\documentclass{article}
\usepackage{microtype}
\usepackage{graphicx}
\usepackage{subfigure}
\usepackage{booktabs}

\usepackage{hyperref}

\usepackage[accepted]{icml2023}

\usepackage
{amsmath}
\usepackage{amssymb}
\usepackage{mathtools}
\usepackage{amsthm}
\usepackage{bm}

\usepackage[capitalize,noabbrev]{cleveref}

\theoremstyle{plain}
\newtheorem{theorem}{Theorem}[section]
\newtheorem{proposition}[theorem]{Proposition}
\newtheorem{lemma}[theorem]{Lemma}
\newtheorem{corollary}[theorem]{Corollary}
\theoremstyle{definition}

\newtheorem{assumption}[theorem]{Assumption}
\theoremstyle{remark}
\newtheorem{remark}[theorem]{Remark}

\usepackage[textsize=tiny]{todonotes}

\icmltitlerunning{Towards Understanding the Generalization of Graph Neural Networks}

\begin{document}

\twocolumn[
\icmltitle{Towards Understanding the Generalization of Graph Neural Networks}

\icmlsetsymbol{equal}{*}

\begin{icmlauthorlist}
\icmlauthor{Huayi Tang}{ruc,laboratory}
\icmlauthor{Yong Liu}{ruc,laboratory}
\end{icmlauthorlist}

\icmlaffiliation{ruc}{Gaoling School of Artificial Intelligence, Renmin University of China, Beijing, China}
\icmlaffiliation{laboratory}{Beijing Key Laboratory of Big Data Management and Analysis Methods, Beijing, China}

\icmlcorrespondingauthor{Yong Liu}{liuyonggsai@ruc.edu.cn}

\icmlkeywords{Machine Learning, ICML}

\vskip 0.3in
]

\printAffiliationsAndNotice{}

\begin{abstract}
Graph neural networks (GNNs) are the most widely adopted model in graph-structured data oriented learning and representation. Despite their extraordinary success in real-world applications, understanding their working mechanism by theory is still on primary stage. In this paper, we move towards this goal from the perspective of generalization. To be specific, we first establish high probability bounds of generalization gap and gradients in transductive learning with consideration of stochastic optimization. After that, we provide high probability bounds of generalization gap for popular GNNs. The theoretical results reveal the architecture specific factors affecting the generalization gap. Experimental results on benchmark datasets show the consistency between theoretical results and empirical evidence. Our results provide new insights in understanding the generalization of GNNs.
\end{abstract}

\section{Introduction}
Graph-structured data \cite{yanqiao} exists widely in real-world applications. As one of the most powerful tools to process graph-structured data, GNNs \cite{gori, scar} are widely adopted in Computer Vision \cite{ DBLP:conf/iccv/QiLJFU17, DBLP:conf/cvpr/JohnsonGF18, DBLP:conf/cvpr/LandrieuS18, DBLP:conf/iclr/SatorrasE18}, Natural Language Processing \cite{DBLP:conf/emnlp/BastingsTAMS17, DBLP:conf/acl/CohnHB18, DBLP:conf/acl/GildeaWZS18}, Recommendation Systems \cite{DBLP:conf/kdd/YingHCEHL18, DBLP:conf/www/Fan0LHZTY19, DBLP:conf/sigir/0001DWLZ020, deng2022graph}, AI for Science \cite{DBLP:conf/icml/Sanchez-Gonzalez20, DBLP:conf/iclr/PfaffFSB21, Shen2021NPIGNNPN, han2022learning}, to name a few. There are two main ways to view modern GNNs, \emph{i.e.}, spatial domain perspective \cite{kipf2017semisupervised, gat, DBLP:conf/icml/XuLTSKJ18, xu2018how} and spectral domain perspective \cite{NIPS2016_04df4d43, gasteiger2018combining, liao2018lanczosnet, chien2021adaptive, he2021bernnet}. The former regards GNN as the process of combining and updating features according to adjacent relationships. The latter treats GNN as a filtering function applied on input features. Recent developments of GNNs are summarized in \cite{ZHOU202057, 9046288, 9039675}.

Despite the empirical success of GNNs, establishing theories to explain their behaviors is still in its infancy. Recent works towards this direction includes understanding over-smoothing \cite{DBLP:conf/aaai/LiHW18, DBLP:conf/iclr/ZhaoA20, DBLP:conf/iclr/OonoS20, Rong2020DropEdge:}, interpretability \cite{DBLP:conf/nips/YingBYZL19, DBLP:conf/nips/LuoCXYZC020, NEURIPS2020_8fb134f2, 10.1145/3394486.3403085, pmlr-v139-yuan21c}, expressiveness \cite{xu2018how, DBLP:conf/nips/ChenVCB19, maron2018invariant, NEURIPS2019_73bf6c41, feng2022how}, and generalization \cite{SCARSELLI2018248, DBLP:conf/nips/DuHSPWX19, verma, pmlr-v119-garg20c, pmlr-v119-zhang20y, oono_2020, Lv2021GeneralizationBF, liao2021a, esser_2021, cong2021on}. This work focuses on the last branch. Some previous works adopt the classical techniques such as Vapnik-Chervonenkis dimension \cite{SCARSELLI2018248}, Rademacher complexity \cite{Lv2021GeneralizationBF, pmlr-v119-garg20c} and algorithm stability \cite{verma} to provide generalization bounds for GCN \cite{kipf2017semisupervised} and more general message passing neural networks. However, in their analysis, the original graph is split into subgraphs composed of central node and its neighbors, which are treated as independent samples. This setting significantly differs from real implementation that training nodes are sampled without replacement from full nodes and the test nodes are visible during training \cite{yaniv_2007, DBLP:conf/iclr/OonoS20}, resulting a gap between theory and practice. To tackle this issue, recent works \cite{oono_2020, esser_2021} incorporate the learning schema of GNNs into the category of transductive learning and derive more realistic results. However, there are still some drawbacks of these works. First, the analysis in \cite{oono_2020} is oriented to multi-scale GNNs that differ a lot from modern GNNs in network architecture. Besides, their analysis is limited to the AdaBoost-like optimization procedure, and whether the technique can be applied to general optimization algorithms such as stochastic gradient descent (SGD) is unknown. Second, the upper bound in \cite{esser_2021} is of slow order and fails to provide meaningful learning guarantee for node classification in large-scale scenarios. Third, \cite{cong2021on} only consider spectral-based GNNs with fixed coefficients, leaving spectral-based GNNs with learnable coefficients \cite{chien2021adaptive} unexplored. 

Motivated by the aforementioned challenges, under transductive setting, we study the generalization gap of GNNs for node classification task with consideration of stochastic optimization algorithm. First, we establish high probability bounds of generalization gap and gradients under transductive setting, and derive high probability bounds of test error under gradient dominant condition. Next, we provide a comprehensive analysis on popular GNNs including both linear and non-linear models and derive the upper bound of the Lipschitz continuity and H{\"o}lder smoothness constants, by which we compare their generalization capability. The results show that SGC \cite{sgc} and APPNP \cite{gasteiger2018combining} can achieve smaller generalization gap than GCN \cite{kipf2017semisupervised}. Besides, the unconstrained coefficients in spectral GNNs may yield a large generalization gap. Our results reveal why shallow models yield comparable and even superior performance from the perspective of learning theory, and provide theoretical supports for widely used techniques such as early stop and drop edge \cite{Rong2020DropEdge:}. Experimental results on benchmark datasets show that the theoretical findings are generally consistent with the practical evidences.

\section{Related Work} 

\subsection{Generalization Analysis of GNNs}
Existing studies on the generalization of GNNs general fall into two categories: graph classification task and node classification task.

\textbf{Graph classification task.} \cite{liao2021a} is the first work to establish generalization bounds of GCN and message passing neural networks by PAC-Bayesian approach. The authors in \cite{ju23generalization} further improve their results and provide the lower bound. Besides, neural tangent kernels \cite{NEURIPS2018_5a4be1fa} are also used to analyze the generalization of infinitely wide GNNs trained by gradient descent \cite{DBLP:conf/nips/DuHSPWX19}. Different from that, this work focus on node classification task that is more challenging. 

\textbf{Node classification task.} The authors in \cite{SCARSELLI2018248} analyze the generalization capability of GNNs by Vapnik–Chervonenkis dimension. \cite{verma} is the first work to provide generalization bounds of one-layer GCN by algorithm stability which is further extended to multi-layer GCNs in \cite{zhou2021enlarge}. The work \cite{pmlr-v119-garg20c} converts the graph into individual local node-wise computation tree and bound their generalization bound respectively by Rademacher Complexity. The aforementioned works rely on the assumption that converting a graph into subgraphs, which differs a lot from realistic implementation. Observing that, \cite{oono_2020} makes the first step that adopting the transductive learning framework to analyze multi-scale GNNs. This framework originates from \cite{DBLP:books/daglib/0097035, DBLP:books/sp/Vapnik06}, and is further developed in \cite{Yaniv06, yaniv_2007} where the authors propose transductive stability and Transductive Rademacher complexity to measure the generalization capability of transductive learner. The work most related to ours is \cite{cong2021on} and \cite{esser_2021}, where the authors establish generalization bound for GNNs and its variants by transductive uniform stability and transductive Rademacher complexity respectively. However, the derived bound in \cite{esser_2021} is of slow order, and whether their technique can be applied on SGD is still unknown. Different from \cite{cong2021on} that analyzing full-batch gradient descant, we analyze a more complex setting, \emph{i.e.}, transductive learning under SGD, due to the involve of randomness in optimization. Besides, there are some works orthogonal to ours, \emph{e.g.}, analyzing the generalization capability of GNNs training with topology-sampling \cite{li22generalization} or on large random graphs \cite{NEURIPS2020_f5a14d49}.

\subsection{Out-of-Distribution (OOD) Generalization on Graphs}

Much efforts are devoted to the study of OOD generalization on graphs \cite{li2022outofdistribution} in recent years, due to the occurs of distribution shift in real-world scenarios. An adversarial learning schema \cite{wu2022handling} is proposed to minimize the mean and variance of risks from multiple environments. The authors in \cite{yang2022learning} propose a two-stage training schema to tackle distribution shift on molecular graphs. Energy-based message passing scheme is show to be effective in enhancing the OOD detection performance of GNNs \cite{wu2023energybased}. Current work \cite{yang2023graph} shows that the spurious performance of GNNs may come from its intrinsic generalization capability rather than expressivity. Besides, there are also some work focus on the reasoning \cite{Xu2020What}, extrapolation ability \cite{xu2021how,BevilacquaZ021}, and generalization from small to large graphs \cite{YehudaiFMCM21}.

\section{Preliminaries}

\subsection{Notations}
Let $\mathcal{G}=\left\{ \mathcal{V}, \mathcal{E} \right\}$ be an given undirected graph with $n=|\mathcal{V}|$ nodes. Each node is an instance $z_i = (\mathbf{x}_i, y_i) $ containing feature $\mathbf{x}_i$ and label $y_i$ from some space $\mathcal{Z}=\mathcal{X} \times \mathcal{Y}$. Let $\mathbf{X}$ be the feature matrix where the $i$-th row $\mathbf{X}_{i*}$ is the node feature $\mathbf{x}_i$. Let $\mathbf{A}$ and $\mathbf{D}$ be the adjacency matrix and the diagonal degree matrix respectively, where $\mathbf{D}_{ii} = \sum_{j=1}^n \mathbf{A}_{ij}$. Denote by $\tilde{\mathbf{A}}=(\mathbf{D}+\mathbf{I}_n)^{-\frac{1}{2}}(\mathbf{A}+\mathbf{I}_n)(\mathbf{D}+\mathbf{I}_n)^{-\frac{1}{2}}$ the normalized adjacency matrix with self-loops and $\sqrt{|\mathcal{Y}|}$ the number of categories. We focus on the transductive learning setting in this work, \emph{i.e.}, all features together with the randomly sampled labels are constructed as training set. Let $S = \{\mathbf{x}_i, y_i\}_{i=1}^{m+u}$ be the set of instances where $m+u=n$. Without loss of generality (w.l.o.g.), let $\{y_{i}\}_{i=1}^m$ be the selected labels, our task is to predict the labels of samples $\{\mathbf{x}_{i}\}_{i=m+1}^{m+u}$ by a learner (model) trained on $\{\mathbf{x}_{i}\}_{i=1}^{m+u} \bigcup \{y_{i}\}_{i=1}^{m}$. This setting is widely adopted in node classification task \cite{DBLP:conf/icml/YangCS16,kipf2017semisupervised} where the training and test nodes are determined by a random partition.

From now on, we limit the scope of the learner to a given GNN and let $\{\mathbf{W}_h\}_{h=1}^H$ be its learnable parameters. Since $\mathbb{R}^{p\times q}$ and $\mathbb{R}^{pq}$ are isomorphic, the analysis in this work is oriented to the vector space for concise. To this end, we use a unified vector $\mathbf{w} = [{\rm vec}\left[ \mathbf{W}_1 \right]; \ldots; {\rm vec}\left[ \mathbf{W}_H \right]]$ to represent the collection of $\{\mathbf{W}_h\}_{h=1}^H$, where ${\rm vec}[\cdot]$ is the vectorization operator that transforms a given matrix into vector, \emph{i.e.}, ${\rm vec}\left[\mathbf{W}\right] = \left[\mathbf{W}_{*1};\cdots;\mathbf{W}_{*q}\right]$ for $\mathbf{W}\in \mathbb{R}^{p\times q}$. Here $\mathbf{W}_{*i}$ is the $i$-th column of $\mathbf{W}$. For $\mathbf{w} \in \mathcal{W}$, the training and test error is defined as $R_m(\mathbf{w}) \triangleq \frac{1}{m}\sum_{i=1}^{m} \ell (\mathbf{w}; z_i)$ and $R_u(\mathbf{w}) \triangleq \frac{1}{u}\sum_{i=m+1}^{m+u} \ell (\mathbf{w}; z_i)$ respectively, where $\ell: \mathcal{W} \times \mathcal{Z} \mapsto \mathbb{R}_+$ is the loss function. In this work, we follow previous studies \cite{yaniv_2007, oono_2020, esser_2021} and define the transductive generalization gap by $\vert R_m(\mathbf{w}) - R_u(\mathbf{w}) \vert$. Since the label of test examples are not available, the optimization process is finding parameters to minimize the the training error $R_m(\mathbf{w})$. Much efforts \cite{duchi11a,KingmaB14} are devoted to solve this stochastic optimization problem, and we mainly focus on SGD (Summarized in Algorithm~\ref{alg}) in this work. 

Now we introduce notations used in the rest of this paper. Denote by ${\Vert \cdot \Vert}_2$ and $\Vert \cdot \Vert$ the $2$-norm of vector and spectral norm of matrix, respectively. Let $\mathbf{w}^{(1)}$ be the initialization weight of model, we focus on the space $\mathcal{W} = B(\mathbf{w}^{(1)}; r), r \geq 1$ in this work, where $B(\mathbf{w}^{(1)}; r) \triangleq \left\{ \mathbf{w}: \left\Vert \mathbf{w} - \mathbf{w}^{(1)} \right\Vert_2 \leq r \right\}$ is the ball with radius $r$. Denote by $\nabla \ell(\cdot;z)$ the gradient of $\ell$ with respective to (w.r.t.) the first argument. Denote by $b_g = \mathop{\rm sup}_{z \in \mathcal{Z}} \left\Vert \nabla \ell(\mathbf{w}^{(1)};z) \right\Vert_2$ the supermum of gradient with initialed parameter and $b_{\ell} = \mathop{\rm sup}_{z \in \mathcal{Z}} \left\vert \ell(\mathbf{w}^{(1)};z) \right\vert_2$ the supermum of loss value with initialed parameter. Let $\hat{\mathbf{w}} \in \mathop{\rm argmin}_{\mathbf{w} \in \mathcal{W}} R_m(\mathbf{w})$ be the parameters of training error minimizer. We denote by $\sigma(\cdot)$ the activation function.

\begin{algorithm}[tb]
   \caption{SGD for Transductive Learning}
   \label{alg}
\begin{algorithmic}
   \STATE {\bfseries Input:} Initial parameter $\mathbf{w}^{(1)}$, learning rates $\{\eta_t\}$, training set $\{\mathbf{x}_i\}_{i=1}^{m+u} \cup \{y_i\}_{i=1}^{m}$.
   \FOR{$t=1$ {\bfseries to} $T$}
   \STATE Randomly draw $j_t$ from the uniform distribution over the set $\{j:j\in [m]\}$.
   \STATE Update parameters by \\
   \quad \quad $\mathbf{w}^{(t+1)} = \mathbf{w}^{(t)} - \eta_t \nabla  \ell(\mathbf{w}^{(t)};z_{j_t})$.\\
   \ENDFOR
\end{algorithmic}
\end{algorithm}

\subsection{Assumptions}
In this part, we present the assumptions used in this paper. 
\begin{assumption}\label{assump1}
Assume that there exists a constant $c_X > 0$ such that $\Vert \mathbf{x} \Vert_2 \leq c_X$ holds for all $\mathbf{x} \in \mathcal{X}$. 
\end{assumption}
\begin{assumption}\label{assump2}
Assume that there exists a constant $c_W > 0$ such that $\Vert \mathbf{W}_h \Vert \leq c_W, h \in [H]$ for $\mathbf{w} \in B(\mathbf{w}^{(1)}; r)$.
\end{assumption}
\begin{remark}
Assumption~\ref{assump1} requires that input features are bounded \cite{verma}. This assumption can be satisfied by applying normalization on features. Assumption~\ref{assump2} means that the parameters during the training process are bounded, which is a common assumption in generalization analysis of GNNs \cite{pmlr-v119-garg20c, liao2021a, cong2021on, esser_2021}. These two assumptions are necessary to analyze the Lipschitz continuity and H\"older smoothness of objective w.r.t. $\mathbf{w}$.
\end{remark}
\begin{assumption}\label{holder}
    Assume that the activation function $\sigma(\cdot)$ is $\tilde{\alpha}$-H\"older smooth. To be specific, let $P > 0$ and $\tilde{\alpha} \in (0,1]$, for all $\mathbf{u}, \mathbf{v} \in \mathbb{R}^d$, 
    \begin{equation*}
        \Vert \sigma'(\mathbf{u}) - \sigma'(\mathbf{v}) \Vert_2 \leq P \Vert \mathbf{u} - \mathbf{v} \Vert_2^{\tilde{\alpha}}.
    \end{equation*}
\end{assumption}
\begin{remark}
    It can be verified that Assumption~\ref{holder} implies Lipschitz continuity of activation function if $\tilde{\alpha}=0$. Besides, Assumption~\ref{holder} implies the smoothness of activation function if $\tilde{\alpha}=1$. Therefore, Assumption~\ref{holder} is much milder than the assumption in previous work \cite{verma,cong2021on} that requires the activation function is smooth. For the convenience of analysis while not yielding a large gap between theory and practice, we construct a modified ReLU function (See Appdendix A) with hyperparameter $q \in (1, 2]$ that satisfies Assumption~\ref{holder} and has a tolerable approximate error to vanilla ReLU function.
\end{remark}

\begin{assumption}\label{bg}
    Assume that there exist a constant $G > 0$ such that for all $z\in S$
    \begin{equation*}
    \begin{aligned}
        & \sqrt{\eta_t} \left\Vert \nabla \ell(\mathbf{w}_t;z) \right\Vert_2 \leq G
    \end{aligned}
    \end{equation*}
    holds $\forall \ t \in \mathbb{N}$, where $\{\eta_t\}_{t=1}^T$ is learning rates.
\end{assumption}
\begin{remark}
    A formal definition of $\nabla \ell(\mathbf{w};z)$ is provided in Lemma~\ref{bound_alpha} in the Appendix. Assumption~\ref{bg} \cite{lei, Li2021ImprovedLR} means that the product of gradient and the square root of learning rate is bounded, which is milder than the widely used bounded gradient assumption \cite{DBLP:conf/icml/HardtRS16, pmlr-v80-kuzborskij18a}, since the learning rate tends to zero during the iteration.
\end{remark}
\begin{assumption}\label{bn}
    Assume that there exists a constant $\sigma_0 > 0$ such that for $\forall \ t \in \mathbb{N}_+$, the following inequality holds
    \begin{equation*}\small
    \begin{aligned}
        \mathbb{E}_{j_t} \left[ \Vert \nabla \ell(\mathbf{w});z_{j_t})) \Vert_2 \right] \leq \sigma^2_0.
    \end{aligned}
    \end{equation*}
\end{assumption}
\begin{remark}
    Assumption~\ref{bn} requires the boundness of variances of stochastic gradients, which is a standard assumption in stochastic optimization studies \cite{pmlr-v80-kuzborskij18a, lei, Li2021ImprovedLR}. 
\end{remark}

\section{Theoretical Results}
In this section, we first present the high probability bounds of generalization gap and excess risks under transductive learning in Section~\ref{transductive}. After that, we turn to specific examples and provide results of some popular GNNs in Section~\ref{gnn_bound}. Please refer to the Appendix for complete proofs.

\subsection{General Results of Transductive SGD}\label{transductive}
We first analyze properties of the objective function $\ell$ and provide the following proposition.

\begin{proposition}[Informal]\label{pro}
Suppose Assumptions~\ref{assump1}, \ref{assump2}, and \ref{holder} hold. Denote by $\mathcal{F}$ a specific GNN, for any $\mathbf{w}, \mathbf{w}' \in \mathcal{W}$ and $z\in S$, the objective $\ell(\mathbf{w};z)$ satisfies
\begin{equation}
    | \ell(\mathbf{w};z) - \ell(\mathbf{w}';z) | \leq L_{\mathcal{F}} \Vert \mathbf{w} - \mathbf{w}' \Vert_2,
\end{equation}
and
\begin{equation}
\begin{aligned}
    & \Vert \nabla \ell(\mathbf{w};z) - \nabla \ell(\mathbf{w}';z) \Vert \\
    \leq & P_{\mathcal{F}} \mathop{\rm max} \left\{ \Vert \mathbf{w} - \mathbf{w}' \Vert^{\tilde{\alpha}}_2, \Vert \mathbf{w} - \mathbf{w}' \Vert_2 \right\},
\end{aligned}
\end{equation}
with constant $L_{\mathcal{F}}$ and $P_{\mathcal{F}}$.
\end{proposition}
\begin{remark}
    We provide more detailed analysis to $L_{\mathcal{F}}$ and $P_{\mathcal{F}}$ in Section~\ref{gnn_bound}. Both $L_{\mathcal{F}}$ and $P_{\mathcal{F}}$ depend on the specific network architecture $\mathcal{F}$ of GNNs. Thus, the upper bound of generalization gap vary by the architecture. 
\end{remark}
Our first main result is high probability bounds on the transductive generalization gap, as presented in Theorem~\ref{th1}.
\begin{theorem}\label{th1}
Suppose Assumptions~\ref{assump1}, \ref{assump2}, \ref{holder}, \ref{bg}, and \ref{bn} hold. Suppose that the learning rate $\{\eta_t\}$ satisfies $\eta_t = \frac{1}{t+t_0}$ such that $t_0 \geq {\rm max}\{({2P})^{1/\alpha}, 1 \}$. For any $\delta \in (0,1)$, with probability $1-\delta$, 
\begin{itemize}
    \item[(a).] If $\alpha \in (0,\frac{1}{2})$, we have
    \begin{equation*}
    \begin{aligned}
        & R_u(\mathbf{w}_1^{(T+1)}) - R_m(\mathbf{w}^{(T+1)}) \\
        = & \mathcal{O} \bigg( L_{\mathcal{F}} \frac{{(m+u)}^{\frac{3}{2}}}{mu} \log^{\frac{1}{2}}(T)T^\frac{1-2\alpha}{2}\log \bigg( \frac{1}{\delta} \bigg) \bigg).
    \end{aligned}
    \end{equation*}
    \item[(b).] If $\alpha = \frac{1}{2}$, we have
    \begin{equation*}
    \begin{aligned}
        & R_u(\mathbf{w}^{(T+1)}) - R_m(\mathbf{w}^{(T+1)}) \\
        = & \mathcal{O} \bigg( L_{\mathcal{F}} \frac{{(m+u)}^{\frac{3}{2}}}{mu} \log(T) \log \bigg( \frac{1}{\delta} \bigg) \bigg).
    \end{aligned}
    \end{equation*}
    \item[(c).] If $\alpha \in (\frac{1}{2}, 1]$, we have
    \begin{equation*}
    \begin{aligned}
        & R_u(\mathbf{w}^{(T+1)}) - R_m(\mathbf{w}^{(T+1)}) \\
        = & \mathcal{O} \bigg( L_{\mathcal{F}} \frac{{(m+u)}^{\frac{3}{2}}}{mu} \log^{\frac{1}{2}}(T)\log \bigg( \frac{1}{\delta} \bigg) \bigg).
    \end{aligned}
    \end{equation*}
\end{itemize}
\end{theorem}
\begin{remark}
    Theorem~\ref{th1} shows that the transductive generalization gap depends on the training/test data size $m/u$, network architecture related Lipschitz continuity constant $L_{\mathcal{F}}$, and the number of iterations $T$. Generally, our upper bounds are of order $\mathcal{O}\left((\frac{1}{m}+\frac{1}{u})\sqrt{m+u} \right)$, which is much sharper than the bound $\mathcal{O}\left((\frac{1}{m}+\frac{1}{u})(m+u) + \log(m+u) \right)$ in previous work \cite{esser_2021}. Note that with the increase of data size $m+u$, the bound in \cite{esser_2021} become increasing larger and fail to provide a reasonable generalization guarantee. This seriously restricts its application in large-scale node classification scenarios where the order of $m+u$ is usually millions. Our results address these drawbacks and provide more applicable generalization guarantee for GNNs. Besides, the bound provided in \cite{esser_2021} does not consider the specific optimization and has difficulty in revealing the influence of $T$ on generalization gap. Our result shows that the generalization gap becomes larger when the number of $T$ increases, resulting in the over-fitting phenomenon. Thus, early stop may be beneficial for yielding a smaller generalization gap, which is widely adopted in implementation of modern GNNs \cite{kipf2017semisupervised, DBLP:conf/icml/ChenWHDL20}. It can be seen that the generalization gap is positively related to the Lipschitz continuity constant $L_{\mathcal{F}}$ determined by specific network architecture $\mathcal{F}$. Thus, larger $L_{\mathcal{F}}$ leads to larger upper bounds of generalization gap, showing that the network architecture of GNN also have a significant influence on the generalization gap (See Section~\ref{gnn_bound} for more detail). The upper bound of generalization gap in \cite{cong2021on} also increase with $T$ when the objective is optimized by full-batch gradient descent. This is not surprise since it can be seen as a special case of SGD where the batch size is equal to the size of traning samples.
\end{remark}
Our second main result is high probability bounds of the gradients on training and test data. 
\begin{theorem}\label{th2}
    Suppose Assumptions~\ref{assump1}, \ref{assump2}, \ref{holder}, \ref{bg}, and \ref{bn} hold. Suppose that the learning rate $\{\eta_t\}$ satisfies $\eta_t = \frac{1}{t+t_0}$ such that $t_0 \geq {\rm max}\{({2P})^{1/\alpha}, 1 \}$.
    For any $\delta \in (0,1)$, with probability $1-\delta$, 
    \begin{itemize}
    \item[(a).] If $\alpha \in (0,\frac{1}{2})$, we have
        \end{itemize}
        \begin{equation*}
        \begin{aligned}
            & \left\Vert \nabla R_m(\mathbf{w}^{(T+1)}) - \nabla R_u(\mathbf{w}^{(T+1)}) \right\Vert_2 \\
            = & \mathcal{O} \bigg( \frac{{(m+u)}^{\frac{3}{2}}}{mu} \log^{\frac{1}{2}}(T)T^\frac{1-2\alpha}{2}\log \bigg( \frac{1}{\delta} \bigg) \bigg).
        \end{aligned}
        \end{equation*}
        \begin{itemize}
        \item[(b).] If $\alpha = \frac{1}{2}$, we have
        \end{itemize}
        \begin{equation*}
        \begin{aligned}
            & \left\Vert \nabla R_m(\mathbf{w}^{(T+1)}) - \nabla R_u(\mathbf{w}^{(T+1)}) \right\Vert_2 \\
            = & \mathcal{O} \bigg( \frac{{(m+u)}^{\frac{3}{2}}}{mu} \log(T) \log \bigg( \frac{1}{\delta} \bigg) \bigg).
        \end{aligned}
        \end{equation*}
        \begin{itemize}
        \item[(c).] If $\alpha \in (\frac{1}{2}, 1]$, we have
        \end{itemize}
        \begin{equation*}
        \begin{aligned}
            & \left\Vert \nabla R_m(\mathbf{w}^{(T+1)}) - \nabla R_u(\mathbf{w}^{(T+1)}) \right\Vert_2 \\
            = & \mathcal{O} \bigg( \frac{{(m+u)}^{\frac{3}{2}}}{mu} \log^{\frac{1}{2}}(T) \log \bigg( \frac{1}{\delta} \bigg) \bigg).
        \end{aligned}
        \end{equation*}
\end{theorem}
\begin{remark}
    Theorem~\ref{th2} provides high probability bounds for the generalization gap of gradients under transductive setting. Overall, the generalization gap we derive is still of order $\mathcal{O}\left((\frac{1}{m}+\frac{1}{u})\sqrt{m+u} \right)$, which is applicable in real-world large-scale graph dataset. Besides, the generalization gap of gradients increases with the increase of $T$, showing that a smaller number of iterations helps achieving a smaller generalization gap of gradients.
\end{remark}
Since the generalization performance is determined by both training error and generalization gap, we provide a upper bound of the test error under a special case that the objective satisfies the following PL condition.
\begin{assumption}\label{pl1}
    Suppose that there exists a constant $\mu$ such that for all $\mathbf{w} \in \mathcal{W}$,
    \begin{equation*}
    \begin{aligned}
        & R_m(\mathbf{w}) - R_m(\hat{\mathbf{w}}^*) \leq \frac{1}{2\mu} \left\Vert \nabla R_m(\mathbf{w}) \right\Vert_2, \\
    \end{aligned}
    \end{equation*}
    holds for the given set $S$ from $\mathcal{Z}$.
\end{assumption}
\begin{remark}
    Assumption~\ref{pl1} is also named as gradient dominance condition in learning theory studies, indicating that the difference between the optimal training error and the current training error can be upper bounded by the quadratic function of the gradient on training instances. This assumption is widely adopted in nonconvex learning \cite{dongruo, xuyun, lei, Li2021ImprovedLR}, and has been verified in over-parameterized systems including wide neural networks \cite{Liu2020TowardAT}. This assumption only appears in Theorem~\ref{th3}.
\end{remark}
\begin{corollary}\label{th3}
    Suppose Assumptions~\ref{assump1}, \ref{assump2}, \ref{holder}, \ref{bg}, \ref{bn}, and \ref{pl1} hold. Suppose that the learning rate $\{\eta_t\}$ satisfies $\eta_t = \frac{2}{\mu(t+t_0)}$ such that $t_0 \geq {\rm max}\{ \frac{2}{\mu}(2P)^{\frac{1}{\alpha}}, 1 \}$. For any $\delta \in (0,1)$, with probability $1-\delta$, 
        \begin{itemize}
        \item[(a).] If $\alpha \in (0, \frac{1}{2})$, we have
        \end{itemize}
        \begin{equation*}\small
        \begin{aligned}
        & R_u(\mathbf{w}^{(T+1)}) - R_m(\mathbf{w}^*) \\
        = & \mathcal{O} \bigg( L_{\mathcal{F}} \frac{{(m+u)}^{\frac{3}{2}}}{mu} \log^{\frac{1}{2}}(T)T^{\frac{1}{2}-\alpha}\log \bigg( \frac{1}{\delta} \bigg) + \frac{1}{T^\alpha} \bigg),
        \end{aligned}
        \end{equation*}
        \begin{itemize}
        \item[(b).] If $\alpha = \frac{1}{2}$, we have
        \end{itemize}
        \begin{equation*}\small
        \begin{aligned}
            & R_u(\mathbf{w}^{(T+1)}) - R_m(\mathbf{w}^*) \\
            = & \mathcal{O} \bigg( L_{\mathcal{F}} \frac{{(m+u)}^{\frac{3}{2}}}{mu} \log(T) \log \bigg( \frac{1}{\delta} \bigg) + \frac{1}{T^\alpha} \bigg).
        \end{aligned}
        \end{equation*}
        \begin{itemize}
        \item[(c).] If $\alpha \in (\frac{1}{2}, 1)$, we have
        \end{itemize}
        \begin{equation*}\small
        \begin{aligned}
            & R_u(\mathbf{w}^{(T+1)}) - R_m(\mathbf{w}^*) \\
            = & \mathcal{O} \bigg( L_{\mathcal{F}} \frac{{(m+u)}^{\frac{3}{2}}}{mu} \log^{\frac{1}{2}}(T) \log (1/\delta) + \frac{1}{T^\alpha} \bigg).
        \end{aligned}
        \end{equation*}
        \begin{itemize}
        \item[(d).] If $\alpha = 1$, we have
        \end{itemize}
        \begin{equation*}\small
        \begin{aligned}
             & R_u(\mathbf{w}^{(T+1)}) - R_u(\mathbf{w}^*) \\
             = & \mathcal{O} \bigg( L_{\mathcal{F}} \frac{{(m+u)}^{\frac{3}{2}}}{mu} \log^{\frac{1}{2}}(T) \log (1/\delta) + \frac{\log(T) \log^3 (1/\delta)}{T} \bigg).
        \end{aligned}
        \end{equation*}
\end{corollary}
\begin{remark}
    Theorem~\ref{th3} shows that under Assumption~\ref{pl1}, the test error are determined by the minimal training error, optimization error and generalization gap. The minimal training error reflects how well the model fits data, which is a measure of the expressive ability. The first and the second term in the slack terms are generalization gap and optimization error, respectively. With the increase of $T$, the generalization gap increase while the optimization error decrease.
    Therefore, it is necessary to carefully choose a proper number of iterations in order to balance the trade-off between optimization and generalization. In the implementation of most GNNs studies \cite{kipf2017semisupervised, gat, chien2021adaptive, he2021bernnet}, early stop is widely adopted and $T$ is determined by the performance of model on validation set. Thus, our results are consistent with real implementations.
\end{remark}

It is worth point out that although the results in this section is oriented to the case that the objective has two parameters (\emph{e.g.}, GCN, APPNP, and GPR-GNN in Section~\ref{gnn_bound}) , results for other cases that the objective has one parameter (\emph{e.g.}, SGC in Section~\ref{gnn_bound}) or three parameters (\emph{e.g.}, GCNII in Section~\ref{gnn_bound}) have the same form when neglecting the constant factors. Meanwhile, the assumptions need to be modified correspondingly. Readers are referred to the Appendix for detailed discussion.

\subsection{Cases Study of Popular GNNs}\label{gnn_bound}
We have established high probability bounds for transductive generalization gap in Theorem~\ref{th1}. In this part, we analyze the upper bounds of architecture related constant $L_{\mathcal{F}}$ and $P_{\mathcal{F}}$, with that the upper bound of generalization gap can be determined. Five representative GNNs, including GCN, GCNII, SGC, APPNP, and GPR-GNN, are selected for analysis. The loss function $\ell$ is cross-entropy loss and denote by $\hat{\mathbf{Y}}$ the prediction. For concise, we do not consider the bias term, since it can be verified that $\langle \mathbf{w}, \mathbf{x} \rangle + b = \langle \tilde{\mathbf{w}}, \tilde{\mathbf{x}} \rangle $ holds with $\tilde{\mathbf{w}} = [\mathbf{w};b]$ and $\tilde{\mathbf{x}}=[\mathbf{x};1]$.

\textbf{GCN.} The work \cite{kipf2017semisupervised} proposes to aggregate features from one-hop neighbor nodes. The feature propagation process of a two-layer GCN model is
\begin{equation}
    \hat{\mathbf{Y}} = {\rm Softmax}\big( g(\tilde{\mathbf{A}})\sigma(g(\tilde{\mathbf{A}})\mathbf{X}\mathbf{W}_1)\mathbf{W}_2 \big),
\end{equation}
where $g(\tilde{\mathbf{A}}) = \tilde{\mathbf{A}}$ and $\mathbf{W}_1 \in \mathbb{R}^{d\times h}, \mathbf{W}_2 \in \mathbb{R}^{h\times |\mathcal{Y}|}$ are parameters.
\begin{proposition}\label{gnn}
    Suppose Assumptions~\ref{assump1}, \ref{assump2}, and \ref{holder} hold, then the objective $\ell(\mathbf{w};z)$ is $L_{\mathcal{F}}$-Lipschitz continuous and H\"older smooth w.r.t. $\mathbf{w} = \left[ {\rm vec}\left[ \mathbf{W}_1 \right]; {\rm vec}\left[ \mathbf{W}_1 \right] \right]$. Concretely, the Lipschitz continuity constant $L_{\mathcal{F}}$ is $L_{\rm GCN} = 2 c_X c_W \big\Vert \tilde{\mathbf{A}} \big\Vert^2_\infty$.
\end{proposition}

Due to the tedious formulation, we provide the concrete value of $P_{\mathcal{F}}$ in the Appendix. Proposition~\ref{gnn} demonstrates that $L_{\rm GCN}$ mainly depends on factors $\Vert g(\tilde{\mathbf{A}}) \Vert_{\infty}$, $c_X$, and $c_W$. Let ${\rm deg}_{\rm min}$ and ${\rm deg}_{\rm max}$ be the minimum and maximum node degree, respectively. By Lemma~\ref{infty_norm} in Appendix A,
\begin{equation}
    \big\Vert \tilde{\mathbf{A}} \big\Vert_{\infty} \leq \sqrt{\frac{{\rm deg}_{\rm max}+1}{{\rm deg}_{\rm min}+1}}.
\end{equation}
It can be found that the generalization gap decreases with the decrease of the maximum node degree, which could be achieved by removing edges. This explains in some sense why the DropEdge \cite{Rong2020DropEdge:} technique is beneficial for alleviating the over-fitting problem from the perspective of learning theory. Besides, for GCN trained on sampled sub-graphs $\{\mathcal{G}_i\}_{i=1}^n$, the Lipschitz continuity constant is $L_{\rm GCN} = 2 c_X c_W \mathop{\rm max}_{i\in [n]} \big\Vert \tilde{\mathbf{A}}^{[i]} \big\Vert^2_\infty$, where $\tilde{\mathbf{A}}^{[i]}$ is the normalized adjacency matrix with self-loop of $\mathcal{G}_i$. Since only a portion of neighboring nodes are preserved during sub-graphs sampling \cite{sage,Zeng2020GraphSAINT,DBLP:conf/nips/ZengZXSMKPJC21}, the maximum node degree of each sub-graph is smaller than that of initial graph, implying $\mathop{\rm max}_{i\in [n]} \big\Vert \tilde{\mathbf{A}}^{[i]} \big\Vert_\infty \leq \big\Vert \tilde{\mathbf{A}} \big\Vert_{\infty}$ holds. Thus, Proposition~\ref{gnn} shows that training on sampled sub-graphs are beneficial to achieve smaller generalization gap. Lastly, the spectral norm of learning parameters also has an effect on the generalization gap. Thus, the commonly used $L_2$ regularization technique is beneficial to reduce the generalization gap.

\textbf{GCNII.} The authors in \cite{DBLP:conf/icml/ChenWHDL20} propose to relieve over-smoothing by initial residual and identity mapping. Denote by $\mathbf{H}^{(0)} = \sigma (\mathbf{X} \mathbf{W}_0)$ the initial representation. The forward propagation of a two-layer GCNII model is
\begin{equation*}
\begin{aligned}
    \mathbf{H}^{(1)} & = \sigma \Big( ((1-\alpha_1)g(\tilde{\mathbf{A}})\mathbf{H}^{(0)} + \alpha_1 \mathbf{H}^{(0)} ) \Psi(\beta_1, \mathbf{W}_1) \Big), \\
    \mathbf{H}^{(2)} & = \sigma \Big( ((1-\alpha_1)g(\tilde{\mathbf{A}})\mathbf{H}^{(1)} + \alpha_1 \mathbf{H}^{(0)} ) \Psi(\beta_2, \mathbf{W}_2) \Big), \\
    \hat{\mathbf{Y}} & = {\rm softmax}\big( \mathbf{H}^{(2)} \mathbf{W}_3 \big),
\end{aligned}
\end{equation*}
where $\Psi(\beta, \mathbf{W}) = (1-\beta) \mathbf{I} + \beta \mathbf{W}$ and $g(\tilde{\mathbf{A}})=\tilde{\mathbf{A}}$. $\mathbf{W}_1 \in \mathbb{R}^{d\times h}$, $\mathbf{W}_2 \in \mathbb{R}^{h \times h}$, and $\mathbf{W}_3 \in \mathbb{R}^{h\times |\mathcal{Y}|}$ are parameters.
\begin{proposition}\label{gcnii}
    Suppose Assumptions~\ref{assump1}, \ref{assump2}, and \ref{holder} hold, then the objective $\ell(\mathbf{w};z)$ is $L_{\mathcal{F}}$ Lipschitz continuous and H\"older smooth w.r.t. 
    $$\mathbf{w} = \left[ {\rm vec}\left[ \mathbf{W}_0 \right]; {\rm vec}\left[ \mathbf{W}_1 \right]; {\rm vec}\left[ \mathbf{W}_2 \right]; {\rm vec}\left[ \mathbf{W}_3 \right] \right].
    $$
    Specifically, denote by $C_{\ell} = 1-\beta_{\ell} + \beta_\ell c_W, \ell \in [2]$ and 
    \begin{equation}
    \begin{split}
        B_1 & = c_X c_W C_1 \big( (1-\alpha_1) \big\Vert \tilde{\mathbf{A}} \big\Vert_\infty + \alpha_1 \big), \\
        B_2 & = \big( (1-\alpha_2) B_1 \big\Vert \tilde{\mathbf{A}} \big\Vert_\infty + \alpha_2 c_X c_W \big) C_2, \\
        L_1 & = 2 \bigg( 2 + \frac{c^2_W \beta^2_2}{C^2_2} \bigg) B^2_2, \\
        L_2 & = 2 (1-\alpha_2)^2 \beta^2_1 c^2_W \big\Vert \tilde{\mathbf{A}} \big\Vert^2_\infty \bigg( \frac{B^2_1 C^2_2}{C^2_1} \bigg).
    \end{split}
    \end{equation}
    The Lipschitz continuity constant is $L_{\rm GCNII} = \sqrt{L_1 + L_2}$.
\end{proposition}

Proposition~\ref{gcnii} shows that $L_{\rm GCNII}$ is a function of $\{\alpha_i \}_{i=1}^2$ and $\{\beta_i \}_{i=1}^2$. Finding the optimal value of $L_{\rm GCNII}$ is a quadratic programming problem with constrain $\alpha_1, \alpha_2 \in [0, 1]$ and $\beta_1, \beta_2 \in [0, 1]$. Now we discuss a special case that $\alpha_1 = \alpha_2 = 0$ and $\beta_1 = \beta_2 = 0$. In this case, we have $L_1 = 4 c^2_X c^2_W \big\Vert \tilde{\mathbf{A}} \big\Vert^4_\infty $ and $L_2 = 0$, which implies that $L_{\rm GCNII} = L_{\rm GCN}$. Note that the optimal value of $L_{\rm GCNII}$ is no larger than any value of objective function over the feasible region. Therefore, we conclude that the value of $L_{\rm GCNII}$ is no higher than $L_{\rm GCN}$. This result is not surprise, since GCNII is a special GCN model under this setting. For proper value of $\{\alpha_i \}_{i=1}^2$ and $\{\beta_i \}_{i=1}^2$, GCNII could achieve smaller generalization gap than GCN. As GCNII can achieve lower training error by relieving the over-smoothing problem, Proposition~\ref{gcnii} indicates that GCNII can achieve superior performance when hyperparameters are set properly. Due to the involve of $\{\alpha_i \}_{i=1}^2$ and $\{\beta_i \}_{i=1}^2$, the growth rate of $L_{\rm GCNII}$ is much smaller than $L_{\rm GCN}$ when propagation depth increases, which makes GCNII maintain generalization capability and achieve stale performance (See Section~\ref{exp}). 

\textbf{SGC.} The work \cite{sgc} proposes to remove all the nonlinear activation in GCN. 
To facilitate comparison with GCN, we consider a two layers SGC model, whose propagation is given by
\begin{equation}\label{sgc_prop}
    \hat{\mathbf{Y}} = {\rm softmax}\big( g(\tilde{\mathbf{A}}) \mathbf{X}\mathbf{W}_1 \mathbf{W}_2 \big),
\end{equation}
where $g(\tilde{\mathbf{A}}) = \tilde{\mathbf{A}}^2$. $\mathbf{W}_1 \in \mathbb{R}^{d \times h}$ and $\mathbf{W}_2 \in \mathbb{R}^{h \times |\mathcal{Y}|}$ is the parameter. 
\begin{proposition}\label{sgc}
    Suppose Assumption~\ref{assump1}, \ref{assump2}, and \ref{holder} hold, then the objective $\ell(\mathbf{w};z)$ is $L_{\mathcal{F}}$-Lipschitz continuous and H\"older smooth w.r.t. $\mathbf{w} = \left[ {\rm vec}\left[ \mathbf{W}_1 \right]; {\rm vec}\left[ \mathbf{W}_2 \right] \right]$. Specifically, the Lipschitz continuity constant $L_{\mathcal{F}}$ is $L_{\rm SGC} = 2 c_X c_W \big\Vert \tilde{\mathbf{A}}^2 \big\Vert_\infty$.
\end{proposition}

Since $\big\Vert \tilde{\mathbf{A}}^2 \big\Vert_\infty \leq \big\Vert \tilde{\mathbf{A}} \big\Vert^2_\infty$, we have $L_{\rm SGC} \leq L_{\rm GCN}$. Surprisingly, this simple linear model can achieve better smaller generalization gap than those nonlinear models \cite{kipf2017semisupervised, DBLP:conf/icml/ChenWHDL20, gasteiger2018combining, chien2021adaptive}, even though its representation ability is inferior than them. Note that the performance on test samples is determined by both training error and generalization gap. If linear GNNs can achieve a small training error, it is natural that they can achieve comparable and even better performance than nonlinear GNNs on test samples. Therefore, Proposition~\ref{sgc} reveals why linear GNNs achieve better performance than nonlinear GNNs from learning theory, as observed in recent works \cite{sgc, DBLP:conf/iclr/ZhuK21, yifei}. Considering the efficiency and scalability of linear GNNs on large-scale datasets, we believe that they have much potential to be exploited.

\textbf{APPNP.} Multi-scale features are aggregated via personalized PageRank schema in \cite{gasteiger2018combining}. Formally, the feature propagation process is formulated as
\begin{equation}\label{app_up}
\begin{aligned}
    \hat{\mathbf{Y}} = {\rm softmax} \big(g(\tilde{\mathbf{A}})\sigma(\sigma(\mathbf{X}\mathbf{W}_1)\mathbf{W}2) \big),
\end{aligned}
\end{equation}
where $g(\tilde{\mathbf{A}}) = \sum_{k=0}^{K-1} \gamma(1-\gamma)^k \tilde{\mathbf{A}}^k + (1-\gamma)^K\tilde{\mathbf{A}}^K$. $\mathbf{W}_1 \in \mathbb{R}^{d\times h}$ and $\mathbf{W}_2 \in \mathbb{R}^{h \times |\mathcal{Y}|}$ are the parameters.
\begin{proposition}\label{appnp}
    Suppose Assumption~\ref{assump1}, \ref{assump2}, and \ref{holder} hold, then the objective $\ell(\mathbf{w};z)$ is $L_{\mathcal{F}}$-Lipschitz continuous and H\"older smooth w.r.t. $\mathbf{w} = \left[ {\rm vec}\left[ \mathbf{W}_1 \right]; {\rm vec}\left[ \mathbf{W}_2 \right] \right]$. Concretely, the Lipschitz continuity constant $L_{\mathcal{F}}$ is $L_{\rm APPNP} = 2 c_X c_W \big\Vert g(\tilde{\mathbf{A}}) \big\Vert_\infty$.
\end{proposition}
The Lipschitz continuity constant in Proposition~\ref{appnp} is positively related to the infinity matrix norm of the polynomial spectral filter. According to \cite{gasteiger2018combining}, $\gamma$ is commonly set to be a small number, yielding that $\big\Vert g(\tilde{\mathbf{A}}) \big\Vert_{\infty} < \big\Vert \tilde{\mathbf{A}} \big\Vert_{\infty}$ holds. Thus, the Lipschitz continuity constant of APPNP is smaller than that of GCN, indicating that APPNP may achieve smaller generalization gap than GCN. Besides, $K$ also affects the value of $\Vert g(\tilde{\mathbf{A}}) \Vert_{\infty}$, and a larger $K$ may yield a larger generalization gap. Therefore, $K$ is usually set as a proper value to guarantee a trade-off between expressive ability and generalization performance.

\textbf{GPR-GNN.} Compared with APPNP, the fixed coefficients are replaced by learnable weights in \cite{chien2021adaptive}, in order to adaptively simulate both high-pass and low-pass graph filters. The feature propagation process is 
\begin{equation}
\begin{aligned}
    \hat{\mathbf{Y}} & = \big(g(\tilde{\mathbf{A}}, \bm \gamma)\sigma(\sigma(\mathbf{X}\mathbf{W}_1)\mathbf{W}2) \big),
\end{aligned}
\end{equation}
where $g(\tilde{\mathbf{A}}, \bm \gamma) = \sum_{k=0}^{K} \gamma_k \tilde{\mathbf{A}}^k$. $\mathbf{W}_1 \in \mathbb{R}^{d\times h}$,  $\mathbf{W}_2 \in \mathbb{R}^{h \times |\mathcal{Y}|}$ and ${\bm \gamma} \in \mathbb{R}^{K+1}$ are the parameters.

\begin{proposition}\label{gpr}
    Suppose Assumption~\ref{assump1}, \ref{assump2}, and \ref{holder} hold, then the objective $\ell(\mathbf{w};z)$ is $L_{\mathcal{F}}$-Lipschitz continuous and H\"older smooth w.r.t. $\mathbf{w} = \left[ {\rm vec}\left[ \mathbf{W}_1 \right]; {\rm vec}\left[ \mathbf{W}_2 \right]; {\bm \gamma} \right]$. Concretely, the Lipschitz continuity constant $L_{\mathcal{F}}$ is $ L_{\rm GPR} = \sqrt{L^2_1 + L^2_2}$, where
    \begin{equation}
    \begin{aligned}
        L_1 & = \sqrt{2} c_X c^2_W \bigg( \sum_{k=0}^K \big\Vert \tilde{\mathbf{A}}^k \big\Vert_{\infty} \bigg), \\
        L_2 & = 2c_X c_W \big\Vert g(\tilde{\mathbf{A}}, {\bm \gamma}) \big\Vert_{\infty}.
    \end{aligned}
    \end{equation}
\end{proposition}
Note that $L_2$ has similar form with $L_{\rm APPNP}$ (the only difference lie on the definition of $g(\tilde{\mathbf{A}}, {\bm \gamma})$). Assume that $g(\tilde{\mathbf{A}}, {\bm \gamma}) = g(\tilde{\mathbf{A}})$ and note that $L_{\rm GPR} = \sqrt{L^2_1 + L^2_2} \geq L_2$, we have $L_{\rm GPR} \geq L_{\rm APPNP}$. Besides, since there is no constraint on $\gamma$, the value of $\big\Vert g(\tilde{\mathbf{A}}, {\bm \gamma}) \big\Vert_{\infty}$ may be larger when the norm of $\bm \gamma$ is large, resulting in larger generalization gap than APPNP. Therefore, adopting regularization technique on the learnable coefficients to restrict the value of $\big\Vert g(\tilde{\mathbf{A}}, {\bm \gamma}) \big\Vert_{\infty}$ is necessary.

To summarize, $L_{\mathcal{F}}$ and $P_{\mathcal{F}}$ are determined by the feature propagation process and graph-structured data. Estimating these constants precisely is challenging \cite{virmaux2018,fazlyab2019}, and the upper bounds we provided are sufficient to reflect the realistic generalization gap of these models (See Section~\ref{exp} for more detail). Besides, we have to emphasize that results for GCN and GCNII with more than two layers can be derived by similar techniques, yet it requires more tedious computation. Exploring new techniques to estimate these constants conveniently and precisely are left for future work.

\begin{figure*}[ht]
\vskip 0.1in
\begin{center}
\subfigure{
    \includegraphics[width=0.31\textwidth]{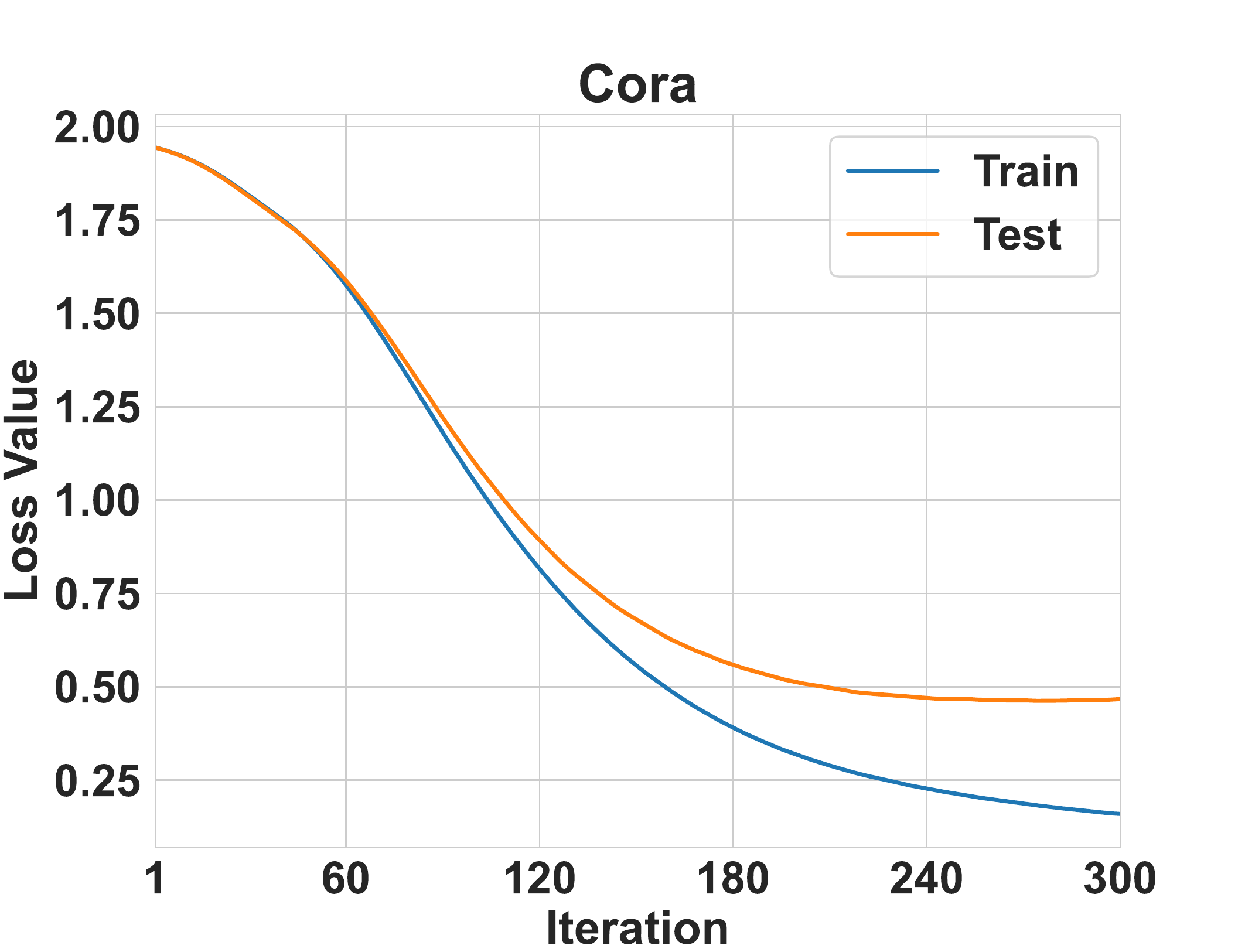}
  }
  \hspace{-0.3mm}
  \subfigure{
    \includegraphics[width=0.31\textwidth]{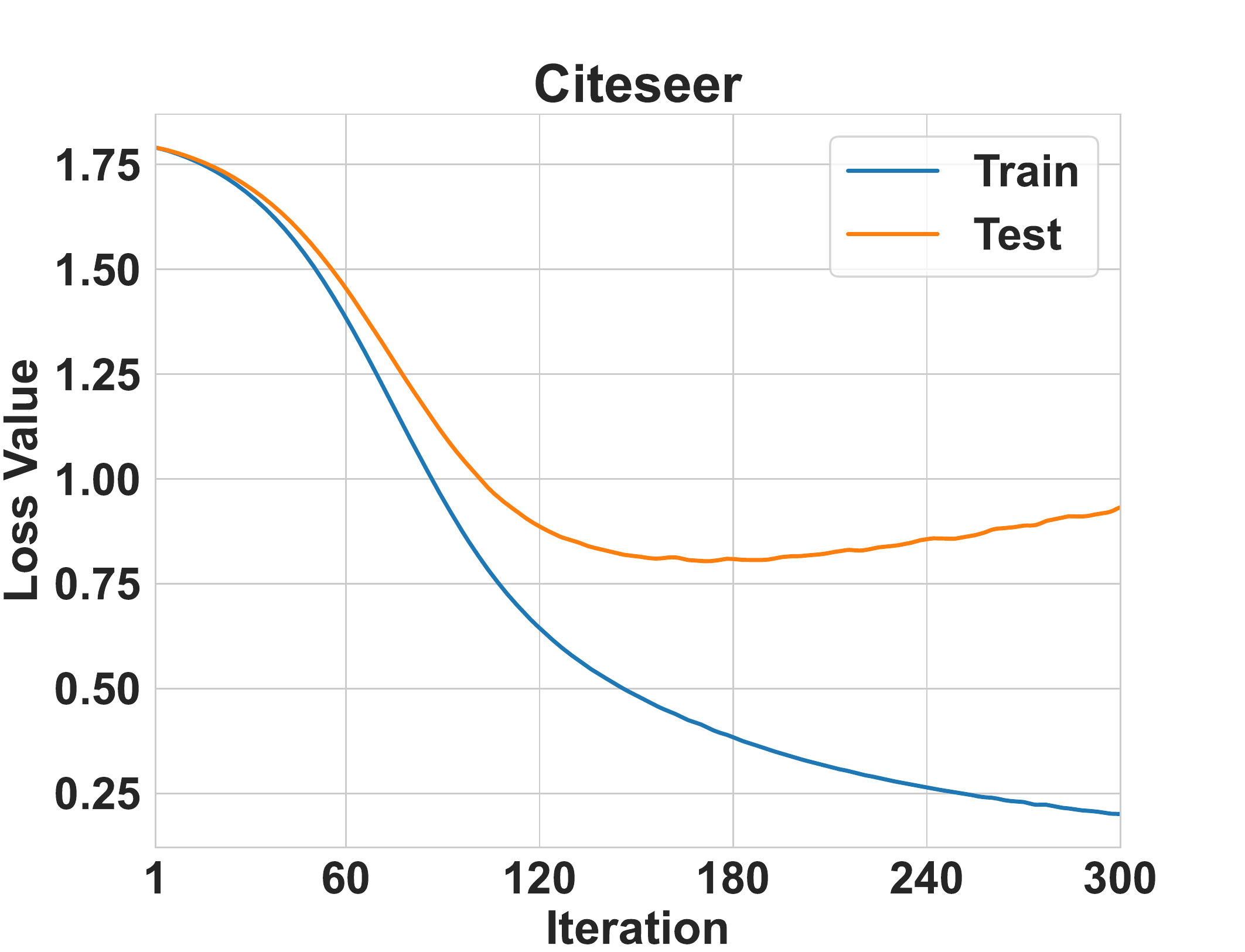}
  }
  \hspace{-0.3mm}
  \subfigure{
    \includegraphics[width=0.31\textwidth]{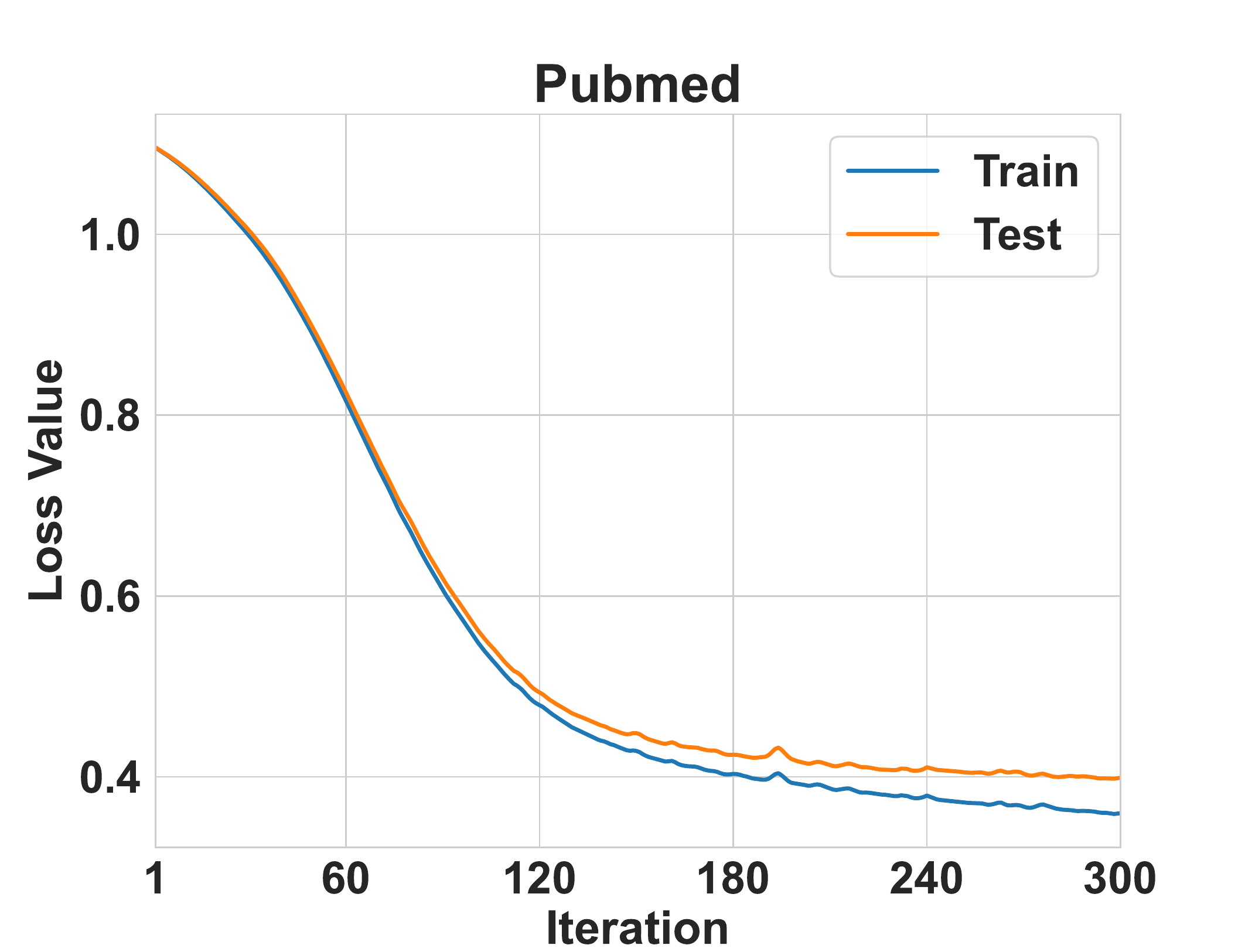}
  }
\caption{The loss value of GCN on training and test samples with the increase of iterations.}
\label{curve}
\end{center}
\vskip -0.2in
\end{figure*}

\section{Experiments}\label{exp}

\textbf{Experimental Setup.}
We conduct experiments on widely adopted benchmark datasets, including Cora, Citeseer, and Pubmed \cite{DBLP:journals/aim/SenNBGGE08, DBLP:conf/icml/YangCS16}. The accuracy and loss gap (\emph{i.e.}, the absolute value of difference between the loss (accuracy) on training and test samples) are used to estimate the generalization gap. Following the standard transductive learning setting, in each run, $30\%$ sampled nodes determined by a random seed are used as training set and the rest nodes are treated as test set. The number of iterations is fixed to $T=300$. We independently repeat the experiments for $10$ times and report the mean value and standard deviations of all runs. Please see Appendix C for more detailed settings.

\begin{table}[t]
\caption{Accuracy gap comparison of different baseline models on Cora, Citeseer and Pubmed.}
\label{acc_gap}
\vskip 0.15in
\begin{center}
\begin{small}
\begin{tabular}{lcccr}
\toprule
 & Cora & Citeseer & Pubmed \\
\midrule
GCN    & 9.76$\pm$1.15 & 22.11$\pm$1.26 & 1.08$\pm$0.52 \\
GCN*   & 13.45$\pm$1.28 & 26.48$\pm$1.21 & 1.49$\pm$0.63 \\
GAT   & 11.00$\pm$0.75 & 22.69$\pm$0.84 & 1.52$\pm$0.43 \\
GCNII & 7.69$\pm$1.48 & 14.85$\pm$0.80 & 0.88$\pm$0.52  \\
GCNII* & 6.24$\pm$1.59 & 13.49$\pm$1.39 & 0.80$\pm$0.50 \\
SGC    & 5.33$\pm$1.58 & 11.50$\pm$1.09 & 0.73$\pm$0.54 \\
APPNP    & 7.72$\pm$1.54 & 9.99$\pm$1.17 & 0.85$\pm$0.46  \\
GPR-GNN     & 8.90$\pm$1.22 & 19.08$\pm$0.95 & 0.96$\pm$0.49 \\
\bottomrule
\end{tabular}
\end{small}
\end{center}
\vskip -0.1in
\end{table}

\begin{table}[t]
\caption{Test accuracy comparison of different baseline models on Cora, Citeseer and Pubmed.}
\label{test_acc_gap}
\vskip 0.15in
\begin{center}
\begin{small}
\begin{tabular}{lcccr}
\toprule
 & Cora & Citeseer & Pubmed \\
\midrule
GCN    & 85.91$\pm$0.53 & 71.78$\pm$0.72 & 85.29$\pm$0.19 \\
GCN*   & 82.49$\pm$0.59 & 66.74$\pm$1.07 & 84.21$\pm$0.26 \\
GAT   & 86.10$\pm$0.51 & 72.90$\pm$0.65 & 85.45$\pm$0.26 \\
GCNII & 82.85$\pm$2.17 & 73.61$\pm$0.64 & 84.70$\pm$0.24  \\
GCNII* & 82.85$\pm$2.44 & 72.89$\pm$0.96 & 83.67$\pm$0.46 \\
SGC    & 82.39$\pm$2.48 & 74.37$\pm$0.56 & 82.00$\pm$0.27 \\
APPNP    & 79.14$\pm$3.17 & 74.12$\pm$0.62 & 82.86$\pm$0.29  \\
GPR-GNN     & 87.24$\pm$0.71 & 73.79$\pm$0.67 & 85.07$\pm$0.34 \\
\bottomrule
\end{tabular}
\end{small}
\end{center}
\vskip -0.1in
\end{table}

\textbf{Experimental Results.} The loss and accuracy comparisons are presented in Table~\ref{loss_gap} and Table~\ref{acc_gap}, respectively. We have the following observations: 
(1) SGC and APPNP have smaller loss and accuracy gap than other model including GCN, which is consistent with the analysis in Proposition~\ref{sgc}. Besides, the test accuracy of SGC surpass GCN on Citeseer. Thus, the reason why linear models sometimes perform is due to their smaller lipschitz continuity constants. 
(2) Compared with GCN, GCNII achieves smaller loss and accuracy gap with the same number of layers. We further estimate the generalization performance of GCN and GCNII with six layers (denoted as GCN* and GCNII*). Interestingly, with the increase of the number of hidden layers, the generalization performance of GCN decreases sharply. On the contrary, the loss and accuracy gap of GCNII remain unchanged. The test accuracy of GCNII also remain unchanged or only drops slightly. Therefore, the superior performance of GCNII comes from two perspectives: the first is learning non-degenerated representations by relieving over-smoothing and the second is robust generalization gap against the increase of the number of layers.
(3) Although GPR-GNN achieve a competitive test accuracy, it has higher accuracy and loss gap than APPNP. Therefore, the unconstrained learning coefficients improve the fitting ability but also weaken generalization capability. Designing weight learning schema to balance the expressive and generalization could be a direction for spectral-based GNNs.
(4) The generalization performance of GAT is slightly worse than GCN. Note that GAT is designed for inductive learning while our experimental setting is transducive. Thus, the superiority of GAT is not so obvious. 

Besides, loss value of GCN on training and test samples w.r.t. iterations are presented in Figure \ref{curve}. It can be seen that the loss gap increases with the increase of iterations, as demonstrated by Theorem~\ref{th1}. In general, the theoretical results are supported by the experimental results. It is worth pointing out that our analysis is only oriented to generalization gap. Smaller generalization gap does not necessarily mean better generalization ability, since the performance on test samples are determined by both training error and generalization gap. 

\begin{table}[t]
\caption{Loss gap comparison of different baseline models on Cora, Citeseer and Pubmed.}
\label{loss_gap}
\vskip 0.15in
\begin{center}
\begin{small}
\begin{tabular}{lcccr}
\toprule
 & Cora & Citeseer & Pubmed \\
\midrule
GCN    & 0.30$\pm$0.03 & 0.77$\pm$0.04 & 0.03$\pm$0.01 \\
GCN* & 0.91$\pm$0.18 & 2.12$\pm$0.16 & 0.05$\pm$0.01 \\
GAT   & 0.29$\pm$0.03 & 0.65$\pm$0.02 & 0.03$\pm$0.01 \\
GCNII & 0.19$\pm$0.03 & 0.43$\pm$0.02 & 0.02$\pm$0.01  \\
GCNII* & 0.16$\pm$0.03 & 0.43$\pm$0.03 &  0.02$\pm$0.01 \\
SGC    & 0.12$\pm$0.03 & 0.28$\pm$0.02 & 0.01$\pm$0.00 \\
APPNP    & 0.16$\pm$0.03 & 0.25$\pm$0.02 & 0.01$\pm$0.00  \\
GPR-GNN     & 0.24$\pm$0.03 & 0.55$\pm$0.02 & 0.02$\pm$0.00 \\
\bottomrule
\end{tabular}
\end{small}
\end{center}
\vskip -0.1in
\end{table}

\section{Discussion and Conclusion}
In this paper, we establish high probability learning guarantees for transductive SGD, by which the upper bound of generalization gap for some popular GNNs are derived. Experimental results on benchmark datasets support the theoretical results. This work sheds light on understanding the generalization of GNNs and provide some insights in designing new GNN architecture with both expressiveness and generalization capabilities. 

Although we have made efforts in generalization theory of GNNs, there are still some limitations in our analysis, which is left for future work to address: (1) The complexity based technique makes the dimension of parameters appearing in the bounds. Further research should focus on establishing dimension-independent bounds under milder assumption and deriving the lower bound that matches the upper bound. (2) We only analyze vanilla SGD in terms of optimization algorithms. Extending our results to SGD with momentum and adaptive learning rates is worth exploring. (3) Our analysis does not explicitly consider the heterophily of graphs. Deriving heterophily-dependent generalization bounds is an meaningful direction.

\bibliography{main}
\bibliographystyle{icml2023}

\newpage
\appendix
\onecolumn

\section{Notations and Lemmas}\label{defs}
In this section, we will present some notations, definitions and lemmas that will be used in subsequent analysis
Let $f: \mathbb{R}^{m\times n} \mapsto \mathbb{R}$ be a real-value function with variable $\mathbf{W} \in \mathbb{R}^{m\times n}$. We stipulate that 
\begin{equation*}
    \nabla_{{\rm vec} \left[\mathbf{W}\right]} f = \left[ \frac{\partial f}{\partial W_{11}}, \ldots, \frac{\partial f}{\partial W_{m1}}, \ldots, \frac{\partial f}{\partial W_{1n}}, \ldots, \frac{\partial f}{\partial W_{mn}} \right]^{\top} \in \mathbb{R}^{mn\times 1}. 
\end{equation*}
Denote by $\frac{\partial f}{\partial {\rm vec}\left[\mathbf{W}\right]}$ the Jacobian matrix, we have $\frac{\partial f}{\partial {\rm vec}\left[\mathbf{W}\right]} = \nabla^\top_{{\rm vec} \left[\mathbf{W}\right]} f \in \mathbb{R}^{1 \times mn}$. Denote by $\mathbf{W}_{i*}$ the $i$-th row of matrix $\mathbf{W}$. We use $\odot$ and $\otimes$ to denote Hadamard product and Kronecker product, respectively. The activation function $\sigma(\cdot)$ in this work is defined as
\begin{equation*}
    \sigma(x) = \begin{cases}
        0, x \leq 0, \\
        x^q, 0 < x \leq \left( \frac{1}{q} \right)^\frac{1}{q-1}, \\
        x - \left( \frac{1}{q} \right)^\frac{1}{q-1} + \left( \frac{1}{q} \right)^\frac{q}{q-1}, x > \left( \frac{1}{q} \right)^\frac{1}{q-1},
    \end{cases}
\end{equation*}
where $q \in (1, 2]$. It can be verify that this activation is differential on $\mathbb{R}$, and its derivation is
\begin{equation*}
    \sigma'(x) = \begin{cases}
        0, x \leq 0, \\
        qx^{q-1}, 0 < x \leq \left( \frac{1}{q} \right)^\frac{1}{q-1}, \\
        1, x > \left( \frac{1}{q} \right)^\frac{1}{q-1}.
    \end{cases}
\end{equation*}
When setting $p \approx 1$ (\emph{e.g.}, $q=1.1$), this activation function has tolerate approximation error to vanilla ReLU function. Now we show some property of $\sigma(\cdot)$ that used in the sequential proofs.
\begin{itemize}
    \item \emph{$\Vert \sigma(\mathbf{u}) \Vert_2 < \Vert \mathbf{u} \Vert_2 $ for any $\mathbf{u} \in \mathbb{R}^d$.} We only need to show that $\vert \sigma(u_i) \vert \leq \vert u_i \vert$ holds for $i \in [d]$. The case that $u_i \in (-\infty, 0]$ is trivial. If $u_i \in (0, \left( 1/q \right)^{1/(q-1)}]$, since $q>1$ and $\left( 1/q \right)^{1/(q-1)} < 1$, we have $\vert \sigma(u_i) \vert = u^q_i \leq u_i = \vert u_i \vert$. If $u_i \in (\left( 1/q \right)^{1/(q-1)}, \infty)$, note that $\left( 1/q \right)^{q/(q-1)} < \left( 1/q \right)^{1/(q-1)}$, we have $\vert \sigma(u_i) \vert \leq  u_i = \vert u_i \vert$.
    \item \emph{$\Vert \sigma'(\mathbf{u}) \odot \mathbf{v} \Vert_2 \leq \Vert \mathbf{v} \Vert_2 $ for any $\mathbf{u}, \mathbf{v} \in \mathbb{R}^d$.} By the formulation of $\sigma'(x)$, we have $\vert \sigma'(x) \vert \leq 1$. Then 
    \begin{equation*}
        \Vert \sigma'(\mathbf{u}) \odot \mathbf{v} \Vert_2 = \sqrt{\sum_{i=1}^d \vert \sigma'(u_i) \vert^2 \vert v_i \vert^2} \leq \sqrt{\sum_{i=1}^d \vert v_i \vert^2} = \Vert \mathbf{v} \Vert_2.
    \end{equation*}
    \item \emph{$\Vert \sigma'(\mathbf{u}) -  \sigma'(\mathbf{v}) \Vert_2 \leq q d^{\frac{2-q}{2}} \Vert \mathbf{u} - \mathbf{v} \Vert^{q-1}_2 $ for any $\mathbf{u}, \mathbf{v} \in \mathbb{R}^d$.} We first show that for any $x, y \in \mathbb{R}$, $\vert \sigma'(x) - \sigma'(y) \vert \leq q \vert x - y \vert^{q-1}$ holds. The case that $x, y \in (-\infty, 0]$ and $x, y \in [\left( 1/q \right)^{1/(q-1)}, \infty)$ are trivial. If $x, y \in (0, \left( 1/q \right)^{1/(q-1)}]$, we have $\vert qx^{q-1} - qy^{q-1} \vert \leq q \vert x - y \vert^{q-1} $. If $x \in (-\infty, 0]$ and $y \in (0, \left( 1/q \right)^{1/(q-1)}]$, we have $\vert qy^{q-1} \vert = q y^{q-1} \leq q \vert x - y \vert^{q-1} $. If $x \in (0, \left( 1/q \right)^{1/(q-1)}]$ and $y \in (\left( 1/q \right)^{1/(q-1)}, \infty)$, we have $\vert qx^{q-1} - 1 \vert \leq \vert qx^{q-1} - qy^{q-1} \vert \leq q \vert x - y \vert^{q-1}$. If $x \in (-\infty, 0]$ and $ y \in (\left( 1/q \right)^{1/(q-1)}, \infty)$, we have $\vert \sigma'(x) - \sigma'(y) \vert \leq q y^{q-1} \leq q \vert x - y \vert^{q-1} $. Thus,
    \begin{equation*}
    \begin{aligned}
        \Vert \sigma'(\mathbf{u}) - \sigma'(\mathbf{v}) \Vert_2 = & \sqrt{\sum_{i=1}^d \vert \sigma'(u_i) - \sigma'(v_i) \vert^2} \leq \sqrt{\sum_{i=1}^d q^2 \vert u_i - v_i \vert^{2(q-1)}}\\
        \leq & \sqrt{ \left( \sum_{i=1}^d \vert u_i - v_i \vert^{2} \right)^{q-1} \left( \sum_{i=1}^d q^{\frac{2}{2-q}} \right)^{2-q} } \leq q d^{\frac{2-q}{2}} \Vert \mathbf{u} - \mathbf{v} \Vert^{q-1}_{2}. \\
    \end{aligned}
    \end{equation*}
\end{itemize}

With the above notations, we give the following lemmas.
\begin{lemma}\label{infty_norm}
Denote by $\tilde{\mathbf{A}}$ the  normalized
adjacency matrix with self-loop, we have $\big\Vert \tilde{\mathbf{A}} \big\Vert_\infty \leq \sqrt{\frac{{\rm deg}_{\rm max}+1}{{\rm deg}_{\rm min}+1}}$.
\end{lemma}
\emph{Proof.} By definition, $\tilde{\mathbf{A}}_{ij} \geq 0$ holds for any $i, j \in [n]$. For any fixed $i \in [n]$, let $\mathcal{N}_i$ be the index set of the $i$-th nodes' one-hop neighbors, we have
\begin{equation*}
\begin{aligned}
    \sum_{j=1}^n \tilde{\mathbf{A}}_{ij} = & \sum_{j=1}^n \frac{\mathbf{A}_{ij}}{\sqrt{{\rm deg}_i+1}\sqrt{{\rm deg}_j+1}} \\
        = & \frac{1}{\sqrt{{\rm deg}_i+1}} \left( \frac{1}{\sqrt{{\rm deg}_i+1}} + \sum_{j\in \mathcal{N}_i} \frac{1}{\sqrt{{\rm deg}_j+1}} \right) \\
        \leq & \frac{1}{\sqrt{{\rm deg}_i+1}} \left( \frac{1}{\sqrt{{\rm deg}_{\rm min}+1}} + \sum_{j\in \mathcal{N}_i} \frac{1}{\sqrt{{\rm deg}_{\rm min}+1}} \right) \\
        \leq & \frac{1}{\sqrt{{\rm deg}_i+1}} \frac{{\rm deg}_i+1}{\sqrt{{\rm deg}_{\rm min}+1}} = \frac{\sqrt{{\rm deg}_i+1}}{\sqrt{{\rm deg}_{\rm min}+1}} \leq \sqrt{\frac{{\rm deg}_{\rm max}+1}{{\rm deg}_{\rm min}+1}}.
\end{aligned}
\end{equation*}

\begin{lemma}
Denote by $\mathbf{u} \in \mathbb{R}^m, \mathbf{v} \in \mathbb{R}^n$, we have $\Vert \mathbf{u} \otimes \mathbf{v} \Vert_2 = \Vert \mathbf{u} \Vert_2 \Vert \mathbf{v} \Vert_2$. 
\end{lemma}
\emph{Proof.} One can find that
\begin{equation*}
    \Vert \mathbf{u} \otimes \mathbf{v} \Vert_2 = \sqrt{\sum_{j=1}^m \Vert u_j \mathbf{v} \Vert^2_2} = \sqrt{\sum_{j=1}^m u^2_j \Vert \mathbf{v} \Vert^2_2} = \sqrt{\Vert \mathbf{u} \Vert^2_2 \Vert \mathbf{v} \Vert^2_2 } = \Vert \mathbf{u} \Vert_2 \Vert \mathbf{v} \Vert_2.
\end{equation*}
\begin{lemma}
Denote by $\mathbf{W}_1, \mathbf{W}_2 \in \mathbb{R}^{m\times n}$, we have
\begin{equation*}
    \Vert \mathbf{W}_1 - \mathbf{W}_2 \Vert \leq \Vert {\rm vec}\left[ \mathbf{W}_1 \right] - {\rm vec} \left[\mathbf{W}_2\right] \Vert_2.
\end{equation*}
\end{lemma}
\emph{Proof.} We have
\begin{equation*}\small
\begin{aligned}
    & \Vert \mathbf{W}_1 - \mathbf{W}_2 \Vert = \mathop{\rm sup}_{\Vert \mathbf{u} \Vert_2=1} \Vert (\mathbf{W}_1 - \mathbf{W}_2)\mathbf{u} \Vert_2 \triangleq \Vert (\mathbf{W}_1 - \mathbf{W}_2)\mathbf{u}^* \Vert_2 \\
    = & \sqrt{\sum_{j=1}^m ({[\mathbf{W}_1]}_{j,:}\mathbf{u}^* - {[\mathbf{W}_2]}_{j,:}\mathbf{u}^* )^2} \leq \sqrt{\sum_{j=1}^m \Vert {[\mathbf{W}_1]}_{j,:} - {[\mathbf{W}_2]}_{j,:} \Vert^2_2} = \Vert {\rm vec}(\mathbf{W}_1) - {\rm vec}(\mathbf{W}_2) \Vert_2,
\end{aligned}
\end{equation*}
where the last inequality follows from the Cauchy-Schwarz inequality and $\Vert \mathbf{u}^* \Vert_2 = 1$. This finishes the proof. 

\begin{lemma}\label{bound_alpha}
Denote by $\{\mathbf{W}_h \}_{h=1}^H$ the learnable parameters (w.l.o.g. we assume that each parameter is matrix since vector is a special case of matrix). If for $h \in [H]$,
\begin{equation*}
    \vert \ell(\mathbf{W}_1, \ldots, \mathbf{W}_h, \ldots, \mathbf{W}_H) - \ell(\mathbf{W}_1, \ldots, \mathbf{W}'_h, \ldots, \mathbf{W}_H) \vert \leq L_h \left\Vert {\rm vec} \left[ \mathbf{W}_h \right] - {\rm vec} \left[ \mathbf{W}'_h \right] \right\Vert_2,
\end{equation*}
and
\begin{equation*}
    \left\Vert \frac{\partial \ell(\mathbf{W}_1, \ldots, \mathbf{W}_H)}{\partial {\rm vec}\left[\mathbf{W}_h\right]} - \frac{\partial \ell(\mathbf{W}'_1, \ldots, \mathbf{W}'_H)}{\partial {\rm vec}\left[\mathbf{W}_h\right]} \right\Vert_2 \leq \sum_{i = 1}^H \left[ P_{hi} \left\Vert {\rm vec} \big[ \mathbf{W}_i \big] - {\rm vec}\big[ \mathbf{W}'_i \big] \right\Vert_2 + \tilde{P}_{hi} \left\Vert {\rm vec} \big[ \mathbf{W}_i \big] - {\rm vec}\big[ \mathbf{W}'_i \big] \right\Vert^{\tilde{\alpha}}_2 \right],
\end{equation*}
then there exist $P, L > 0$ such that $\left\vert \ell(\mathbf{w}) - \ell(\mathbf{w}') \right\vert \leq L \left\Vert \mathbf{w} - \mathbf{w}' \right\Vert_2$ and $\Vert \nabla \ell(\mathbf{w}) - \ell(\mathbf{w}) \Vert_2 \leq P_{\mathcal{F}} \mathop{\rm max} \{ \Vert \mathbf{w} - \mathbf{w} \Vert_2, \Vert \mathbf{w} - \mathbf{w} \Vert^{\tilde{\alpha}}_2 \}$ hold, where $\mathbf{w} = [{\rm vec}\left[ \mathbf{W}_1 \right]; \ldots; {\rm vec}\left[ \mathbf{W}_H \right]]$.
\end{lemma}
\emph{Proof.} By definition, the gradient of $\ell$ w.r.t $\mathbf{w}$ is $\nabla \ell (\mathbf{w}) = \left[ \frac{\partial \ell}{\partial {\rm vec}\left[\mathbf{W}_1 \right]}, \ldots, \frac{\partial \ell}{\partial {\rm vec}\left[\mathbf{W}_H \right]} \right]^\top$. Then we have
\begin{equation*}
\begin{aligned}
    \left\vert \ell(\mathbf{w}) - \ell(\mathbf{w}') \right\vert
    \leq & \sum_{h=1}^H \vert \ell(\mathbf{W}_1, \ldots, \mathbf{W}_h, \ldots, \mathbf{W}_H) - \ell(\mathbf{W}_1, \ldots, \mathbf{W}'_h, \ldots, \mathbf{W}_H) \vert \\
    \leq & \sum_{h=1}^H L_h \left\Vert {\rm vec} \big[ \mathbf{W}_h \big] - {\rm vec}\big[ \tilde{\mathbf{W}}_h \big] \right\Vert_2 \\
    \leq & \left( \sum_{h=1}^H L^2_h \right)^{\frac{1}{2}} \left( \sum_{h=1}^H \left\Vert {\rm vec} \big[ \mathbf{W}_h \big] - {\rm vec}\big[ \mathbf{W}'_h \big] \right\Vert^2_2 \right)^{\frac{1}{2}} \\
    = & L \left\Vert \mathbf{w} - \tilde{\mathbf{w}} \right\Vert_2,
\end{aligned}
\end{equation*}
where we obtain the last inequality by Cauchy-Schwarz inequality. Similarly, we have
\begin{equation}\label{gradient_l}
\begin{aligned}
    & \left\Vert \nabla \ell (\mathbf{w}) - \nabla \ell (\mathbf{w}') \right\Vert_2 \\
    \leq & \left\Vert \frac{\partial \ell(\mathbf{W}_1, \ldots, \mathbf{W}_H)}{\partial {\rm vec}\left[\mathbf{W}_h\right]} - \frac{\partial \ell(\mathbf{W}'_1, \ldots, \mathbf{W}'_H)}{\partial {\rm vec}\left[\mathbf{W}_h\right]} \right\Vert_2 \\
    \leq & \sum_{h=1}^H \left[ \sum_{i = 1}^H P_{hi} \left\Vert {\rm vec} \big[ \mathbf{W}_i \big] - {\rm vec}\big[ \mathbf{W}'_i \big] \right\Vert_2 \right] + \sum_{h=1}^H \left[ \sum_{i = 1}^H \tilde{P}_{hi} \left\Vert {\rm vec} \big[ \mathbf{W}_i \big] - {\rm vec}\big[ \mathbf{W}'_i \big] \right\Vert^{\tilde{\alpha}}_2 \right] \\
    = & \sum_{i=1}^H \left[ \sum_{h = 1}^H P_{hi} \left\Vert {\rm vec} \big[ \mathbf{W}_i \big] - {\rm vec}\big[ \mathbf{W}'_i \big] \right\Vert_2 \right] + \sum_{i=1}^H \left[ \sum_{h = 1}^H \tilde{P}_{hi} \left\Vert {\rm vec} \big[ \mathbf{W}_i \big] - {\rm vec}\big[ \mathbf{W}'_i \big] \right\Vert^{\tilde{\alpha}}_2 \right] \\
    = & \sum_{i=1}^H P_{i} \left\Vert {\rm vec} \big[ \mathbf{W}_i \big] - {\rm vec}\big[ \mathbf{W}'_i \big] \right\Vert_2 + \sum_{i=1}^H \tilde{P}_{i} \left\Vert {\rm vec} \big[ \mathbf{W}_i \big] - {\rm vec}\big[ \mathbf{W}'_i \big] \right\Vert^{\tilde{\alpha}}_2 \\
    \leq & \left( \sum_{i=1}^H P^2_i \right)^{\frac{1}{2}} \left( \sum_{i=1}^H \left\Vert {\rm vec} \big[ \mathbf{W}_i \big] - {\rm vec}\big[ \mathbf{W}'_i \big] \right\Vert^2_2 \right)^{\frac{1}{2}} + \left( \sum_{i=1}^H P^{\frac{2}{2-\tilde{\alpha}}}_i \right)^{1-\frac{\tilde{\alpha}}{2}} \left( \sum_{i=1}^H \left\Vert {\rm vec} \big[ \mathbf{W}_i \big] - {\rm vec}\big[ \mathbf{W}'_i \big] \right\Vert^2_2 \right)^{\frac{\tilde{\alpha}}{2}} \\
    = & \left( \sum_{i=1}^H P^2_i \right)^{\frac{1}{2}} \left\Vert \mathbf{w} - \mathbf{w}' \right\Vert_2 + \left( \sum_{i=1}^H P^{\frac{2}{2-\tilde{\alpha}}}_i \right)^{1-\frac{\tilde{\alpha}}{2}} \left\Vert \mathbf{w} - \mathbf{w}' \right\Vert^{\tilde{\alpha}}_2 \\
    \leq & \mathop{\rm max} \left\{ \left( \sum_{i=1}^H P^2_i \right)^{\frac{1}{2}} + \left( \sum_{i=1}^H P^{\frac{2}{2-\tilde{\alpha}}}_i \right)^{1-\frac{\tilde{\alpha}}{2}} \right\} \mathop{\rm max} \left\{ \left\Vert \mathbf{w} - \mathbf{w}' \right\Vert_2, \left\Vert \mathbf{w} - \mathbf{w}' \right\Vert^{\tilde{\alpha}}_2 \right\} \\
    = & P \mathop{\rm max} \left\{ \left\Vert \mathbf{w} - \mathbf{w}' \right\Vert_2, \left\Vert \mathbf{w} - \mathbf{w}' \right\Vert^{\tilde{\alpha}}_2 \right\}.
\end{aligned}
\end{equation}
where we define $P_i = \sum_{h=1}^H P_{hi}$ and $\tilde{P}_i = \sum_{h=1}^H \tilde{P}_{hi}$. The second inequality is due to the H\"older inequality.

\begin{lemma}\label{lipschitz}
Denote by $\mathbf{v} \in \mathbb{R}^d$. Let $f: \mathbb{R}^d \mapsto \mathbb{R}^d $ be $f(\mathbf{v})_j = \frac{e^{v_j}}{\sum_{i=1}^d e^{v_i}}$. For any $\mathbf{v}, \mathbf{v}' \in \mathbb{R}^d$, we have $\Vert f(\mathbf{v}) - f(\mathbf{v}') \Vert_2 \leq 2 \Vert \mathbf{v} - \mathbf{v}' \Vert_2$.
\end{lemma}
\emph{Proof.} By \cite{federer1969geometric}, we have $\Vert f(\mathbf{v}) - f(\mathbf{v}') \Vert_2 \leq \mathop{\rm sup}_{\mathbf{v} \in \mathbb{R}^d} \Vert J(\mathbf{v}) \Vert \Vert \mathbf{v} - \mathbf{v}' \Vert_2$, where $J$ is the Jacobian. For the aforementioned $f$, we have $J(\mathbf{v}) = {\rm diag} \left( f(\mathbf{v}) \right) - f(\mathbf{v}) f(\mathbf{v})^\top$. Then
\begin{equation}\label{lipschitz_1}
    \Vert J(\mathbf{v}) \Vert = \Vert {\rm diag} \left( f(\mathbf{v}) \right) - f(\mathbf{v}) f(\mathbf{v})^\top \Vert \leq \Vert {\rm diag} \left( f(\mathbf{v}) \right) \Vert + \Vert f(\mathbf{v}) f(\mathbf{v})^\top \Vert.
\end{equation}
First, 
\begin{equation}\label{lipschitz_2}
    \Vert {\rm diag} \left( f(\mathbf{v}) \right) \Vert = \mathop{\rm sup}_{\Vert \mathbf{w} \Vert_2 = 1} \Vert {\rm diag} \left( f(\mathbf{v}) \right) \mathbf{w} \Vert_2 = \mathop{\rm sup}_{\Vert \mathbf{w} \Vert_2 = 1} \sqrt{ \sum_{i=1}^d f^2(\mathbf{v})_i w^2_i } = \mathop{\rm max}_{i \in [d]} f(\mathbf{v})_i \leq 1.
\end{equation}
Besides,
\begin{equation}\label{lipschitz_3}
    \Vert f(\mathbf{v}) f(\mathbf{v})^\top \Vert = \mathop{\rm sup}_{\Vert \mathbf{w} \Vert_2 = 1} \Vert f(\mathbf{v}) f(\mathbf{v})^\top \mathbf{w} \Vert_2 = \Vert f(\mathbf{v}) \Vert_2  \mathop{\rm sup}_{\Vert \mathbf{w} \Vert_2 = 1} | f(\mathbf{v})^\top \mathbf{w}| \leq \Vert f(\mathbf{v}) \Vert^2_2 \leq 1, 
\end{equation}
where the last inequality is due to $\sum_{i=1}^d f(\mathbf{v})_i = 1$.
Plugging Eq.~(\ref{lipschitz_2}) and Eq.~(\ref{lipschitz_3}) into Eq.~(\ref{lipschitz_1}), the proof is completed.

\begin{lemma}\label{cross_entropy}
Denote by $\mathbf{v} \in \mathbb{R}^d$. Let $f: \mathbb{R}^d \mapsto \mathbb{R} $ be $f(\mathbf{v})_j = - \sum_{k=1}^K y_k \log \hat{y}_k $, where $\hat{y}_k = \frac{e^{v_k}}{\sum_{i=1}^K e^{v_i}}$. For any $\mathbf{v}, \mathbf{v}' \in \mathbb{R}^d$, we have $\vert f(\mathbf{v}) - f(\mathbf{v}') \vert \leq \sqrt{2} \Vert \mathbf{v} - \mathbf{v}' \Vert_2$.
\end{lemma}
\emph{Proof.} By the chain rule, the Jacobian is $J(\mathbf{v}) = \hat{\mathbf{y}}(\mathbf{v}) - \mathbf{y}$. Note that $\vert f(\mathbf{v}) - f(\mathbf{v}') \vert \leq \mathop{\rm sup}_{\mathbf{v} \in \mathbb{R}^d} \Vert J(\mathbf{v}) \Vert_2 \Vert \mathbf{v} - \mathbf{v}' \Vert_2$. W.o.l.g, we assume that $y_1=1$, then
\begin{equation*}
    \Vert J(\mathbf{v}) \Vert_2 = \sqrt{\sum_{k=1}^K (\hat{y}_k - y_k)^2} = \sqrt{ (1-\hat{y}_1)^2 + \sum_{k=2}^K \hat{y}^2_k} \leq \sqrt{ 1 + \sum_{k=1}^K \hat{y}^2_k} \leq \sqrt{2},
\end{equation*}
where the last inequality is due to $\sum_{k=1}^K \hat{y}_k = 1$.

\begin{lemma}\label{chaos}
Let $\mathcal{F}:\mathcal{Z} \times \mathcal{Z} \mapsto \mathbb{R}$ be a function class with $\mathop{\rm sup}_{f \in \mathcal{F}} d_S(f, 0) \leq D$ and $S = \{z_1, \ldots, z_n \} \subset \mathcal{Z}$, where $d_S$ is a empirical metric defind on $\mathcal{F}$:
\begin{equation*}
    d_S(f, g) = \left( \frac{1}{n^2} \sum_{1 \leq i < j \leq n} \vert f(x_i, x_j) - g(x_i, x_j) \vert^2 \right)^{\frac{1}{2}}.
\end{equation*}
We have
\begin{equation*}
    \mathcal{U}(\mathcal{F}) \triangleq \frac{1}{n} \mathbb{E}_{\bm \sigma} \left[ \mathop{\rm sup}_{f \in \mathcal{F}} \sum_{1\leq i < j \leq n} \sigma_i \sigma_j f(z_i, z_j) \right] \leq 24e \int^D_{0} \log (\mathcal{N}(r, \mathcal{F}, d_S) + 1){\,\mathrm{d}}r,
\end{equation*}
where $\bm \sigma$ is the transductive Rademacher variable.
\end{lemma}
\emph{Proof.} The proof extend Theorem~2 in \cite{ying} to the transductive Rademancher chaos complexity. The first step is to show that the following inequality holds for $1 < p \leq q < \infty$ and $d \geq 1$:
\begin{equation}\label{khin}
\begin{aligned}
    & \left[ \mathbb{E} \Big\Vert x + \gamma \sum_{i=1}^n x_i \sigma_i + \gamma^2 \sum_{i_1 < i_2 \leq n} x_{i_1 i_2} \sigma_{i_1} \sigma_{i_2} + \cdots + \gamma^d \sum_{i_1 < \cdots < i_d \leq n} x_{i\ldots i_d} \sigma_{i_1} \cdots \sigma_{i_d} \Big\Vert_2^q \right]^{\frac{1}{q}} \\
    \leq & \left[ \mathbb{E} \Big\Vert x + \sum_{i=1}^n x_i \epsilon_i + \sum_{i_1 < i_2 \leq n} x_{i_1 i_2} \epsilon_{i_1} \epsilon_{i_2} + \cdots + \sum_{i_1 < \cdots < i_d \leq n} x_{i\cdots i_d} \epsilon_{i_1} \cdots \epsilon_{i_d} \Big\Vert_2^p \right]^{\frac{1}{p}},
\end{aligned}
\end{equation}
where $\sigma$ and $\epsilon$ are transductive and standard Rademacher variable, respectively. The process generally follows that of Theorem 3.2.2 in \cite{Gin1998DecouplingFD}. First, consider the case that $n=1$, we have to show that $\left( \mathbb{E} {|x + \gamma \sigma y|}^q \right)^{\frac{1}{q}} \leq \left( \mathbb{E} {|x + \epsilon y|}^p \right)^{\frac{1}{p}}$ holds. This inequality naturally holds when $x=y=0$ or $y=0$. When $x=0$ and $y \neq 0$, we have
\begin{equation*}
    \left( \mathbb{E} {|x + \gamma \sigma y|}^q \right)^{\frac{1}{q}} \Big|_{x=0} = \left( 2 p_0 \vert \gamma y \vert^q \right)^{\frac{1}{q}} \leq \left( \vert \gamma y \vert^q \right)^{\frac{1}{q}} \leq \vert \gamma y \vert = \left( \mathbb{E} {|x + \gamma \epsilon y|}^q \right)^{\frac{1}{q}} \Big|_{x=0},
\end{equation*}
where the inequality is due to $p_0 \leq \frac{1}{2}$. When $x\neq 0$ and $y \neq 0$, let $u = \frac{y}{x}$, then
\begin{equation*}
    \left( \mathbb{E} {|x + \gamma \sigma y|}^q \right)^{\frac{1}{q}} \leq \left( \mathbb{E} {|x + \epsilon y|}^p \right)^{\frac{1}{p}} \Leftrightarrow \left( \mathbb{E} {|1+\gamma \sigma u|}^q \right)^{\frac{1}{q}} \leq \left( \mathbb{E} {|1+\gamma \sigma u|}^p \right)^{\frac{1}{p}}.
\end{equation*}
By symmetric, we only have to discuss the case that $u \geq 0$.
For $0 \leq u \leq 1$:
\begin{equation*}
\begin{aligned}
    \left( \mathbb{E} {|1+\gamma \sigma u|}^q \right)^{\frac{1}{q}} & = (p_0 {|1+ \gamma u|}^q + p_0 {|1- \gamma u|}^q + (1-2p_0) )^{\frac{1}{q}} \\
    & = \left[ p_0 + \sum_{k=1}^\infty p_0 \binom{q}{k} \gamma^k u^k + p_0 + \sum_{k=1}^\infty p_0 \binom{q}{k} (-1)^k \gamma^k u^k + (1-2p_0)\right]^{\frac{1}{q}} \\
    & = \left[ 2p_0 + 2p_0 \sum_{k=1}^\infty \binom{q}{2k} \gamma^{2k} {u}^{2k} + (1-2p_0)\right]^{\frac{p}{q}} \\
    & = \left[1 + 2p_0 \sum_{k=1}^\infty \binom{q}{2k} \gamma^{2k} {u}^{2k} \right]^{\frac{p}{q}} \leq \left[1 + \sum_{k=1}^\infty \binom{q}{2k} \gamma^{2k} {u}^{2k} \right]^{\frac{p}{q}} \\
    & = \left[ \frac{1}{2} \vert 1 + \gamma u \vert^q + \vert 1 - \gamma u \vert^q  \right]^{\frac{1}{q}} \leq \left( \mathbb{E} {|1+\epsilon u|}^q \right)^{\frac{1}{q}},
\end{aligned}
\end{equation*}
where the first inequality is due to $p_0 \leq \frac{1}{2}$, and the last inequality is from Eq.~(3.2.4') in \cite{Gin1998DecouplingFD}. For $ u \geq 1$, we have $\vert 1 \pm \gamma u \vert \leq \vert u \pm \gamma \vert$ since $u^2 (1 - \gamma^2 ) \geq 1 - \gamma^2$. Then we have
\begin{equation*}
\begin{aligned}
    & (p_0 {|1+ \gamma u|}^q + p_0 {|1- \gamma u|}^q + (1-2p_0) )^{\frac{1}{q}} \\
    \leq & \left( p_0 \vert u \vert^q \vert 1 + \gamma/u \vert + p_0 \vert u \vert^q \vert 1 - \gamma/u \vert + \vert u \vert^q (1-2p_0) \right)^{\frac{1}{q}} \\
    = & \vert u \vert \left( p_0 \vert 1 + \gamma/u \vert + p_0 \vert 1 - \gamma/u \vert + (1-2p_0) \right)^{\frac{1}{q}} \\
    \leq & \vert u \vert \left[ \frac{1}{2} \vert 1 + \gamma/u \vert^q + \vert 1 - \gamma/u \vert^q  \right]^{\frac{1}{q}} \leq \vert u \vert \left[ \frac{1}{2} \vert 1 + 1/u \vert^q + \vert 1 - 1/u \vert^q  \right]^{\frac{1}{q}} = \left( \mathbb{E} {|1+\epsilon u|}^q \right)^{\frac{1}{q}}.
\end{aligned}
\end{equation*}
where the last inequality is obtained by applying Eq.~(3.2.4') in \cite{Gin1998DecouplingFD} and replacing $u$ with $1/u$. Second, consider the case where $x, y$ are vectors. Let $z_1 = x + y$, $z_2 = x - y$, $u = \frac{\Vert z_1 \Vert + \Vert z_2 \Vert}{2}$ and $v = \frac{\Vert z_1 \Vert - \Vert z_2 \Vert}{2}$. In the second step, let $\kappa = \frac{v}{u}$, we have 
\begin{equation*}
\begin{aligned}
    \left( \mathbb{E} \Vert x + \gamma \sigma y \Vert^q \right)^{\frac{1}{q}} & = \left[ p_0 \Vert x + \gamma y \Vert^q + p_0 \Vert x - \gamma y \Vert^q + (1-2p_0) \Vert x \Vert^q \right]^{\frac{1}{q}} \\
    & \leq \left[ p_0 \left( \frac{1+\gamma}{2} \Vert z_1 \Vert + \frac{1-\gamma}{2} \Vert z_2 \Vert \right)^q + p_0 \left( \frac{1-\gamma}{2} \Vert z_1 \Vert + \frac{1+\gamma}{2} \Vert z_2 \Vert \right)^q + (1-2p_0) \Vert x \Vert^q \right]^{\frac{1}{q}} \\
    & = \left[ p_0 \left\vert u + \gamma v \right\vert^q + p_0 \left\vert u - \gamma v \right\vert^q + (1-2p_0) \Vert x \Vert^q \right]^{\frac{1}{q}} \\
    & \leq \left[ p_0 \left\vert u + \gamma v \right\vert^q + p_0 \left\vert u - \gamma v \right\vert^q + (1-2p_0) u^q \right]^{\frac{1}{q}} \\
    & = \vert u \vert \left[ p_0 \left\vert 1 + \gamma \kappa \right\vert^q + p_0 \left\vert 1 - \gamma \kappa \right\vert^q + (1-2p_0) \right]^{\frac{1}{q}} \\
    & \leq \vert u \vert \left[ \frac{\vert 1 + \kappa \vert^p + \vert 1 - \kappa \vert^p}{2} \right]^{\frac{1}{p}} = \left[ \frac{\vert u + v \vert^p + \vert u - v \vert^p}{2} \right]^{\frac{1}{p}},
\end{aligned}
\end{equation*}
where we use the result for the the case that $n=1$ to obtain the last inequality. The second inequality is due to
\begin{equation*}
    \Vert x \Vert = \left\Vert \frac{1}{2}(x + y) + \frac{1}{2}(x - y) \right\Vert \leq \frac{1}{2} \Vert z_1 \Vert + \frac{1}{2} \Vert z_2 \Vert = u.
\end{equation*}
Third, we use induction to obtain the final result. Following \cite{Gin1998DecouplingFD}, we only show $n=1$ implies $n=2$. Denote by $\mu$ the measure on the probability space, by Fubini theorem:
\begin{equation*}
\begin{aligned}
    & \left[ \mathbb{E}_{\sigma_1, \sigma_2} \Big\Vert x + \gamma x_1 \sigma_1 + \gamma x_2 \sigma_2 + \gamma^2 x_{12} \sigma_1 \sigma_2 \Big\Vert_2^q \right]^{\frac{1}{q}} \\
    = & \left[ \int \left( \int \Big\Vert x + \gamma x_1 \sigma_1 + \gamma x_2 \sigma_2 + \gamma^2 x_{12} \sigma_1 \sigma_2 \Big\Vert_2^q {\,\mathrm{d}}\mu(\sigma_2) \right) {\,\mathrm{d}} \mu(\sigma_1) \right]^{\frac{1}{q}} \\
    \leq & \left[ \int \left( \int \Big\Vert x + \gamma x_1 \sigma_1 + x_2 \epsilon_2 + \gamma x_{12} \sigma_1 \epsilon_2 \Big\Vert_2^p {\,\mathrm{d}}\mu(\epsilon_2) \right)^{\frac{q}{p}} {\,\mathrm{d}} \mu(\sigma_1) \right]^{\frac{1}{q}} \\
    \leq & \left[ \int \left( \int \Big\Vert x + \gamma x_1 \sigma_1 + x_2 \epsilon_2 + \gamma x_{12} \sigma_1 \epsilon_2 \Big\Vert_2^q {\,\mathrm{d}}\mu(\sigma_1) \right)^{\frac{p}{q}} {\,\mathrm{d}} \mu(\epsilon_2) \right]^{\frac{1}{p}} \\
    \leq & \left[ \int \int \Big\Vert x + \gamma x_1 \sigma_1 + x_2 \epsilon_2 + x_{12} \sigma_1 \epsilon_2 \Big\Vert_2^p {\,\mathrm{d}}\mu(\epsilon_1) {\,\mathrm{d}} \mu(\epsilon_2) \right]^{\frac{1}{p}} \\
    = & \left[ \mathbb{E}_{\epsilon_1, \epsilon_2} \Big\Vert x + x_1 \sigma_1 + x_2 \sigma_2 + x_{12} \sigma_1 \sigma_2 \Big\Vert_2^p \right]^{\frac{1}{p}},
\end{aligned}
\end{equation*}
where the first and the third inequality is due to the inducition hypothesis, and the second inequality is due to the Minkowski inequality. The remaining steps are the same as that in \cite{ying}, by applying Eq.~(\ref{khin}) to obtain Eq.~(21) in \cite{ying}. We omit the detail proof for concise.

\section{Proof of Main Results}
In this part, we present detailed proof of the results in the main body.
\subsection{Proof of Section~\ref{transductive}}
\subsubsection{Proof of Theorem~\ref{th1}}
\emph{Proof.} Following \cite{yaniv_2007}, let $p=\frac{mu}{(m+u)^2}$, the Transductive Rademacher complexity is defined as
\begin{equation*}
    \mathcal{R}_{m+u}(\mathbf{w}) = \left( \frac{1}{m}+\frac{1}{u} \right) \mathbb{E}_{\bm \sigma} \left[ \mathop{\rm sup}_{\mathbf{w} \in B_R} \sum_{i=1}^{m+u} {\sigma}_i \ell(\mathbf{w};z_i) \right],
\end{equation*}
where ${\sigma}_i$ is a random variable taking value in $\{\pm 1\}$ with probability $p$ and $0$ with probability $1-2p$.
By Theorem~1 in \cite{yaniv_2007}, with probability at least $1-\delta/2$,
\begin{equation}\label{general}
    R_u(\mathbf{w}^{(T+1)}) \leq R_m(\mathbf{w}^{(T+1)}) + \mathcal{R}_{m+u}(\mathbf{w}) + c_0Q\sqrt{{\rm min}(m,u)} + \sqrt{\frac{SQ}{2}\log \frac{2}{\delta}},
\end{equation}
where $Q \triangleq (\frac{1}{m}+\frac{1}{u})$ and $S \triangleq \frac{m+u}{(m+u-\frac{1}{2})(1-1/2({\rm max}(m,u)))}$. $c_0 \triangleq \sqrt{\frac{32\log (4e)}{3}}$ is a constant. Applying Lemma~1 in \cite{yaniv_2007} with $p_2=\frac{1}{2}$, we obtain
\begin{equation}\label{trc1}
    \mathcal{R}_{m+u}(\mathbf{w}) \leq \left( \frac{1}{m}+\frac{1}{u} \right) \mathbb{E}_{\bm \epsilon} \left[ \mathop{\rm sup}_{\mathbf{w} \in B_R} \sum_{i=1}^{m+u} {\epsilon}_i \ell(\mathbf{w};z_i) \right],
\end{equation}
where $\epsilon_i$ is the standard Rademacher random variable. Now we give an upper bound of the Transductive Rademacher Complexity by Dudley’s integral technique. Denote by $d_{\mathcal{H}_S}(\mathbf{w}, \widetilde{\mathbf{w}}) = \left( \frac{1}{m+u} \sum_{i=1}^{m+u} \left[ \ell(\mathbf{w}; z_i) - \ell(\widetilde{\mathbf{w}};z_i) \right]^2 \right)^{\frac{1}{2}}$. For $j \in \mathbb{N}$, let $\alpha_j = 2^{-j}M$ with $M = \mathop{\rm sup}_{\mathbf{w}\in B_R} d_{\mathcal{H}_S}(\mathbf{w}, \mathbf{w}^{(1)})$. Denote by $T_j$ the minimal $\alpha_j$-cover of $B_R$ and $\ell(\mathbf{w}^j;z)[\mathbf{w}]$ the element in $T_j$ that covers $\ell(\mathbf{w};z)$. Specifically, since $\{ \ell(\mathbf{w}^{(1)};z)\}$ is a $M$-cover of $B_R$, we set $\ell(\mathbf{w}^0;z)[\mathbf{w}] = \ell(\mathbf{w}^{(1)};z)$ (recall that $\mathbf{w}^{(1)}$ is the initialization parameter and $\mathbf{w}^j$ is the associated parameter of $\ell$ in $T_j$). For arbitrary $N \in \mathbb{N}$:
\begin{equation}\label{tr2}
\begin{aligned}
    & \mathbb{E}_{\bm \epsilon} \left[ \mathop{\rm sup}_{\mathbf{w} \in B_R} \sum_{i=1}^{m+u} {\epsilon}_i \ell(\mathbf{w};z_i) \right] \\
    = & \mathbb{E}_{\bm \epsilon} \left[ \mathop{\rm sup}_{\mathbf{w} \in B_R} \bigg( \sum_{i=1}^{m+u} \Big( {\epsilon}_i (\ell(\mathbf{w};z_i) - \ell(\mathbf{w}^N;z_i))[\mathbf{w}] + \sum_{j=1}^N {\epsilon}_i (\ell(\mathbf{w}^j;z_i)[\mathbf{w}] - \ell(\mathbf{w}^{j-1};z_i)[\mathbf{w}]) + {\epsilon}_i \ell(\mathbf{w}^{(1)};z_i) \Big) \bigg) \right] \\
    \leq & \mathbb{E}_{\bm \epsilon} \left[ \mathop{\rm sup}_{\mathbf{w} \in B_R} \bigg( \sum_{i=1}^{m+u} {\epsilon}_i (\ell(\mathbf{w};z_i) - \ell(\mathbf{w}^N;z_i)[\mathbf{w}]) \bigg) \right] + \sum_{j=1}^N  \mathbb{E}_{\bm \epsilon} \left[ \mathop{\rm sup}_{\mathbf{w} \in B_R} \bigg( \sum_{i=1}^{m+u} {\epsilon}_i (\ell(\mathbf{w}^j;z_i)[\mathbf{w}] - \ell(\mathbf{w}^{j-1};z_i)[\mathbf{w}]) \bigg) \right] \\
    & + \mathbb{E}_{\bm \epsilon} \left[ \sum_{i=1}^{m+u} {\epsilon}_i \ell(\mathbf{w}^{(1)};z_i) \right].
\end{aligned}
\end{equation}
For the first term, we apply Cauchy-Schwarz inequality and obtain
\begin{equation}\label{term1}
\begin{aligned}
    & \mathbb{E}_{\bm \epsilon} \left[ \mathop{\rm sup}_{\mathbf{w} \in B_R} \bigg( \sum_{i=1}^{m+u} {\epsilon}_i (\ell(\mathbf{w};z_i) - \ell(\mathbf{w}^N;z_i)[\mathbf{w}]) \bigg) \right] \\
    \leq & \left(\mathbb{E}_{\bm \epsilon} \left[ \sum_{i=1}^{m+u} \epsilon^2_i \right] \right)^{\frac{1}{2}} \left( \mathop{\rm sup}_{\mathbf{w} \in B_R} \sum_{i=1}^{m+u} (\ell(\mathbf{w};z_i) - \ell(\mathbf{w}^N;z_i)[\mathbf{w}])^2 \right)^{\frac{1}{2}} \leq (m+u)\alpha_N.
\end{aligned}
\end{equation}
By Massart's Lemma, we have
\begin{equation}\label{massart}
\begin{aligned}
    & \mathbb{E}_{\bm \epsilon} \left[ \mathop{\rm sup}_{\mathbf{w} \in B_R} \bigg( \sum_{i=1}^{m+u} {\epsilon}_i (\ell(\mathbf{w}^j;z_i)[\mathbf{w}] - \ell(\mathbf{w}^{j-1};z_i)[\mathbf{w}]) \bigg) \right] \leq \sqrt{m+u} \mathop{\rm sup}_{\mathbf{w} \in B_R} d_{\mathcal{H}_S} (\mathbf{w}^j, \mathbf{w}^{j-1}) \sqrt{2 \log \vert T_j \vert \vert T_{j-1} \vert}.
\end{aligned}
\end{equation}
By the Minkowski inequality,
\begin{equation}\label{mink}
\begin{aligned}
    & \mathop{\rm sup}_{\mathbf{w} \in B_R} d_{\mathcal{H}_S} (\mathbf{w}^j, \mathbf{w}^{j-1}) \\
    = & \mathop{\rm sup}_{\mathbf{w} \in B_R} \left( \frac{1}{m+u} \sum_{i=1}^{m+u} \left[ \ell(\mathbf{w}^j; z_i)[\mathbf{w}] - \ell(\mathbf{w};z) + \ell(\mathbf{w};z) - \ell(\mathbf{w}^{j-1};z_i)[\mathbf{w}] \right]^2 \right)^{\frac{1}{2}} \\
    \leq & \mathop{\rm sup}_{\mathbf{w} \in B_R} \left( \frac{1}{m+u} \sum_{i=1}^{m+u} \left[ \ell(\mathbf{w}^j; z_i)[\mathbf{w}] - \ell(\mathbf{w};z) \right]^2 \right)^{\frac{1}{2}} + \mathop{\rm sup}_{\mathbf{w} \in B_R} \left( \frac{1}{m+u} \sum_{i=1}^{m+u} \left[ \ell(\mathbf{w};z) - \ell(\mathbf{w}^{j-1};z_i)[\mathbf{w}] \right]^2 \right)^{\frac{1}{2}} \\
    = & \mathop{\rm sup}_{\mathbf{w} \in B_R} d_{\mathcal{H}_S} (\mathbf{w}^j, \mathbf{w}) + \mathop{\rm sup}_{\mathbf{w} \in B_R} d_{\mathcal{H}_S} (\mathbf{w}, \mathbf{w}^{j-1}) \leq \alpha_j + \alpha_{j-1} = 3\alpha_j.
\end{aligned}
\end{equation}
Plugging Eq.~(\ref{mink}) into Eq.~(\ref{massart}), using facts that $\alpha_j = 2(\alpha_j - \alpha_{j+1})$ and $|T_j| \geq |T_{j-1}|$, taking summation over $j$,
\begin{equation}\label{term2}
\begin{aligned}
    & \sum_{j=1}^N \mathbb{E}_{\bm \epsilon} \left[ \mathop{\rm sup}_{\mathbf{w} \in B_R} \bigg( \sum_{i=1}^{m+u} {\epsilon}_i (\ell(\mathbf{w}^j;z_i)[\mathbf{w}] - \ell(\mathbf{w}^{j-1};z_i)[\mathbf{w}]) \bigg) \right] \\
    \leq & 6\sqrt{m+u} \sum_{j=1}^N \alpha_j \sqrt{\log |T_j|} = 12\sqrt{m+u} \sum_{j=1}^N (\alpha_j - \alpha_{j+1}) \sqrt{\log |T_j|} \\
    = & 12\sqrt{m+u} \sum_{j=1}^N (\alpha_j - \alpha_{j+1}) \sqrt{\log \mathcal{N}(\alpha_j, \mathcal{H}_R, d_{\mathcal{H}_S})} \\
    \leq & 12 \sqrt{m+u} \int_{\alpha_{N+1}}^{\alpha_0} \sqrt{\log \mathcal{N}(\alpha, \mathcal{H}_R, d_{\mathcal{H}_S})} {\,\mathrm{d}}\alpha \leq 12 \sqrt{m+u} \int_{\alpha_{N+1}}^{\infty} \sqrt{\log \mathcal{N}(\alpha, \mathcal{H}_R, d_{\mathcal{H}_S})} {\,\mathrm{d}}\alpha .
\end{aligned}
\end{equation}
For the last term, by Khintchine-Kahane inequality \cite{latala1994},
\begin{equation}\label{term3}
    \mathbb{E}_{\bm \epsilon} \left[ \sum_{i=1}^{m+u} {\epsilon}_i \ell(\mathbf{w}^{(1)};z_i) \right] \leq \left( \sum_{i=1}^{m+u} \ell^2(\mathbf{w}^{(1)};z_i) \right)^{\frac{1}{2}} \leq b_{\ell}\sqrt{m+u}.
\end{equation}
Taking the limit as $N \to \infty$, plugging Eq.~(\ref{term1}), Eq.~(\ref{term2}) and Eq.~(\ref{term3}) into Eq.~(\ref{tr2}) and combining with Eq.~(\ref{trc1}) yield 
\begin{equation}\label{trc3}
    \mathcal{R}_{m+u}(\mathbf{w}) \leq b_{\ell} \frac{{(m+u)}^{\frac{3}{2}}}{mu} + 12 \frac{{(m+u)}^{\frac{3}{2}}}{mu} \int_{0}^\infty \sqrt{\log \mathcal{N}(r, \mathcal{H}_R, d_{\mathcal{H}_S})} {\,\mathrm{d}}r,
\end{equation}
where $\epsilon_i$ is the standard Rademacher random variable. Let $\mathcal{H}_R = \left\{ z \mapsto \ell(\mathbf{w};z) \big| \mathbf{w} \in B_R \right\}$ be the parametric function space. One can verify that $d_{H_R}(\ell(\mathbf{w};\cdot), \ell(\widetilde{\mathbf{w}};\cdot)) = \mathop{\text{max}}_{z\in \mathcal{Z}} \left| \ell(\mathbf{w}; z) - \ell(\widetilde{\mathbf{w}};z) \right|$ is a metric in $\mathcal{H}_R$. we have
\begin{equation*}
\begin{aligned}
    d_{\mathcal{H}_S} \leq \left( \frac{1}{m+u} \sum_{i=1}^{m+u} \left[ \mathop{\text{max}}_{\mathbf{w}, \widetilde{\mathbf{w}} \in B_R, z\in \mathcal{Z}} \ell(\mathbf{w}; z_i) - \ell(\widetilde{\mathbf{w}};z_i) \right]^2 \right)^{\frac{1}{2}} \leq d_{\mathcal{H}_R}.
\end{aligned}
\end{equation*}
By the definition of covering number, we have $\mathcal{N}(r, \mathcal{H}_R, d_{\mathcal{H}_S}) \leq \mathcal{N}(r, \mathcal{H}_R, d_{\mathcal{H}_R})$. Besides, applying Proposition~\ref{pro} yields
\begin{equation*}
\begin{aligned}
    d_{\mathcal{H}_R} = & \mathop{\text{max}}_{z\in \mathcal{Z}} \left| \ell(\mathbf{w}; z) - \ell(\widetilde{\mathbf{w}}; z)  \right| \leq L_{\mathcal{F}} \Vert \mathbf{w} - \widetilde{\mathbf{w}} \Vert_2.
\end{aligned}
\end{equation*}
By the definition of covering number, we have $\mathcal{N}(r, \mathcal{H}_R, d_{\mathcal{H}_R}) \leq \mathcal{N} \left( \frac{r}{L_{\mathcal{F}}}, B_R, d_2 \right)$. According to \cite{pisier_1989}, $\log \mathcal{N} \left(r, B_R, d_2 \right) \leq d \log (3R/r)$ holds. Therefore, we obtain 
\begin{equation}\label{cover1}
    \log \mathcal{N}(r, \mathcal{H}_R, d_{\mathcal{H}_S}) \leq d \log \left( \frac{3L_{\mathcal{F}}R}{r} \right).
\end{equation} 
Furthermore,
\begin{equation*}
\begin{aligned}
    d^2_{\mathcal{H}_S}(\mathbf{w}, \mathbf{w}^{(1)}) = & \frac{1}{m+u} \sum_{i=1}^{m+u} \left[ \ell(\mathbf{w}; z_i) - \ell(\mathbf{w}^{(1)}; z_i) \right]^2 \leq L^2_{\mathcal{F}} R^2,
\end{aligned}
\end{equation*}
where the last inequality is due to Proposition~\ref{pro}. This implies that 
\begin{equation}\label{cover2}
    \int_{0}^\infty \sqrt{ \log \mathcal{N}(r, \mathcal{H}_R, d_{\mathcal{H}_S})} \text{d}r = \int_{0}^{L_{\mathcal{F}}R} \sqrt{ \log \mathcal{N}(r, \mathcal{H}_R, d_{\mathcal{H}_S})} {\,\mathrm{d}}r.
\end{equation}
Combining Eq.~(\ref{trc3}), Eq.~(\ref{cover1}), and Eq.~(\ref{cover2}) yields
\begin{equation}\label{int}
\begin{aligned}
    \mathcal{R}_{m+u}(\mathbf{w}) \leq & 12 \frac{{(m+u)}^{\frac{3}{2}}}{mu} \sqrt{d} \int_{0}^{L_{\mathcal{F}}R} \sqrt{\log \left( 3L_{\mathcal{F}}R/r \right)} {\,\mathrm{d}}r \\
    \leq & 12 \frac{{(m+u)}^{\frac{3}{2}}}{mu} \sqrt{d} \left( \sqrt{\log 3} + \frac{3}{2} \sqrt{\pi} \right) L_{\mathcal{F}}R.
\end{aligned}
\end{equation}
Applying Theorem~47 in \cite{Li2021ImprovedLR} to bound $R$ in Eq.~(\ref{int}) and plugging in Eq.~(\ref{general}) with probability $1-{\delta}/2$, we conclude that with probability at least $1-\delta$,
\begin{equation*}
    R_u(\mathbf{w}^{(T+1)}) = \begin{cases}
         \mathcal{O} \left( L_{\mathcal{F}} \frac{(m+u)^{\frac{3}{2}}}{mu} \log^{\frac{1}{2}}(T) T^{\frac{1}{2}-\alpha} \log \left( \frac{1}{\delta} \right) \right) & \text{If $\alpha \in \left(0, \frac{1}{2} \right)$} \\
         \mathcal{O}\Big( L_{\mathcal{F}} \frac{(m+u)^{\frac{3}{2}}}{mu} \log(T)\log(\frac{1}{\delta}) \Big) & \text{If $\alpha=\frac{1}{2}$} \\
         \mathcal{O}\Big( L_{\mathcal{F}} \frac{(m+u)^{\frac{3}{2}}}{mu} \log^{\frac{1}{2}}(T)\log(\frac{1}{\delta}) \Big) & \text{If $\alpha \in \left(\frac{1}{2}, 1 \right]$}.
        \end{cases}
\end{equation*} 

\subsubsection{Proof of Theorem~\ref{th2}}
This proof extends the proof of Theorem~1 in literature \cite{yaniv_2007} from scalar to vector. Let $p=\frac{mu}{(m+u)^2}$, we define the vector-valued Transductive Rademacher complexities:
\begin{equation*}
    \mathcal{R}_{m+u}(\mathbf{w};p) = \mathbb{E}_{\bm{\sigma}} \left[ \mathop{\rm sup}_{\mathbf{w}\in \mathcal{W}} \left\Vert \Big(\frac{1}{m}+\frac{1}{u}\Big) \sum_{i=1}^{m+u} \sigma_i \nabla \ell(\mathbf{w};z_i) \right\Vert_2 \right],
\end{equation*}
where ${\sigma}_i$ is a random variable taking value in $\{\pm 1\}$ with probability $p$ and $0$ with probability $1-2p$. Following \cite{yaniv_2007}, we introduce the pairwise Rademacher variables $\tilde{\bm \sigma}=\{(\tilde{\sigma}_{i,1}, \tilde{\sigma}_{i,2})\}_{i=1}^{m+u}$ that satisfies: $\mathbb{P}\{(\tilde{\sigma}_{i,1}, \tilde{\sigma}_{i,2})= (\frac{1}{m}, \frac{1}{u})\}=\frac{mu}{(m+u)^2}$, $\mathbb{P}\{(\tilde{\sigma}_{i,1}, \tilde{\sigma}_{i,2})= (-\frac{1}{u}, -\frac{1}{m})\}=\frac{mu}{(m+u)^2}$, $\mathbb{P}\{(\tilde{\sigma}_{i,1}, \tilde{\sigma}_{i,2})= (\frac{1}{m}, -\frac{1}{m})\}=\frac{m^2}{(m+u)^2}$, $\mathbb{P}\{(\tilde{\sigma}_{i,1}, \tilde{\sigma}_{i,2})= (-\frac{1}{u}, \frac{1}{u})\}=\frac{u^2}{(m+u)^2}$. It can be verify that
\begin{equation*}
    \mathcal{R}_{m+u}(\mathbf{w};p) = \mathbb{E}_{\tilde{\bm \sigma}} \left[ \mathop{\rm sup}_{\mathbf{w} \in \mathcal{W}} \left\Vert \sum_{i=1}^{m+u} (\sigma_{i,1} + \sigma_{i,2}) \nabla \ell(\mathbf{w};z_i) \right\Vert_2 \right].
\end{equation*}
Denote by $F_m(\mathbf{w}) \triangleq \frac{1}{m}\sum_{i=1}^m \nabla \ell(\mathbf{w}; z_i)$ and $F_u(\mathbf{w}) \triangleq \frac{1}{u}\sum_{i=m+1}^{m+u} \nabla \ell(\mathbf{w};z_i)$ the population gradient calculated on training samples and test samples. Let $\Gamma: [m+u] \mapsto [m+u]$ be a symmetric group and $\pi \in \Gamma$ a specific permutation on samples. We have
\begin{equation*}
\begin{aligned}
    & \Vert F_{m}(\mathbf{w};\pi) - F_u(\mathbf{w};\pi) \Vert_2 \\
    \triangleq & \mathop{\rm sup}_{\mathbf{w} \in \mathcal{W}} \left\Vert \frac{1}{m} \sum_{i=1}^m \nabla \ell(\mathbf{w};z_{\pi(i)}) - \frac{1}{u} \sum_{i=m+1}^{m+u} \nabla \ell(\mathbf{w};z_{\pi(i)}) \right\Vert_2\\
    = & \mathop{\rm sup}_{\mathbf{w} \in \mathcal{W}} \bigg\Vert \frac{1}{m} \sum_{i=1}^m \nabla \ell(\mathbf{w};z_{\pi(i)}) - \frac{1}{m} \sum_{\pi' \in \Gamma} \sum_{i=1}^{m} \frac{\nabla \ell(\mathbf{w};z_{\pi'(i)}) }{(m+u)!}  + \frac{1}{u} \sum_{\pi' \in \Gamma} \sum_{i=m+1}^{m+u} \frac{\nabla \ell(\mathbf{w};z_{\pi'(i)})}{(m+u)!} - \frac{1}{u} \sum_{i=m+1}^{m+u} \nabla \ell(\mathbf{w};z_{\pi(i)}) \bigg\Vert_2 \\
    = & \mathop{\rm sup}_{\mathbf{w} \in \mathcal{W}} \bigg\Vert \sum_{\pi' \in \Gamma} \frac{1}{(m+u)!} \Big[ \frac{1}{m} \sum_{i=1}^m \big(\nabla \ell(\mathbf{w};z_{\pi(i)}) - \nabla \ell(\mathbf{w};z_{\pi'(i)}) \big) + \frac{1}{u} \sum_{i=m+1}^{m+u} \big(\nabla \ell(\mathbf{w};z_{\pi'(i)}) - \nabla \ell(\mathbf{w};z_{\pi(i)}) \big) \Big] \bigg\Vert_2\\
    \leq & \sum_{\pi' \in \Gamma} \frac{1}{(m+u)!} \mathop{\rm sup}_{\mathbf{w}} \left\Vert \frac{1}{m} \sum_{i=1}^m \big(\nabla \ell(\mathbf{w};z_{\pi(i)}) - \nabla \ell(\mathbf{w};z_{\pi'(i)}) \big) + \frac{1}{u} \sum_{i=m+1}^{m+u} \big(\nabla \ell(\mathbf{w};z_{\pi'(i)}) - \nabla \ell(\mathbf{w};z_{\pi(i)}) \big) \right\Vert_2\\
    = & \mathbb{E}_{\pi'} \mathop{\rm sup}_{\mathbf{w}} \left\Vert \frac{1}{m} \sum_{i=1}^m \big(\nabla \ell(\mathbf{w};z_{\pi(i)}) - \nabla \ell(\mathbf{w};z_{\pi'(i)}) \big) + \frac{1}{u} \sum_{i=m+1}^{m+u} \big(\nabla \ell(\mathbf{w};z_{\pi'(i)}) - \nabla \ell(\mathbf{w};z_{\pi(i)}) \big) \right\Vert_2\\
    \triangleq & \Phi(\pi).
\end{aligned}
\end{equation*}
By Proposition~\ref{pro},
\begin{equation}\label{norm_g}
\begin{aligned}
    \left\Vert \nabla \ell(\mathbf{w};z) \right\Vert_2 \leq \left\Vert \nabla \ell(\mathbf{w};z) - \nabla \ell(\mathbf{w}^{(1)};z) \right\Vert_2 + \left\Vert \ell(\mathbf{w}^{(1)};z) \right\Vert_2 \leq P_{\mathcal{F}} R + b_g. 
\end{aligned}
\end{equation}
where the second inequality is due to $\Vert \mathbf{w} - \mathbf{w}^{(1)} \Vert_2 \leq R$ and $R > 1$. By Lemma 2 in \cite{yaniv_2007}, with probability at least $1-\delta/2$ over the random permutation $\pi$, for $\mathbf{w} \in \mathcal{W}$,
\begin{equation}\label{high_pro_gradient}
    \Vert F_{m}(\mathbf{w}) - F_u(\mathbf{w}) \Vert_2 \leq \mathbb{E}_{\pi} [\Phi(\pi)] + (P_{\mathcal{F}} R + b_g)\sqrt{\frac{SQ}{2}\log \frac{2}{\delta}}.
\end{equation}
Next we discuss how to give a upper bound of $\mathbb{E}_{\pi} [\Phi(\pi)]$. To achieve this, we have to build connection between $\mathbb{E}_{\pi} [\Phi(\pi)]$ and $\mathcal{R}_{m+u}(\mathbf{w};p)$. For a given permutation $\pi$, denote by $\bm {a} \in \mathbb{R}^{m+u}$ a random vectors where $a_i = \frac{1}{m}$ if $i\in \{\pi(1),\ldots,\pi(m)\}$ else $-\frac{1}{u}$ and $\bm {b} \in \mathbb{R}^{m+u}$ a random vectors where $b_i = -\frac{1}{m}$ if $i\in \{\pi(1),\ldots,\pi(m)\}$ else $\frac{1}{u}$. To build the connection between ${\bm a}, {\bm b}$ and $\tilde{\bm \sigma}$, a extra distribution that conditioned on $\tilde{\bm \sigma}$ is introduced. Denote by $n_1(\tilde{\bm \sigma}) \triangleq \sum_{i=1}^{m+u} \mathbb{I}\{(\tilde{\sigma}_{i,1}, \tilde{\sigma}_{i,2})= (\frac{1}{m}, \frac{1}{u})\}$, $n_2(\tilde{\bm \sigma}) \triangleq \sum_{i=1}^{m+u} \mathbb{I}\{(\tilde{\sigma}_{i,1}, \tilde{\sigma}_{i,2})= (\frac{1}{m}, -\frac{1}{m}) \}$, and $n_3(\tilde{\bm \sigma}) = \sum_{i=1}^{m+u} \mathbb{I}\{(\tilde{\sigma}_{i,1}, \tilde{\sigma}_{i,2})= (-\frac{1}{u}, -\frac{1}{m})\}$ the random variables conditioned on $\tilde{\bm \sigma}$, which indicate the number of pairs appearing in elements of $\tilde{\sigma}$. Let $N_1(\tilde{\bm \sigma})=n_1(\tilde{\bm \sigma})+n_2(\tilde{\bm \sigma})$ and $N_2(\tilde{\bm \sigma})=n_2(\tilde{\bm \sigma})+n_3(\tilde{\bm \sigma})$, we denote by $\mathfrak{R}(N_1, N_2)$ the distribution of $\tilde{\bm \sigma}$ conditioned on $n_1$, $n_2$, and $n_3$, which has fixed number of pairs and the randomness comes from permutations. Thus, ${\bm a} + {\bm b}$ and $\tilde{\bm \sigma} \sim \mathfrak{R}(N_1(\tilde{\bm \sigma})=m, N_2(\tilde{\bm \sigma})=m)$ have the same distribution.
Then we have
\begin{equation*}
\begin{aligned}
    & \mathbb{E}_{\pi} [\Phi(\pi)] \\
    = & \mathbb{E}_{\pi,\pi'} \mathop{\rm sup}_{\mathbf{w} \in \mathcal{W}} \left\Vert \frac{1}{m} \sum_{i=1}^m \big(\nabla \ell(\mathbf{w};z_{\pi(i)}) - \nabla \ell(\mathbf{w};z_{\pi'(i)}) \big) + \frac{1}{u} \sum_{i=m+1}^{m+u} \big(\nabla \ell(\mathbf{w};z_{\pi'(i)}) - \nabla \ell(\mathbf{w};z_{\pi(i)}) \big) \right\Vert_2 \\
    = & \mathbb{E}_{\pi,\pi'} \mathop{\rm sup}_{\mathbf{w} \in \mathcal{W}} \left\Vert \sum_{i=1}^{m+u} (a_{\pi(i)}+b_{\pi'(i)}) \nabla \ell(\mathbf{w};z_i) \right\Vert_2 \\
    = & \mathbb{E}_{\tilde{\bm \sigma}\sim \mathfrak{R}(m,m)} \mathop{\rm sup}_{\mathbf{w} \in \mathcal{W}} \left\Vert \sum_{i=1}^{m+u} (\tilde{\sigma}_{i,1}+\tilde{\sigma}_{i,2}) \nabla \ell(\mathbf{w};z_i) \right\Vert_2.
\end{aligned}
\end{equation*}
Denote by
\begin{equation*}
    \psi(N, N') = \mathbb{E}_{\tilde{\bm \sigma} \sim \mathfrak{R}(N_1, N_2)} \left[ \mathop{\rm sup}_{\mathbf{w} \in \mathcal{W}} \left\Vert \sum_{i=1}^{m+u} (\sigma_{i,1} + \sigma_{i,2}) \nabla \ell(\mathbf{w};z_i) \right\Vert_2 \right],
\end{equation*}
the Transductive Rademacher complexity where $\tilde{\bm \sigma}$ follows $\mathfrak{R}(N, N')$ for given $N_1$ and $N_2$. One can find that 
\begin{equation}\label{exp_psi1}
    \mathcal{R}_{m+u}(\mathbf{w};p)=\mathbb{E}_{N_1(\tilde{\bm \sigma}), N_2(\tilde{\bm \sigma})}[\psi(N_1(\tilde{\bm \sigma}), N_2(\tilde{\bm \sigma}))].   
\end{equation}
Besides, one can find that $\mathbb{E}_{\tilde{\bm \sigma}}[N_1(\tilde{\bm \sigma})]=m$ and $\mathbb{E}_{\tilde{\bm \sigma}}[N_2(\tilde{\bm \sigma})]=m$ hold. Therefore, we have 
\begin{equation}\label{exp_psi2}
    \mathbb{E}_{\pi}[\Phi(\pi)]=\psi(\mathbb{E}_{\tilde{\bm \sigma}}[N_1(\tilde{\bm \sigma})],\mathbb{E}_{\tilde{\bm \sigma}}[N_2(\tilde{\bm \sigma})]).    
\end{equation}
The last step is to give a upper bound of $\psi(\mathbb{E}_{\tilde{\bm \sigma}}[N_1(\tilde{\bm \sigma})],\mathbb{E}_{\tilde{\bm \sigma}}[N_2(\tilde{\bm \sigma})]) - \mathbb{E}_{N_1(\tilde{\bm \sigma}), N_2(\tilde{\bm \sigma})}[\psi(N_1(\tilde{\bm \sigma}), N_2(\tilde{\bm \sigma}))]$. Recall the definitions of $\psi(N_1, N_2)$ and $\psi(N'_1, N_2)$ are:
\begin{equation*}
\begin{aligned}
    & \psi(N_1, N_2) \\
    = & \mathbb{E}_{\pi,\pi'} \mathop{\rm sup}_{\mathbf{w} \in \mathcal{W}} \left\Vert \frac{1}{m} \sum_{i=1}^{N_1} \nabla \ell(\mathbf{w};z_{\pi(i)}) - \frac{1}{m} \sum_{i=1}^{N_2} \nabla \ell(\mathbf{w};z_{\pi'(i)}) + \frac{1}{u} \sum_{i=N_2+1}^{m+u} \nabla \ell(\mathbf{w};z_{\pi'(i)}) - \frac{1}{u} \sum_{i=N_1+1}^{m+u} \nabla \ell(\mathbf{w};z_{\pi(i)}) \right\Vert_2, \\
    & \psi(N'_1, N_2) \\
    = & \mathbb{E}_{\pi,\pi'} \mathop{\rm sup}_{\mathbf{w} \in \mathcal{W}} \left\Vert \frac{1}{m} \sum_{i=1}^{N'_1} \nabla \ell(\mathbf{w};z_{\pi(i)}) - \frac{1}{m} \sum_{i=1}^{N_2} \nabla \ell(\mathbf{w};z_{\pi'(i)}) + \frac{1}{u} \sum_{i=N_2+1}^{m+u} \nabla \ell(\mathbf{w};z_{\pi'(i)}) - \frac{1}{u} \sum_{i=N'_1+1}^{m+u} \nabla \ell(\mathbf{w};z_{\pi(i)}) \right\Vert_2.
\end{aligned}
\end{equation*}
Without loss of generality, assume that $N'_1 \leq N_1$. Then we have
\begin{equation}\label{psi1}
\begin{aligned}
    & \vert \psi(N_1,N_2) - \psi(N'_1,N_2) \vert \\
    \leq & \mathbb{E}_{\pi} \mathop{\rm sup}_{\mathbf{w} \in \mathcal{W}} \left\Vert \left(\frac{1}{u}+\frac{1}{m} \right)\sum_{i=N'_1+1}^{N_1} \nabla \ell(\mathbf{w};z_{\pi(i)}) \right\Vert_2 \leq \vert N_1 - N'_1 \vert (PR+b_g) \left(\frac{1}{m}+\frac{1}{u} \right).
\end{aligned}
\end{equation}
Similarly, we have
\begin{equation}\label{psi2}
\begin{aligned}
    & \vert \psi(N_1,N_2) - \psi(N_1,N'_2) \vert \\
    \leq & \mathbb{E}_{\pi} \mathop{\rm sup}_{\mathbf{w} \in \mathcal{W}} \left\Vert \left(\frac{1}{u}+\frac{1}{m} \right) \sum_{i=N'_2+1}^{N_2} \nabla \ell(\mathbf{w};z_{\pi(i)}) \right\Vert_2 \leq \vert N_2 - N'_2 \vert (PR+b_g) \left(\frac{1}{m}+\frac{1}{u} \right).
\end{aligned}
\end{equation}
Combining Eq.~(\ref{psi1}), Eq.~(\ref{psi2}) and the inequality from \cite{devroye}, we have
\begin{equation*}
\begin{aligned}
    & \mathbb{P}_{N_1(\tilde{\bm \sigma}), N_2(\tilde{\bm \sigma})} \{ |\psi(N_1(\tilde{\bm \sigma}), N_2(\tilde{\bm \sigma})) - \psi(\mathbb{E}_{\tilde{\bm \sigma}}[N_1(\tilde{\bm \sigma})], \mathbb{E}_{\tilde{\bm \sigma}}[N_2(\tilde{\bm \sigma})])| \geq \epsilon \} \\
    \leq & \mathbb{P}_{N_1(\tilde{\bm \sigma}), N_2(\tilde{\bm \sigma})} \{ |\psi(N_1(\tilde{\bm \sigma}), N_2(\tilde{\bm \sigma})) - \psi(N_1, \mathbb{E}_{\tilde{\bm \sigma}}[N_2(\tilde{\bm \sigma})])| \geq \epsilon/2 \} \\
    & + \mathbb{P}_{\tilde{\bm \sigma}} \{ |\psi(N_1, \mathbb{E}_{\tilde{\bm \sigma}}[N_2(\tilde{\bm \sigma})]) - \psi(\mathbb{E}_{\tilde{\bm \sigma}}[N_1(\tilde{\bm \sigma})], \mathbb{E}_{\tilde{\bm \sigma}}[N_2(\tilde{\bm \sigma})])| \geq \epsilon/2 \} \\
    \leq & \mathbb{P}_{N_1(\tilde{\bm \sigma}), N_2(\tilde{\bm \sigma})} \{ |N_2(\tilde{\bm \sigma}) - \mathbb{E}_{\tilde{\bm \sigma}}[N_2(\tilde{\bm \sigma})]|(PR+b_g)Q \geq \epsilon/2 \} + \mathbb{P}_{\tilde{\bm \sigma}} \{ |N_1(\tilde{\bm \sigma}) - \mathbb{E}_{\tilde{\bm \sigma}}[N_1(\tilde{\bm \sigma})] (PR+b_g)Q \geq \epsilon/2 \} \\
    \leq & 4\exp\{-3\epsilon^2/(32m (PR^\alpha+b_g)^2 Q)\}.
\end{aligned}
\end{equation*}
Applying the fact from Problem~12.1 in \cite{devroye}, the following inequality holds
\begin{equation}\label{exp_psi3}
    \mathbb{E}_{N_1(\tilde{\bm \sigma}), N_2(\tilde{\bm \sigma})} |\psi(N_1(\tilde{\bm \sigma}), N_2(\tilde{\bm \sigma})) - \psi(\mathbb{E}_{\tilde{\bm \sigma}}[N_1(\tilde{\bm \sigma})], \mathbb{E}_{\tilde{\bm \sigma}}[N_2(\tilde{\bm \sigma})])| \leq c_0 (PR+b_g) Q \sqrt{ {\min }(m, u)},
\end{equation}
where $c_0=\sqrt{\frac{32\log(4e)}{3}}$. Plugging Eq.~(\ref{exp_psi1}), Eq.~(\ref{exp_psi2}) and Eq.~(\ref{exp_psi3}) into Eq.~(\ref{high_pro_gradient}), with probability at least $1-\delta/2$,
\begin{equation*}
\begin{aligned}
    & \Vert F_{m}(\mathbf{w}) - F_u(\mathbf{w}) \Vert_2 \\
    \leq & \mathbb{E}_{\pi} [\Phi(\pi)] + (P_{\mathcal{F}}R + b_g)\sqrt{\frac{SQ}{2}\log \frac{2}{\delta}} \\
    = & \psi(\mathbb{E}_{\tilde{\bm \sigma}}[N_1(\tilde{\bm \sigma})],\mathbb{E}_{\tilde{\bm \sigma}}[N_2(\tilde{\bm \sigma})]) + (P_{\mathcal{F}}R + b_g)\sqrt{\frac{SQ}{2}\log \frac{2}{\delta}} \\
    \leq & \mathbb{E}_{N_1(\tilde{\bm \sigma}), N_2(\tilde{\bm \sigma})}[\psi(N_1(\tilde{\bm \sigma}), N_2(\tilde{\bm \sigma}))] + c_0 (P_{\mathcal{F}}R+b_g) Q \sqrt{ {\min }(m, u)} + (P_{\mathcal{F}}R+b_g) \sqrt{\frac{SQ}{2}\log \frac{2}{\delta}} \\
    = & \mathcal{R}_{m+u} (\mathbf{w};p) + c_0 (P_{\mathcal{F}}R+b_g) Q \sqrt{ {\min }(m, u)} + (P_{\mathcal{F}}R+b_g) \sqrt{\frac{SQ}{2}\log \frac{2}{\delta}}.
\end{aligned}
\end{equation*}
Till now, we have obtained the following inequality holds with probability at least $1-\delta/2$:
\begin{equation}\label{main_gradient}
\begin{aligned}
    & \mathop{\rm sup}_{\mathbf{w} \in B_R} \left\Vert \frac{1}{m} \sum_{i=1}^m \nabla \ell(\mathbf{w};z_i) - \frac{1}{u} \sum_{i=m+1}^{m+u} \nabla \ell(\mathbf{w};z_i) \right\Vert_2 \\
    \leq & \Big(\frac{1}{m}+\frac{1}{u}\Big) \mathbb{E}_{\bm{\epsilon}} \left[ \mathop{\rm sup}_{\mathbf{w} \in B_R } \Big\Vert \sum_{i=1}^{m+u} \epsilon_i {\rm vec}(\nabla \ell(\mathbf{w};z_i)) \Big\Vert_2 \right] + c_0 (PR+b_g) Q \sqrt{ {\min }(m, u)} + (PR+b_g) \sqrt{\frac{S}{2}(\frac{1}{m}+\frac{1}{u})\log \frac{1}{\delta}}.
\end{aligned}
\end{equation}
Let $\mathcal{H}_R = \left\{ (z,z') \mapsto \langle \nabla \ell(\mathbf{w}; z), \nabla \ell(\mathbf{w}; z') \rangle \big| \mathbf{w} \in B_R \right\}$ be the parametric function space, one can verify that 
\begin{equation*}
    d_{\mathcal{H}_R} = \mathop{\rm max}_{z, z' \in \mathcal{Z}} \vert \langle \nabla \ell(\mathbf{w}; z), \nabla \ell(\mathbf{w}; z') \rangle - \langle \nabla \ell(\tilde{\mathbf{w}}; z), \nabla \ell(\tilde{\mathbf{w}}; z') \rangle \vert
\end{equation*}
is a metric in $\mathcal{H}_R$. Define
\begin{equation*}
\begin{aligned}
    d_{\mathcal{H}_S} (\mathbf{w}, \tilde{\mathbf{w}}) = \left( \frac{1}{(m+u)^2} \sum_{1\leq i < j \leq m+u} \left| \langle \nabla \ell(\mathbf{w}; z_i), \nabla \ell(\mathbf{w}; z_j) \rangle - \langle \nabla \ell(\tilde{\mathbf{w}}; z_i), \nabla \ell(\tilde{\mathbf{w}}; z_j) \rangle \right|^2 \right)^{\frac{1}{2}},
\end{aligned}
\end{equation*}
we have $d_{\mathcal{H}_S} (\mathbf{w}, \tilde{\mathbf{w}}) \leq d_{\mathcal{H}_R}$. By the definition of covering number, we have $\mathcal{N}(r, \mathcal{H}_R, d_{\mathcal{H}_S}) \leq \mathcal{N}(r, \mathcal{H}_R, d_{\mathcal{H}_R})$. Besides, applying Proposition~\ref{pro} yields
\begin{equation*}
\begin{aligned}
    & (m+u)^2d^2_S(\mathbf{w}, \tilde{\mathbf{w}}) \\
    = & \sum_{1\leq i < j \leq m+u} \big| \langle \nabla \ell(\mathbf{w}; z_i), \nabla \ell(\mathbf{w}; z_j) \rangle - \langle \nabla \ell(\tilde{\mathbf{w}}; z_i), \nabla \ell(\tilde{\mathbf{w}}; z_j) \rangle \big|^2 \\
    \leq & \sum_{1\leq i < j \leq m+u} 2\big| \langle \nabla \ell(\mathbf{w}; z_i) - \nabla \ell(\tilde{\mathbf{w}}; z_i), \nabla \ell(\mathbf{w}; z_j) \rangle\big|^2 + 2\big| \langle \nabla \ell(\tilde{\mathbf{w}}; z_i), \nabla \ell({\mathbf{w}}; z_j) - \nabla \ell(\tilde{\mathbf{w}}; z_j) \rangle \big|^2 \\
    \leq & \sum_{1\leq i < j \leq m+u} 2\left\Vert \nabla \ell(\mathbf{w}; z_i) - \nabla \ell(\tilde{\mathbf{w}}; z_i) \right\Vert^2_2 \left\Vert \nabla \ell(\mathbf{w}; z_j) \right\Vert^2_2 + 2\left\Vert \nabla \ell(\tilde{\mathbf{w}}; z_i) \right\Vert^2_2 \left\Vert \nabla \ell({\mathbf{w}}; z_j) - \nabla \ell(\tilde{\mathbf{w}}; z_j) \rangle \right\Vert^2_2 \\
    \leq & 2(m+u)(m+u-1)P^2_{\mathcal{F}}{(P_{\mathcal{F}}R + b_g)}^2 \mathop{\rm max} \left\{ \left\Vert \mathbf{w} - \tilde{\mathbf{w}} \right\Vert^{2\alpha}_2, \left\Vert \mathbf{w} - \tilde{\mathbf{w}} \right\Vert^2_2 \right\}.
\end{aligned}
\end{equation*}
By the definition of covering number, we have 
\begin{equation*}
 \mathcal{N}(r, \mathcal{H}_R, d_{\mathcal{H}_R}) \leq \mathcal{N} \left( \mathop{\rm min} \left\{ \left(\frac{r}{\sqrt{2}P_{\mathcal{F}}(P_{\mathcal{F}}R + b_g)}\right)^{\frac{1}{\tilde{\alpha}}}, \frac{r}{\sqrt{2} P_{\mathcal{F}}(P_{\mathcal{F}}R + b_g)} \right\}, B_R, d_2 \right).  
\end{equation*}
According to \cite{pisier_1989}, $\log \mathcal{N} \left(r, B_R, d_2 \right) \leq d \log (3R/r)$ holds. Therefore, we obtain 
\begin{equation}\label{cover3}
\begin{aligned}
    \log \mathcal{N}(r, \mathcal{H}_R, d_{\mathcal{H}_S}) \leq & \mathop{\rm max} \left\{ d\log \left( \frac{3R {(\sqrt{2}P_{\mathcal{F}})}^{\frac{1}{\tilde{\alpha}}}(P_{\mathcal{F}}R + b_g)^{\frac{1}{\tilde{\alpha}}}}{r^{\frac{1}{\tilde{\alpha}}}} \right), d\log \left( \frac{3\sqrt{2}P_{\mathcal{F}}R(P_{\mathcal{F}}R+b_g)}{r} \right) \right\}.
\end{aligned}
\end{equation} 
Denote by $\frac{1}{m+u}\mathbb{E}_{\bm \epsilon} \sum_{1\leq i <j \leq m+u} \sigma_i \sigma_j h(z_i, z_j) $ the transductive Rademacher chaos complexity, we have
\begin{equation}\label{complexity_gradient}
\begin{aligned}
    & \left( \mathbb{E}_{\bm{\sigma}} \left[ \mathop{\rm sup}_{\mathbf{w} \in B_R } \left\Vert \sum_{i=1}^{m+u} \sigma_i \nabla \ell(\mathbf{w};z_i) \right\Vert_2 \right] \right)^2 \\
    \leq & \mathbb{E}_{\bm{\sigma}} \left[ \mathop{\rm sup}_{\mathbf{w} \in B_R } \left\Vert \sum_{i=1}^{m+u} \sigma_i \nabla \ell(\mathbf{w};z_i) \right\Vert^2_2 \right] \\
    = & \mathbb{E}_{\bm{\sigma}} \left[ \mathop{\rm sup}_{\mathbf{w}\in B_R } \sum_{i,j=1}^{m+u} \sigma_i \sigma_j \left\langle \nabla \ell(\mathbf{w};z_i), \nabla \ell(\mathbf{w};z_j) \right\rangle \right] \\
    = & \mathbb{E}_{\bm{\sigma}} \left[ \mathop{\rm sup}_{\mathbf{w} \in B_R } \sum_{i=1}^{m+u} \sigma^2_i \left\Vert \nabla \ell(\mathbf{w};z_i) \right\Vert^2_2 \right] + \mathbb{E}_{\bm{\sigma}} \left[ \mathop{\rm sup}_{\mathbf{w} \in B_R } \sum_{i,j=1,i\ne j}^{m+u} \sigma_i \sigma_j \langle \nabla \ell(\mathbf{w};z_i), \nabla \ell(\mathbf{w};z_j) \rangle \right] \\
    \leq & (m+u)(P_{\mathcal{F}}R+b_g)^2 + 2(m+u)\mathcal{U}(\mathcal{H}_R),
\end{aligned}
\end{equation}
Note that
\begin{equation*}
\begin{aligned}
    & (m+u)^2d^2_S(\mathbf{w}, \mathbf{w}^{(1)}) \\
    = & \sum_{1\leq i < j \leq m+u} \big| \langle \nabla \ell(\mathbf{w}; z_i), \nabla \ell(\mathbf{w}; z_j) \rangle - \langle \nabla \ell(\mathbf{w}^{(1)}; z_i), \nabla \ell(\mathbf{w}^{(1)}; z_j) \rangle \big| \\
    \leq & \sum_{1\leq i < j \leq m+u} 2\left\Vert \nabla \ell(\mathbf{w}; z_i) - \nabla \ell({\mathbf{w}}^{(1)}; z_i) \right\Vert^2_2 \left\Vert \nabla \ell(\mathbf{w}; z_j) \right\Vert^2_2 + 2\left\Vert \nabla \ell({\mathbf{w}}^{(1)}; z_i) \right\Vert^2_2 \left\Vert \nabla \ell({\mathbf{w}}; z_j) - \nabla \ell({\mathbf{w}}^{(1)}; z_j) \rangle \right\Vert^2_2 \\
    \leq & 2(m+u)(m+u-1)P^2_{\mathcal{F}}R^2{(P_{\mathcal{F}}R + b_g)}^2.
\end{aligned}
\end{equation*}
By Lemma~\ref{chaos}, we have $\mathcal{U}(\mathcal{H}_R) \leq 24e \int_{0}^{\sqrt{2}P_{\mathcal{F}}R{(P_{\mathcal{F}}R + b_g)}} \log \mathcal{N}(r, \mathcal{H}_R, d_{\mathcal{H}_S}) {\rm d}r$. Plugging in Eq.~(\ref{cover3}) yields
\begin{equation*}
\begin{aligned}
    \mathcal{U}(\mathcal{H}_R) \leq & 24e \int_{0}^{2P_{\mathcal{F}}R{(P_{\mathcal{F}}R + b_g)}} \log \left(1 + \mathcal{N}(r, \mathcal{H}_R, d_{\mathcal{H}_S}) \right) {\,\mathrm{d}}r \\
    \leq & 24e \int_{0}^{\sqrt{2}P_{\mathcal{F}}R{(P_{\mathcal{F}}R + b_g)}} \left(\log 2 + \log \mathcal{N}(r, \mathcal{H}_R, d_{\mathcal{H}_S}) \right) {\,\mathrm{d}}r \\
    = & 24 \sqrt{2} e P_{\mathcal{F}}R(P_{\mathcal{F}}R + b_g)\log 2 +  24e d \int_{0}^{\sqrt{2}P_{\mathcal{F}}(P_{\mathcal{F}}R + b_g)} \log \left( \frac{3R {(2P_{\mathcal{F}})}^{\frac{1}{\tilde{\alpha}}}(P_{\mathcal{F}}R + b_g)^{\frac{1}{\tilde{\alpha}}}}{r^{\frac{1}{\tilde{\alpha}}}} \right) {\,\mathrm{d}}r \\
    & + 24e d \int_{\sqrt{2}P_{\mathcal{F}}(P_{\mathcal{F}}R + b_g)}^{\sqrt{2}P_{\mathcal{F}}R(P_{\mathcal{F}}R + b_g)} \log \left( \frac{3\sqrt{2}P_{\mathcal{F}}R(P_{\mathcal{F}}R+b_g)}{r} \right) {\,\mathrm{d}}r \\
    \leq &  24\sqrt{2} e P_{\mathcal{F}}(P_{\mathcal{F}}R+b_g) \left[ d\log (3e^{\frac{1}{\alpha}}R) + d R \log (3e) + R \log 2 \right]. 
\end{aligned}
\end{equation*}
Applying Theorem~47 in \cite{Li2021ImprovedLR} to bound $R$ in Eq.~(\ref{int}) with probability $1-{\delta}/2$ and combining it with Eqs.~(\ref{complexity_gradient}, \ref{main_gradient}), with probability at least $1-\delta$,
\begin{equation*}
\begin{aligned}
    & \mathop{\rm sup}_{\mathbf{w} \in B_R} \left\Vert \frac{1}{m} \sum_{i=1}^m \nabla \ell(\mathbf{w};z_i) - \frac{1}{u} \sum_{i=m+1}^{m+u} \nabla \ell(\mathbf{w};z_i) \right\Vert_2 \\
    = & \begin{cases}
        \mathcal{O}\Big( \frac{{(m+u)}^{\frac{3}{2}}}{mu} \log^{\frac{1}{2}}(T)T^\frac{1-2\alpha}{2}\log(\frac{1}{\delta}) \Big) & \text{if $\alpha \in (0, \frac{1}{2})$} \\
        \mathcal{O}\Big( \frac{{(m+u)}^{\frac{3}{2}}}{mu} \log(T)\log(\frac{1}{\delta}) \Big) & \text{if $\alpha = \frac{1}{2}$} \\
        \mathcal{O}\Big( \frac{{(m+u)}^{\frac{3}{2}}}{mu} \log^{\frac{1}{2}}(T)\log(\frac{1}{\delta}) \Big) & \text{if $\alpha \in (\frac{1}{2}, 1]$}.  
    \end{cases}
\end{aligned}
\end{equation*}

\subsubsection{Proof of Theorem~\ref{th3}}

By Lemma~43 in \cite{Li2021ImprovedLR}, we have
\begin{equation}\label{th3_eq1}
    R_m(\mathbf{w}^{T+1}) - R_m(\hat{\mathbf{w}}^*) =
    \begin{cases}
        \mathcal{O} \Big( \frac{1}{T^\alpha} \Big) & \text{if $\alpha \in (0, 1)$} \\
        \mathcal{O} \Big( \frac{\log(T) \log^3 (1/\delta)}{T} \Big) & \text{if $\alpha = 1$.}
    \end{cases}
\end{equation}
By Theorem~\ref{th1},
\begin{equation}\label{th3_eq2}
    R_u(\mathbf{w}^{(T+1)}) - R_m(\mathbf{w}^{(T+1)}) = \begin{cases}
         \mathcal{O} \left( L_{\mathcal{F}} \frac{(m+u)^{\frac{3}{2}}}{mu} \log^{\frac{1}{2}}(T) T^{\frac{1}{2}-\alpha} \log \left( \frac{1}{\delta} \right) \right) & \text{if $\alpha \in \left(0, \frac{1}{2} \right)$} \\
         \mathcal{O}\Big( L_{\mathcal{F}} \frac{(m+u)^{\frac{3}{2}}}{mu} \log(T)\log(\frac{1}{\delta}) \Big) & \text{if $\alpha=\frac{1}{2}$} \\
         \mathcal{O}\Big( L_{\mathcal{F}} \frac{(m+u)^{\frac{3}{2}}}{mu} \log^{\frac{1}{2}}(T)\log(\frac{1}{\delta}) \Big) & \text{if $\alpha \in \left(\frac{1}{2}, 1 \right]$}.
        \end{cases}
\end{equation}
Combing Eq.~(\ref{th3_eq1}) and Eq.~(\ref{th3_eq2}) yields the result.

\subsection{Proof of Section~\ref{gnn_bound}}

\subsubsection{Proof of Proposition~\ref{gnn}}
We first analyze the Lipschitz continuity. Denote by $\mathbf{Z}^{(1)}=g(\tilde{\mathbf{A}})\mathbf{X}$, $\mathbf{H}^{(1)}=\sigma(\mathbf{Z}^{(1)}\mathbf{W}_1)$ and $\mathbf{Z}^{(2)}=g(\tilde{\mathbf{A}})\mathbf{H}^{(1)}$, the forward process of GCN is given by $\hat{\mathbf{Y}} = {\rm Softmax}\left( \mathbf{Z}^{(2)} \mathbf{W}_2 \right)$. First, we have
\begin{equation}\label{hidden_gcn}
\begin{aligned}
    \mathop{\rm max}_{i\in [n]} \Vert \mathbf{H}^{(1)}_{i*} \Vert_2 = & \left\Vert \sigma \left( \sum_{j=1}^n \left[g(\tilde{\mathbf{A}}) \right]_{ij} \mathbf{X}_{j*} \mathbf{W}_1 \right) \right\Vert_2 \leq \left\Vert \sum_{j=1}^n \left[g(\tilde{\mathbf{A}}) \right]_{ij} \mathbf{X}_{j*} \mathbf{W}_1 \right\Vert_2 \\
    \leq & \sum_{j=1}^n \left[g(\tilde{\mathbf{A}}) \right]_{ij} \left\Vert \mathbf{X}_{j*} \mathbf{W}_1 \right\Vert_2 \leq c_X c_W \Vert g(\tilde{\mathbf{A}}) \Vert_\infty,
\end{aligned}
\end{equation}
where the first inequality is due to the definition of $\sigma(\cdot)$. Similarly, 
\begin{equation}\label{z_gcn}
    \mathop{\rm max}_{i\in [n]} \Vert \mathbf{Z}^{(1)}_{i*} \Vert_2 = \left\Vert  \sum_{j=1}^n \left[g(\tilde{\mathbf{A}}) \right]_{ij} \mathbf{X}_{j*} \right\Vert_2 \leq \sum_{j=1}^n \left[g(\tilde{\mathbf{A}}) \right]_{ij} \left\Vert \mathbf{X}_{j*} \right\Vert_2 \leq c_X \Vert g(\tilde{\mathbf{A}}) \Vert_\infty
\end{equation}
holds. Besides,
\begin{equation*}
    \mathop{\rm max}_{i\in [n]} \Vert \mathbf{Z}^{(2)}_{i*} \Vert_2 = \left\Vert  \sum_{j=1}^n \left[g(\tilde{\mathbf{A}}) \right]_{ij} \mathbf{H}^{(1)}_{j*} \right\Vert_2 \leq \sum_{j=1}^n \left[g(\tilde{\mathbf{A}}) \right]_{ij} \left\Vert \mathbf{H}^{(1)}_{j*} \right\Vert_2 \leq c_X c_W \Vert g(\tilde{\mathbf{A}}) \Vert^2_\infty.
\end{equation*}
Then we analyze how $\ell(\mathbf{W}_1, \mathbf{W}_2; z_i)$ change w.r.t. $\mathbf{W}_2$ for fixed $\mathbf{W}_1$ and $i \in [n]$: 
\begin{equation*}
\begin{aligned}
    & \vert \ell(\mathbf{W}_1, \mathbf{W}_2, z_i) - \ell(\mathbf{W}_1, \mathbf{W}'_2, z_i) \vert \\
    \leq & \sqrt{2} \left\Vert \mathbf{Z}^{(1)}_{i*} (\mathbf{W}_2 - \mathbf{W}'_2) \right\Vert_2 = \sqrt{2} \left\Vert \sum_{j=1}^n \left[g(\tilde{\mathbf{A}}) \right]_{ij} \mathbf{H}^{(1)}_{j*} (\mathbf{W}_2 - \mathbf{W}'_2) \right\Vert_2 \\
    \leq & \sqrt{2} \sum_{j=1}^n \left[g(\tilde{\mathbf{A}}) \right]_{ij} \left\Vert \mathbf{H}^{(1)}_{j*} \right\Vert_2 \left\Vert \mathbf{W}_2 - \mathbf{W}_2 \right\Vert \leq \sqrt{2} \left\Vert g(\tilde{\mathbf{A}}) \right\Vert_{\infty} \mathop{\rm max}_{i\in [n]} \Vert \mathbf{H}^{(1)}_{i*} \Vert_2 \left\Vert \mathbf{W}_2 - \mathbf{W}_2 \right\Vert \\
    \leq & c_X c_W \sqrt{2} \left\Vert g(\tilde{\mathbf{A}}) \right\Vert^2_{\infty} \left\Vert \mathbf{W}_2 - \mathbf{W}_2 \right\Vert \leq c_X c_W \sqrt{2} \left\Vert g(\tilde{\mathbf{A}}) \right\Vert^2_{\infty} \left\Vert {\rm vec} \left[\mathbf{W}_2 \right] - {\rm vec}\left[\mathbf{W}'_2\right] \right\Vert_2 ,
\end{aligned}
\end{equation*}
where the first inequality is due to the Lipschitz continuity property of ${\rm softmax}$, and the last inequality is obtained by Eq.~(\ref{hidden_gcn}). 
Then we analyze the change of $\ell(\mathbf{W}_1, \mathbf{W}_2; z_i)$ change w.r.t. $\mathbf{W}_1$ for fixed $\mathbf{W}_2$ and $i \in [n]$. Note that $\mathbf{Z}^{(1)}$ and $\mathbf{H}^{(1)}$ are function of $\mathbf{W}_1$ in this case, which we denote by $\mathbf{Z}^{(1)}(\mathbf{W}_1)$ and $\mathbf{H}^{(1)}(\mathbf{W}_1)$, respectively. Then,
\begin{equation*}
\begin{aligned}
    & \vert \ell(\mathbf{W}_1, \mathbf{W}_2, z_i) - \ell(\mathbf{W}'_1, \mathbf{W}_2, z_i) \vert \\
    \leq & \sqrt{2} \left\Vert (\mathbf{Z}^{(1)}_{i*}(\mathbf{W}_1) - \mathbf{Z}^{(1)}_{i*}(\mathbf{W}'_1))\mathbf{W}_2 \right\Vert_2 \\
    \leq & c_W \sqrt{2} \left\Vert \sum_{j=1}^n \left[g(\tilde{\mathbf{A}}) \right]_{ij} (\mathbf{H}^{(1)}_{j*}(\mathbf{W}_1) - \mathbf{H}^{(1)}_{j*}(\mathbf{W}'_1)) \right\Vert_2 \\
    \leq & c_W \sqrt{2} \sum_{j=1}^n \left[g(\tilde{\mathbf{A}}) \right]_{ij} \left\Vert \mathbf{Z}^{(1)}_{j*} (\mathbf{W}_1 - \mathbf{W}'_1) \right\Vert_2 \\
    \leq & c_W \sqrt{2} \left\Vert g(\tilde{\mathbf{A}}) \right\Vert_\infty \mathop{\rm max}_{i\in [n]} \left\Vert \mathbf{Z}^{(1)}_{i*} \right\Vert_2 \left\Vert \mathbf{W}_1 - \mathbf{W}_1 \right\Vert \leq c_X c_W \sqrt{2} \left\Vert g(\tilde{\mathbf{A}}) \right\Vert^2_{\infty} \left\Vert {\rm vec} \left[\mathbf{W}_1 \right] - {\rm vec}\left[\mathbf{W}'_1\right] \right\Vert_2.
\end{aligned}
\end{equation*}
Let $L_1 = L_2 = c_X c_W \sqrt{2} \left\Vert g(\tilde{\mathbf{A}}) \right\Vert^2_{\infty}$, we conclude that $\vert \ell(\mathbf{w}) - \ell(\mathbf{w}') \vert \leq L_{\mathcal{F}} \Vert \mathbf{w} - \mathbf{w}' \Vert_2$ holds with $L_{\mathcal{F}} = 2 c_X c_W \left\Vert g(\tilde{\mathbf{A}}) \right\Vert^2_{\infty}$. By the chain rule, we have
\begin{equation*}
\begin{aligned}
    & \frac{\partial \ell(\mathbf{W}_1, \mathbf{W}_2; z_i)}{\partial {\rm vec}\left[\mathbf{W}_2\right]} = (\hat{\mathbf{y}}_i - \mathbf{y}_i) \otimes \mathbf{Z}^{(2)}_{i*}, \\
    & \frac{\partial \ell(\mathbf{W}_1, \mathbf{W}_2; z_i)}{\partial {\rm vec}\left[\mathbf{W}_1\right]} = \sum_{j=1}^n \left[ g(\tilde{\mathbf{A}}) \right]_{ij} \left( \sigma'\left(\mathbf{Z}^{(1)}_{j*} \mathbf{W}_1 \right) \odot (\hat{\mathbf{y}}_i - \mathbf{y}_i)\mathbf{W}^\top_2 \right) \otimes \mathbf{Z}^{(1)}_{j*}.
\end{aligned}
\end{equation*}
We first analyze how $\frac{\partial \ell(\mathbf{W}_1, \mathbf{W}_2; z_i)}{\partial {\rm vec}\left[\mathbf{W}_2\right]}$ change w.r.t. $\mathbf{W}_1$ and $\mathbf{W}_2$. Note that
\begin{equation*}
\begin{aligned}
    & \left\Vert \frac{\partial \ell(\mathbf{W}_1, \mathbf{W}_2; z_i)}{\partial {\rm vec}\left[\mathbf{W}_2\right]}(\mathbf{W}_1) - \frac{\partial \ell(\mathbf{W}_1, \mathbf{W}_2; z_i)}{\partial {\rm vec}\left[\mathbf{W}_2\right]}(\mathbf{W}'_1) \right\Vert_2 \\
    = & \Vert \hat{\mathbf{y}}_i - \mathbf{y}_i \Vert_2 \left\Vert \mathbf{Z}^{(2)}_{i*}(\mathbf{W}_1) - \mathbf{Z}^{(2)}_{i*}(\mathbf{W}'_1) \right\Vert_2 + \Vert \hat{\mathbf{y}}_i(\mathbf{W}_1) - \hat{\mathbf{y}}_i(\mathbf{W}'_1) \Vert_2 \left\Vert \mathbf{Z}^{(2)}_{i*} \right\Vert_2 \\ 
    \leq & \left( \sqrt{2} + 2c_X c^2_W \left\Vert g(\tilde{\mathbf{A}}) \right\Vert^2_\infty \right) \left\Vert \mathbf{Z}^{(2)}_{i*}(\mathbf{W}_1) - \mathbf{Z}^{(2)}_{i*}(\mathbf{W}'_1) \right\Vert_2 \\
    = & \left( \sqrt{2} + 2c_X c^2_W \left\Vert g(\tilde{\mathbf{A}}) \right\Vert^2_\infty \right) \left\Vert \sum_{j=1}^n \left[g(\tilde{\mathbf{A}})\right]_{ij} \left( \mathbf{H}^{(1)}_{j*}(\mathbf{W}_1) - \mathbf{H}^{(1)}_{j*}(\mathbf{W}'_1) \right) \right\Vert_2 \\
    \leq & \left( \sqrt{2} \left\Vert g(\tilde{\mathbf{A}}) \right\Vert_\infty + 2c_X c^2_W \left\Vert g(\tilde{\mathbf{A}}) \right\Vert^3_\infty \right) \left\Vert \mathbf{W}_1 - \mathbf{W}'_1 \right\Vert \mathop{\rm max}_{i\in [n]} \left\Vert \mathbf{Z}^{(1)}_{i*} \right\Vert_2 \\
    \leq & \left( \sqrt{2} c_X \left\Vert g(\tilde{\mathbf{A}}) \right\Vert^2_\infty + 2c_X c^2_W \left\Vert g(\tilde{\mathbf{A}}) \right\Vert^4_\infty \right) \left\Vert {\rm vec}\left[\mathbf{W}_1\right] - {\rm vec}\left[\mathbf{W}'_1\right] \right\Vert. 
\end{aligned}
\end{equation*}
Besides,
\begin{equation*}
\begin{aligned}
    & \left\Vert \frac{\partial \ell(\mathbf{W}_1, \mathbf{W}_2; z_i)}{\partial {\rm vec}\left[\mathbf{W}_2\right]}(\mathbf{W}_2) - \frac{\partial \ell(\mathbf{W}_1, \mathbf{W}_2; z_i)}{\partial {\rm vec}\left[\mathbf{W}_2\right]}(\mathbf{W}'_2) \right\Vert_2 \\
    \leq & \Vert \hat{\mathbf{y}}_i (\mathbf{W}_2) - \hat{\mathbf{y}}_i (\mathbf{W}_2) \Vert_2 \mathop{\rm max}_{i\in [n]} \left\Vert \mathbf{Z}^{(2)}_{i*} \right\Vert_2 \leq 2 \left\Vert \mathbf{Z}^{(1)} (\mathbf{W}_2 - \mathbf{W}'_2) \right\Vert_2 \mathop{\rm max}_{i\in [n]} \left\Vert \mathbf{Z}^{(2)}_{i*} \right\Vert_2 \\
    \leq & 2 \left\Vert \mathbf{W}_2 - \mathbf{W}'_2 \right\Vert \mathop{\rm max}_{i\in [n]} \left\Vert \mathbf{Z}^{(2)}_{i*} \right\Vert_2 \leq 2c_X c_W \Vert g(\mathbf{A}) \Vert^2_\infty \left\Vert {\rm vec}\left[\mathbf{W}_2\right] - {\rm vec}\left[\mathbf{W}'_2\right] \right\Vert.
\end{aligned}
\end{equation*}

Denote by $P_{21} = \sqrt{2} c_X \left\Vert g(\tilde{\mathbf{A}}) \right\Vert^2_\infty + 2c_X c^2_W \left\Vert g(\tilde{\mathbf{A}}) \right\Vert^4_\infty $, $P_{22} = 2c_X c_W \Vert g(\mathbf{A}) \Vert^2_\infty$, $\tilde{P}_{21} = \tilde{P}_{22}=0$, we obtain that $\left\Vert \frac{\partial \ell(\mathbf{w}; z_i)}{\partial {\rm vec}\left[\mathbf{W}_2\right]} - \frac{\partial \ell(\mathbf{w}'; z_i)}{\partial {\rm vec}\left[\mathbf{W}_2\right]} \right\Vert_2 \leq \sum_{i=1}^2 P_{2i} \left\Vert {\rm vec}\left[\mathbf{W}_i\right] - {\rm vec}\left[\mathbf{W}'_i\right] \right\Vert_2 + \tilde{P}_{2i} \left\Vert {\rm vec}\left[\mathbf{W}_i\right] - {\rm vec}\left[\mathbf{W}'_i\right] \right\Vert^{\alpha_{2i}}_2$. Then we analyze how $\frac{\partial \ell(\mathbf{W}_1, \mathbf{W}_2; z_i)}{\partial {\rm vec}\left[\mathbf{W}_1\right]}$ change w.r.t. $\mathbf{W}_1$ and $\mathbf{W}_2$.
Note that
\begin{equation*}
\begin{aligned}
    & \left\Vert \frac{\partial \ell(\mathbf{W}_1, \mathbf{W}_2; z_i)}{\partial {\rm vec}\left[\mathbf{W}_1\right]}(\mathbf{W}_1) - \frac{\partial \ell(\mathbf{W}_1, \mathbf{W}_2; z_i)}{\partial {\rm vec}\left[\mathbf{W}_1\right]}(\mathbf{W}'_1) \right\Vert_2 \\
    = & \left\Vert \sum_{j=1}^n \left[ g(\tilde{\mathbf{A}}) \right]_{ij} \left( \left( \sigma'\big(\mathbf{Z}^{(1)}_{j*} \mathbf{W}_1 \big) - \sigma'\big(\mathbf{Z}^{(1)}_{j*} \mathbf{W}'_1 \big) \right) \odot (\hat{\mathbf{y}}_i - \mathbf{y}_i)\mathbf{W}^\top_2 \right) \otimes \mathbf{Z}^{(1)}_{j*} \right\Vert_2 \\
    & + \left\Vert \sum_{j=1}^n \left[ g(\tilde{\mathbf{A}}) \right]_{ij} \left( \sigma' \big(\mathbf{Z}^{(1)}_{j*} \mathbf{W}_1 \big) \odot ((\hat{\mathbf{y}}_i(\mathbf{W}_1) - \hat{\mathbf{y}}_i(\mathbf{W}'_1) ))\mathbf{W}^\top_2 \right) \otimes \mathbf{Z}^{(1)}_{j*} \right\Vert_2 \\
    \leq & \left\Vert g(\tilde{\mathbf{A}}) \right\Vert_\infty \left\Vert  (\hat{\mathbf{y}}_i - \mathbf{y}_i)\mathbf{W}^\top_2 \right\Vert_2 \mathop{\rm max}_{j \in [n]} \left\Vert \mathbf{Z}^{(1)}_{j*} \right\Vert_2 \left\Vert \sigma'\big(\mathbf{Z}^{(1)}_{j*} \mathbf{W}_1 \big) - \sigma'\big(\mathbf{Z}^{(1)}_{j*} \mathbf{W}'_1 \big) \right\Vert_2 \\
    & + \left\Vert g(\tilde{\mathbf{A}}) \right\Vert_\infty \mathop{\rm max}_{j \in [n]} \left\Vert \mathbf{Z}^{(1)}_{j*} \right\Vert_2 \left\Vert \hat{\mathbf{y}}_i(\mathbf{W}_1) - \hat{\mathbf{y}}_i(\mathbf{W}'_1) \right\Vert_2 \\
    \leq & c_X c_W P \sqrt{|\mathcal{Y}|} \left\Vert g(\tilde{\mathbf{A}}) \right\Vert^2_\infty \mathop{\rm max}_{j \in [n]} \left\Vert \mathbf{Z}^{(1)}_{j*} \left( \mathbf{W}_1 - \mathbf{W}'_1 \right) \right\Vert^{\tilde{\alpha}}_2 + c_X c_W \left\Vert g(\tilde{\mathbf{A}}) \right\Vert^2_\infty \mathop{\rm max}_{j \in [n]} \left\Vert \mathbf{Z}^{(2)}_{i*}(\mathbf{W}_1) - \mathbf{Z}^{(2)}_{i*}(\mathbf{W}'_1) \right\Vert_2\\
    \leq & c_X c_W P \sqrt{|\mathcal{Y}|} \left\Vert g(\tilde{\mathbf{A}}) \right\Vert^2_\infty \left\Vert \mathbf{W}_1 - \mathbf{W}'_1 \right\Vert^{\tilde{\alpha}}_2 \mathop{\rm max}_{j \in [n]} \left\Vert \mathbf{Z}^{(1)}_{j*} \right\Vert_2 + c_X c_W \left\Vert g(\tilde{\mathbf{A}}) \right\Vert^2_\infty \mathop{\rm max}_{j \in [n]} \left\Vert \mathbf{Z}^{(2)}_{i*}(\mathbf{W}_1) - \mathbf{Z}^{(2)}_{i*}(\mathbf{W}'_1) \right\Vert_2\\
    \leq & c^{1+\tilde{\alpha}}_X c_W P \sqrt{|\mathcal{Y}|} \left\Vert g(\tilde{\mathbf{A}}) \right\Vert^{2+\tilde{\alpha}}_\infty \left\Vert {\rm vec}\left[\mathbf{W}_1\right] - {\rm vec}\left[\mathbf{W}'_1\right] \right\Vert^{\tilde{\alpha}}_2 + c^2_X c_W \left\Vert g(\tilde{\mathbf{A}}) \right\Vert^4_\infty \left\Vert {\rm vec}\left[\mathbf{W}_1\right] - {\rm vec}\left[\mathbf{W}'_1\right] \right\Vert_2,
\end{aligned}
\end{equation*}
where we use the fact that the absolute value of each element of $\sigma'\big(\mathbf{Z}^{(1)}_{j*} \mathbf{W}_1 \big)$ is less than $1$. Similarly,
\begin{equation*}
\begin{aligned}
    & \left\Vert \frac{\partial \ell(\mathbf{W}_1, \mathbf{W}_2; z_i)}{\partial {\rm vec}\left[\mathbf{W}_1\right]}(\mathbf{W}_2) - \frac{\partial \ell(\mathbf{W}_1, \mathbf{W}_2; z_i)}{\partial {\rm vec}\left[\mathbf{W}_2\right]}(\mathbf{W}'_2) \right\Vert_2 \\
    = & \left\Vert \sum_{j=1}^n \left[ g(\tilde{\mathbf{A}}) \right]_{ij} \left( \sigma'\left(\mathbf{Z}^{(1)}_{j*} \mathbf{W}_1 \right) \odot (\hat{\mathbf{y}}_i - \mathbf{y}_i) \left(\mathbf{W}_2 - \mathbf{W}'_2 \right)^\top \right) \otimes \mathbf{Z}^{(1)}_{j*} \right\Vert_2 \\
    & + \left\Vert \sum_{j=1}^n \left[ g(\tilde{\mathbf{A}}) \right]_{ij} \left( \sigma' \big(\mathbf{Z}^{(1)}_{j*} \mathbf{W}_1 \big) \odot ((\hat{\mathbf{y}}_i(\mathbf{W}_2) - \hat{\mathbf{y}}_i(\mathbf{W}'_2) ))\mathbf{W}^\top_2 \right) \otimes \mathbf{Z}^{(1)}_{j*} \right\Vert_2 \\
    \leq & \left\Vert g(\tilde{\mathbf{A}}) \right\Vert_\infty \left\Vert  (\hat{\mathbf{y}}_i - \mathbf{y}_i)\left(\mathbf{W}_2 - \mathbf{W}'_2 \right)^\top \right\Vert_2 \mathop{\rm max}_{j \in [n]} \left\Vert \mathbf{Z}^{(1)}_{j*} \right\Vert_2 + \left\Vert g(\tilde{\mathbf{A}}) \right\Vert_\infty \mathop{\rm max}_{j \in [n]} \left\Vert \mathbf{Z}^{(1)}_{j*} \right\Vert_2 \left\Vert \hat{\mathbf{y}}_i(\mathbf{W}_2) - \hat{\mathbf{y}}_i(\mathbf{W}'_2) \right\Vert_2\\
    \leq & \left( \sqrt{2} c_X \left\Vert g(\tilde{\mathbf{A}}) \right\Vert^{2}_\infty + 2 c^2_X c_W \left\Vert g(\tilde{\mathbf{A}}) \right\Vert^4_\infty \right) \left\Vert {\rm vec}\left[\mathbf{W}_2\right] - {\rm vec}\left[\mathbf{W}'_2\right] \right\Vert_2.
\end{aligned}
\end{equation*}
Denote by
\begin{equation*}
\begin{aligned}
    & P_{11} = c^{1+\tilde{\alpha}}_X c_W P \sqrt{|\mathcal{Y}|} \left\Vert g(\tilde{\mathbf{A}}) \right\Vert^{2+\tilde{\alpha}}_\infty, \tilde{P}_{11} = c^2_X c_W \left\Vert g(\tilde{\mathbf{A}}) \right\Vert^4_\infty, \\
    & P_{12} = \sqrt{2} c_X \left\Vert g(\tilde{\mathbf{A}}) \right\Vert^{2}_\infty + 2 c^2_X c_W \left\Vert g(\tilde{\mathbf{A}}) \right\Vert^4_\infty, \tilde{P}_{12} = 0,
\end{aligned}
\end{equation*}
we obtain $\left\Vert \frac{\partial \ell(\mathbf{w}; z_i)}{\partial {\rm vec}\left[\mathbf{W}_1\right]} - \frac{\partial \ell(\mathbf{w}'; z_i)}{\partial {\rm vec}\left[\mathbf{W}_1\right]} \right\Vert_2 \leq \sum_{i=1}^2 P_{1i} \left\Vert {\rm vec}\left[\mathbf{W}_i\right] - {\rm vec}\left[\mathbf{W}'_i\right] \right\Vert_2 + \tilde{P}_{1i} \left\Vert {\rm vec}\left[\mathbf{W}_i\right] - {\rm vec}\left[\mathbf{W}'_i\right] \right\Vert^{\tilde{\alpha}}_2 $. By Lemma~\ref{bound_alpha}, we conclude that $\Vert \nabla \ell(\mathbf{w}) - \nabla \ell(\mathbf{w}') \Vert_2 \leq P_{\mathcal{F}} \mathop{\rm max} \{ \Vert \mathbf{w} - \mathbf{w}' \Vert_2, \Vert \mathbf{w} - \mathbf{w}' \Vert^{\tilde{\alpha}}_2 \}$ where $\mathbf{w}=\left[{\rm vec}\left[ \mathbf{W}_1 \right]; {\rm vec}\left[ \mathbf{W}_2 \right]\right]$.

\subsubsection{Proof of Proposition~\ref{gcnii}}
We first analyze the Lipschitz continuity. Denote by 
\begin{equation*}
\begin{aligned}
    \mathbf{H}^{(0)} & = \sigma\left( \mathbf{X}\mathbf{W}_0 \right), \\
    \mathbf{H}^{(1)} & = \sigma \left( ( (1-\alpha_1) g(\tilde{\mathbf{A}}) \mathbf{H}^{(0)} + \alpha_1 \mathbf{H}^{(0)} ) ( (1-\beta_1) \mathbf{I} + \beta_1 \mathbf{W}_1 ) \right), \\
    \mathbf{H}^{(2)} & = \sigma \left( ( (1-\alpha_2) g(\tilde{\mathbf{A}}) \mathbf{H}^{(1)} + \alpha_2 \mathbf{H}^{(0)} ) ( (1-\beta_2) \mathbf{I} + \beta_2 \mathbf{W}_2 ) \right), \\
\end{aligned}
\end{equation*}
the forward process of GCNII is given by $\hat{\mathbf{Y}}={\rm Softmax} \left( \mathbf{H}^{(2)}\mathbf{W}_3 \right)$. First, we have
\begin{equation*}
    \mathop{\rm max}_{i \in [n]} \left\Vert \mathbf{H}^{(0)}_{i*} \right\Vert_2 = \mathop{\rm max}_{i \in [n]} \left\Vert \sigma \left( \mathbf{X}_{i*}\mathbf{W}_0 \right) \right\Vert_2 \leq c_X c_W.
\end{equation*}
Similarly, for $\ell = 1$ and $\ell = 2$, denote by $C_{\ell} = (1 - \beta_{\ell}) + \beta_{\ell} c_W$, we have
\begin{equation*}
\begin{aligned}
    & \mathop{\rm max}_{i \in [n]} \left\Vert \mathbf{H}^{(\ell)}_{i*} \right\Vert_2 \\
    = & \mathop{\rm max}_{i \in [n]} \left\Vert \sigma\left( \sum_{j=1}^n ((1-\alpha_\ell) \left[ g(\tilde{\mathbf{A}}) \right]_{ij} \mathbf{H}^{(\ell-1)}_{j*} + \alpha_\ell \mathbf{H}^{(0)}_{i*})((1-\beta_\ell)\mathbf{I}+\beta_\ell \mathbf{W}_\ell) \right) \right\Vert_2 \\
    \leq & \mathop{\rm max}_{i \in [n]} \left\{ (1-\alpha_\ell) \left\Vert \sum_{j=1}^n \left[ g(\tilde{\mathbf{A}}) \right]_{ij} \mathbf{H}^{(\ell-1)}_{j*} ((1-\beta_\ell)\mathbf{I}+\beta_\ell \mathbf{W}_\ell) \right\Vert_2 + \alpha_\ell \left\Vert \mathbf{H}^{(0)}_{i*} ((1-\beta_\ell)\mathbf{I}+\beta_\ell \mathbf{W}_\ell) \right\Vert_2 \right\} \\
    \leq & \mathop{\rm max}_{i \in [n]} \left\{ (1-\alpha_\ell) \sum_{j=1}^n \left[ g(\tilde{\mathbf{A}}) \right]_{ij} \left\Vert \mathbf{H}^{(\ell-1)}_{j*} ((1-\beta_\ell)\mathbf{I}+\beta_\ell \mathbf{W}_\ell) \right\Vert_2 + \alpha_\ell \left\Vert \mathbf{H}^{(0)}_{i*} ((1-\beta_\ell)\mathbf{I}+\beta_\ell \mathbf{W}_\ell) \right\Vert_2 \right\} \\
    \leq & (1-\alpha_\ell) \left\Vert g(\tilde{\mathbf{A}}) \right\Vert_\infty \left\Vert (1-\beta_\ell)\mathbf{I}+\beta_\ell \mathbf{W}_\ell \right\Vert_2 \mathop{\rm max}_{i \in [n]} \left\Vert \mathbf{H}^{(\ell-1)}_{i*} \right\Vert_2  + \alpha_\ell \left\Vert (1-\beta_\ell)\mathbf{I}+\beta_\ell \mathbf{W}_\ell \right\Vert_2 \mathop{\rm max}_{i \in [n]} \left\Vert \mathbf{H}^{(0)}_{i*} \right\Vert_2 \\
    \leq & (1-\alpha_\ell) C_{\ell} \left\Vert g(\tilde{\mathbf{A}}) \right\Vert_\infty \mathop{\rm max}_{i \in [n]} \left\Vert \mathbf{H}^{(\ell-1)}_{i*} \right\Vert + \alpha_\ell c_X c_W C_{\ell}.
\end{aligned}
\end{equation*}
Let $B_1 = \mathop{\rm max}_{i \in [n]} \left\Vert \mathbf{H}^{(1)}_{i*} \right\Vert_2 $ and $B_2 = \mathop{\rm max}_{i \in [n]} \left\Vert \mathbf{H}^{(2)}_{i*} \right\Vert_2 $, we have obtain $B_1 = c_X c_W C_1 ((1-\alpha_1) \left\Vert g(\tilde{\mathbf{A}}) \right\Vert_\infty + \alpha_1 )$ and $B_2 = (1-\alpha_2) C_2 \left\Vert g(\tilde{\mathbf{A}}) \right\Vert_\infty B_1  + \alpha_2 c_X c_W C_2$. Next, we analyze the change of $\mathbf{H}^{(1)}$ and $\mathbf{H}^{(2)}$ w.r.t. $\mathbf{W}_0$, $\mathbf{W}_1$ and $\mathbf{W}_2$:

\textbf{Part A.} Note that
\begin{equation*}
\begin{aligned}
    \Delta_{11} \triangleq & \left\Vert \mathbf{H}^{(1)}_{i*}(\mathbf{W}_1) - \mathbf{H}^{(1)}_{i*}(\mathbf{W}'_1) \right\Vert_2 \\
    \leq & \beta_1 \left\Vert \Big( (1-\alpha_1)\sum_{j=1}^n {\left[g(\tilde{\mathbf{A}})\right]_{ij} \mathbf{H}^{(0)}_{j*}} + \alpha_1 \mathbf{H}^{(0)}_{i*} \Big) ( \mathbf{W}_1 - \mathbf{W}'_1 ) \right\Vert_2 \\
    \leq & \beta_1 \left( (1-\alpha_1) \sum_{j=1}^n {\left[ g(\tilde{\mathbf{A}}) \right]_{ij} \left\Vert \mathbf{H}^{(0)}_{j*} \right\Vert_2} + \alpha_1 \left\Vert \mathbf{H}^{(0)}_{i*} \right\Vert_2 \right) \left\Vert \mathbf{W}_1 - \mathbf{W}'_1 \right\Vert \\
    \leq & \beta_1 \left( (1-\alpha_1) \left\Vert g(\tilde{\mathbf{A}}) \right\Vert_\infty + \alpha_1 \right) c_X c_W \left\Vert \mathbf{W}_1 - \mathbf{W}'_1 \right\Vert \\
    \leq & \beta_1 \left( (1-\alpha_1) \left\Vert g(\tilde{\mathbf{A}}) \right\Vert_\infty + \alpha_1 \right) c_X c_W \left\Vert {\rm vec}\left[\mathbf{W}_1 \right] - {\rm vec}\left[\mathbf{W}'_1 \right] \right\Vert_2 = \frac{\beta_1 B_1}{C_1} \left\Vert {\rm vec}\left[\mathbf{W}_1 \right] - {\rm vec}\left[\mathbf{W}'_1 \right] \right\Vert_2 .
\end{aligned}
\end{equation*}
Similarly, we have
\begin{equation*}
\begin{aligned}
    \Delta_{10} \triangleq & \left\Vert \mathbf{H}^{(1)}_{i*}(\mathbf{W}_0) - \mathbf{H}^{(1)}_{i*}(\mathbf{W}'_0) \right\Vert_2 \\
    \leq & \left\Vert (1-\alpha_1)\sum_{j=1}^n \left[g(\tilde{\mathbf{A}})\right]_{ij} {(\mathbf{H}^{(0)}_{j*}(\mathbf{W}_0) - \mathbf{H}^{(0)}_{j*}(\mathbf{W}'_0))} + \alpha_1 (\mathbf{H}^{(0)}_{i*}(\mathbf{W}_0) - \mathbf{H}^{(0)}_{i*}(\mathbf{W}'_0)) \right\Vert_2 \left\Vert (1-\beta_1)\mathbf{I} + \beta_1 \mathbf{W}_1 \right\Vert \\ 
    \leq & C_1 \left( (1-\alpha_1)\sum_{j=1}^n \left[g(\tilde{\mathbf{A}})\right]_{ij} {\left\Vert \mathbf{H}^{(0)}_{j*}(\mathbf{W}_0) - \mathbf{H}^{(0)}_{j*}(\mathbf{W}'_0) \right\Vert_2} + \alpha_1 \left\Vert \mathbf{H}^{(0)}_{i*}(\mathbf{W}_0) - \mathbf{H}^{(0)}_{i*}(\mathbf{W}'_0) \right\Vert_2 \right) \\
    \leq & C_1 \left( (1-\alpha_1) \left\Vert g(\tilde{\mathbf{A}}) \right\Vert_\infty + \alpha_1 \right) c_X \left\Vert \mathbf{W}_0 - \mathbf{W}'_0 \right\Vert = \frac{B_1}{c_W} \left\Vert \mathbf{W}_0 - \mathbf{W}'_0 \right\Vert.
\end{aligned}
\end{equation*}

\textbf{Part B.}
Note that
\begin{equation*}
\begin{aligned}
    \Delta_{22} \triangleq &  \left\Vert \mathbf{H}^{(2)}_{i*}(\mathbf{W}_2) - \mathbf{H}^{(2)}_{i*}(\mathbf{W}'_2) \right\Vert_2 \\
    \leq & \beta_2 \left\Vert \Big( (1-\alpha_2)\sum_{j=1}^n {\left[g(\tilde{\mathbf{A}})\right]_{ij} \mathbf{H}^{(1)}_{j*}} + \alpha_2 \mathbf{H}^{(0)}_{i*} \Big) ( \mathbf{W}_2 - \mathbf{W}'_2 ) \right\Vert_2 \\
    \leq & \beta_2 \left( (1-\alpha_2) \sum_{j=1}^n \left[g(\tilde{\mathbf{A}})\right]_{ij} \left\Vert \mathbf{H}^{(1)}_{j*} \right\Vert_2 + \alpha_2 \left\Vert \mathbf{H}^{(0)}_{i*} \right\Vert_2 \right) \left\Vert \mathbf{W}_2 - \mathbf{W}'_2 \right\Vert \\
    \leq & \beta_2 \left( (1-\alpha_2) \left\Vert g(\tilde{\mathbf{A}}) \right\Vert_\infty B_1 + \alpha_2 c_X c_W \right) \left\Vert \mathbf{W}_2 - \mathbf{W}'_2 \right\Vert \\
    \leq & \beta_2 \left( (1-\alpha_2) \left\Vert g(\tilde{\mathbf{A}}) \right\Vert_\infty B_1 + \alpha_2 c_X c_W \right) \left\Vert {\rm vec}\left[\mathbf{W}_2\right] - {\rm vec}\left[\mathbf{W}'_2\right] \right\Vert_2 = \frac{\beta_2 B_2}{C_2} \left\Vert {\rm vec}\left[\mathbf{W}_1 \right] - {\rm vec}\left[\mathbf{W}'_1 \right] \right\Vert_2.
\end{aligned}
\end{equation*}
Similarly,
\begin{equation*}
\begin{aligned}
    \Delta_{21} \triangleq & \left\Vert \mathbf{H}^{(2)}_{i*}(\mathbf{W}_1) - \mathbf{H}^{(2)}_{i*}(\mathbf{W}'_1) \right\Vert_2 \\
    \leq & (1-\alpha_2) \sum_{j=1}^n {\left[g(\tilde{\mathbf{A}})\right]_{ij} \left\Vert \mathbf{H}^{(1)}_{j*}(\mathbf{W}_1) - \mathbf{H}^{(1)}_{j*}(\mathbf{W}'_1) \right\Vert_2} \left\Vert (1-\beta_2)\mathbf{I} + \beta_2 \mathbf{W}_2 \right\Vert_2 \\
    \leq & (1-\alpha_2) C_2 \left\Vert g(\tilde{\mathbf{A}}) \right\Vert_\infty \mathop{\rm max}_{i \in [n]} \left\Vert \mathbf{H}^{(1)}_{i*}(\mathbf{W}_1) - \mathbf{H}^{(1)}_{i*}(\mathbf{W}'_1) \right\Vert_2 \leq (1-\alpha_2) \beta_1 \frac{B_1 C_2}{C_1} \left\Vert g(\tilde{\mathbf{A}}) \right\Vert_\infty \left\Vert {\rm vec}\left[\mathbf{W}_1\right] - {\rm vec}\left[\mathbf{W}'_1\right] \right\Vert_2.
\end{aligned}
\end{equation*}
Besides,
\begin{equation*}
\begin{aligned}
    \Delta_{20} \triangleq & \left\Vert \mathbf{H}^{(2)}_{i*}(\mathbf{W}_0) - \mathbf{H}^{(2)}_{i*}(\mathbf{W}'_0) \right\Vert_2 \\
    \leq & (1-\alpha_2) \sum_{j=1}^n {\left[g(\tilde{\mathbf{A}})\right]_{ij} \left\Vert \mathbf{H}^{(1)}_{j*}(\mathbf{W}_0) - \mathbf{H}^{(1)}_{j*}(\mathbf{W}'_0) \right\Vert_2} \left\Vert (1-\beta_2)\mathbf{I} + \beta_2 \mathbf{W}_2 \right\Vert_2 \\
    & + \alpha_2 \left\Vert \mathbf{H}^{(0)}_{j*}(\mathbf{W}_0) - \mathbf{H}^{(0)}_{j*}(\mathbf{W}'_0) \right\Vert_2 \left\Vert (1-\beta_2)\mathbf{I} + \beta_2 \mathbf{W}_2 \right\Vert_2 \\
    \leq & \left( (1-\alpha_2) \frac{B_1 C_2}{c_W} \left\Vert g(\tilde{\mathbf{A}}) \right\Vert_\infty + \alpha_2 c_X C_2 \right) \left\Vert {\rm vec}\left[\mathbf{W}_0\right] - {\rm vec}\left[\mathbf{W}'_0\right] \right\Vert_2 = \frac{B_2}{c_W} \left\Vert {\rm vec}\left[\mathbf{W}_0\right] - {\rm vec}\left[\mathbf{W}'_0\right] \right\Vert_2.
\end{aligned}
\end{equation*}

Now we are ready to analyze the Lipschitz continuity and Holder smoothness. Note that
\begin{equation*}
\begin{aligned}
    & |\ell(\mathbf{W}_0, \mathbf{W}_1, \mathbf{W}_2, \mathbf{W}_3;z_i) - \ell(\mathbf{W}_0, \mathbf{W}_1, \mathbf{W}_2, \mathbf{W}'_3;z_i)| \\
    \leq & \sqrt{2} \left\Vert \mathbf{H}^{(2)}_{i*} (\mathbf{W}_3 - \mathbf{W}'_3) \right\Vert_2 \leq \sqrt{2} \mathop{\rm max}_{i \in [n]} \left\Vert \mathbf{H}^{(2)}_{i*} \right\Vert_2 \left\Vert \mathbf{W}_3 - \mathbf{W}'_3 \right\Vert_2 \\
    \leq & \sqrt{2} B_2 \left\Vert \mathbf{W}_3 - \mathbf{W}'_3 \right\Vert_2 \leq  \sqrt{2} B_2 \left\Vert {\rm vec}\left[\mathbf{W}_3\right] - {\rm vec}\left[\mathbf{W}'_3\right] \right\Vert_2.
\end{aligned}
\end{equation*}
Since $\mathbf{H}^{(2)}$ is a variable related to $\mathbf{W}_2$, we have
\begin{equation*}
\begin{aligned}
    & |\ell(\mathbf{W}_0, \mathbf{W}_1, \mathbf{W}_2, \mathbf{W}_3;z_i) - \ell(\mathbf{W}_0, \mathbf{W}_1, \mathbf{W}'_2, \mathbf{W}_3;z_i)| \\
    \leq & \sqrt{2} \left\Vert (\mathbf{H}^{(2)}_{i*}(\mathbf{W}_2) - \mathbf{H}^{(2)}_{i*}(\mathbf{W}'_2)) \mathbf{W}_3 \right\Vert_2 \\
    \leq & c_W \sqrt{2} \Delta_{22} = \frac{\beta_2 B_2}{C_2} c_W \sqrt{2} \left\Vert {\rm vec}\left[\mathbf{W}_2\right] - {\rm vec}\left[\mathbf{W}'_2\right] \right\Vert_2.
\end{aligned}
\end{equation*}
Similarly, since $\mathbf{H}^{(1)}$ and $\mathbf{H}^{(2)}$ are variables related to $\mathbf{W}_1$,
\begin{equation*}
\begin{aligned}
    & |\ell(\mathbf{W}_0, \mathbf{W}_1, \mathbf{W}_2, \mathbf{W}_3;z_i) - \ell(\mathbf{W}_0, \mathbf{W}'_1, \mathbf{W}_2, \mathbf{W}_3;z_i)| \\
    \leq & \sqrt{2} \left\Vert (\mathbf{H}^{(2)}_{i*}(\mathbf{W}_1) - \mathbf{H}^{(2)}_{i*}(\mathbf{W}'_1)) \mathbf{W}_3 \right\Vert_2 \\
    \leq & c_W \sqrt{2} \Delta_{21} = (1-\alpha_2) \beta_1 c_W \frac{B_1 C_2}{C_1} \left\Vert g(\tilde{\mathbf{A}}) \right\Vert_\infty \sqrt{2} \left\Vert {\rm vec}\left[\mathbf{W}_1\right] - {\rm vec}\left[\mathbf{W}'_1\right] \right\Vert_2.
\end{aligned}
\end{equation*}
Lastly, since $\mathbf{H}^{(0)}$, $\mathbf{H}^{(1)}$ and $\mathbf{H}^{(2)}$ are variables related to $\mathbf{W}_0$,
\begin{equation*}
\begin{aligned}
    & |\ell(\mathbf{W}_0, \mathbf{W}_1, \mathbf{W}_2, \mathbf{W}_3;z_i) - \ell(\mathbf{W}'_0, \mathbf{W}_1, \mathbf{W}_2, \mathbf{W}_3;z_i)| \\
    \leq & \sqrt{2} \left\Vert (\mathbf{H}^{(2)}_{i*}(\mathbf{W}_0) - \mathbf{H}^{(2)}_{i*}(\mathbf{W}'_0)) \mathbf{W}_3 \right\Vert_2 \\
    \leq & c_W \sqrt{2} \Delta_{20} = \sqrt{2} B_2 \left\Vert {\rm vec}\left[\mathbf{W}_0\right] - {\rm vec}\left[\mathbf{W}'_0\right] \right\Vert_2.
\end{aligned}
\end{equation*}
Denote by
\begin{equation*}
\begin{aligned}
    L_{\mathcal{F}} = \sqrt{4B^2_2 + 2 c^2_W \frac{\beta^2_2 B^2_2}{C^2_2} + 2(1-\alpha_2)^2 \beta^2_1 c^2_W \frac{B^2_1 C^2_2}{C^2_1} \left\Vert g(\tilde{\mathbf{A}}) \right\Vert^2_\infty}, \\
\end{aligned}
\end{equation*}
by Lemma~\ref{bound_alpha}, we conclude that $\vert \ell(\mathbf{w}) - \ell(\mathbf{w}') \vert \leq L_{\mathcal{F}} \Vert \mathbf{w} - \mathbf{w}' \Vert_2$ holds. Then we discuss the smoothness. By the chain rule, we have
\begin{equation*}
\begin{aligned}
    & \frac{\partial \ell(\mathbf{W}_0, \mathbf{W}_1, \mathbf{W}_2, \mathbf{W}_3;z_i)}{\partial {\rm vec} \left[ \mathbf{W}_3 \right]} = (\hat{\mathbf{y}}_i - \mathbf{y}_i) \otimes \mathbf{H}^{(2)}_{i*}, \\
    & \frac{\partial \ell(\mathbf{W}_0, \mathbf{W}_1, \mathbf{W}_2, \mathbf{W}_3;z_i)}{\partial {\rm vec} \left[ \mathbf{W}_2 \right]} = \alpha_2 \beta_2 \delta_i \otimes \mathbf{H}^{(0)}_{i*} + (1-\alpha_2)\beta_2 \sum_{j=1}^n \left[ g(\tilde{\mathbf{A}}) \right]_{ij} \delta_i \otimes \mathbf{H}^{(1)}_{j*}, \\
    & \frac{\partial \ell(\mathbf{W}_0, \mathbf{W}_1, \mathbf{W}_2, \mathbf{W}_3;z_i)}{\partial {\rm vec} \left[ \mathbf{W}_1 \right]} = \alpha_1 \beta_1 \sum_{j=1}^n  \left[ g(\tilde{\mathbf{A}}) \right]_{ij} \delta_{ij} \otimes \mathbf{H}^{(0)}_{j*} + (1-\alpha_1)\beta_1 \sum_{j=1}^n \sum_{k=1}^n \left[ g(\tilde{A}) \right]_{ij} \left[ g(\tilde{A}) \right]_{jk} \delta_{ij} \otimes \mathbf{H}^{(0)}_{k*}, \\
    & \frac{\partial \ell(\mathbf{W}_0, \mathbf{W}_1, \mathbf{W}_2, \mathbf{W}_3;z_i)}{\partial {\rm vec} \left[ \mathbf{W}_0 \right]} \\
    = & \alpha_2 ((\delta_i ((1-\beta_2)\mathbf{I}+\beta_2 \mathbf{W}^{\top}_2)) \odot \mathbf{H}^{(0)}_{i*}) \otimes \mathbf{X}_{i*} \\
    & + \alpha_1 \sum_{j=1}^n  \left[g(\tilde{\mathbf{A}})\right]_{ij} ((\delta_{ij} ((1-\beta_1)\mathbf{I}+\beta_1\mathbf{W}_1^{\top})) \odot \mathbf{H}^{(0)}_{j*}) \otimes \mathbf{X}_{j*} \\
    & + (1-\alpha_1) \sum_{j=1}^n \sum_{k=1}^n \left[ g(\tilde{\mathbf{A}})\right]_{ij} \left[ g(\tilde{\mathbf{A}})\right]_{jk} ((\delta_{ij} ((1-\beta_{1})\mathbf{I}+\beta_{1}\mathbf{W}_1^{\top})) \odot \mathbf{H}^{(0)}_{k*}) \otimes \mathbf{X}_{k*},
\end{aligned}
\end{equation*}
where
\begin{equation*}
\begin{aligned}
    \delta_i = & ((\hat{\mathbf{y}}_i - \mathbf{y}_i)\mathbf{W}^\top_3) \odot \sigma'(\mathbf{H}^{(2)}_{i*}), \\
    \delta_{ij} = & (1-\alpha_2)\sigma'(\mathbf{H}^{(1)}_{j*}) \odot (\delta_i ((1-\beta_2)\mathbf{W}^\top_2+\beta_2 \mathbf{I})). \\
\end{aligned}
\end{equation*}
We first analyze how $\delta_i$ and $\delta_{ij}$ change w.r.t. $\mathbf{W}_0$, $\mathbf{W}_1$, $\mathbf{W}_2$ and $\mathbf{W}_3$.

\textbf{Part C.} 
For $i \in [n]$, we have
\begin{equation}\label{delta13}
\begin{aligned}
    & \left\Vert \delta_i(\mathbf{W}_3) - \delta_i(\mathbf{W}'_3) \right\Vert_2 \\
    \leq & \left\Vert (\hat{\mathbf{y}}_i - \mathbf{y}_i)(\mathbf{W}_3 - \mathbf{W}'_3)^\top \odot \sigma'(\mathbf{H}^{(2)}_{i*}) \right\Vert_2 + \left\Vert (\hat{\mathbf{y}}_i(\mathbf{W}_3) - \hat{\mathbf{y}}_i(\mathbf{W}'_3))\mathbf{W}_3^{'\top} \odot \sigma'(\mathbf{H}^{(2)}_{i*}) \right\Vert_2 \\ 
    \leq & \left\Vert (\hat{\mathbf{y}}_i - \mathbf{y}_i) (\mathbf{W}_3 - \mathbf{W}'_3) \right\Vert_2 + c_W \left\Vert (\hat{\mathbf{y}}_i(\mathbf{W}_3) - \hat{\mathbf{y}}_i(\mathbf{W}'_3) \right\Vert_2 \\
    \leq & \sqrt{2} \left\Vert \mathbf{W}_3 - \mathbf{W}'_3 \right\Vert_2 +
    2 c_W \left\Vert \mathbf{W}_3 - \mathbf{W}'_3 \right\Vert_2 \mathop{\rm max}_{i\in [n]} \left\Vert \mathbf{H}^{(2)}_{i*} \right\Vert_2 = (\sqrt{2} + 2 c_W B_2) \left\Vert {\rm vec}\left[\mathbf{W}_3\right] - {\rm vec}\left[\mathbf{W}'_3\right] \right\Vert_2,
\end{aligned}
\end{equation}
where we use the fact that the absolute value of each component in $\sigma'(\mathbf{H}^{(2)}_{i*})$ is less than $1$. Similarly, we have
\begin{equation}\label{delta12}
\begin{aligned}
    & \left\Vert \delta_i(\mathbf{W}_2) - \delta_i(\mathbf{W}'_2) \right\Vert_2 \\
    \leq & \left\Vert (\hat{\mathbf{y}}_i - \mathbf{y}_i)\mathbf{W}^\top_3 \odot (\sigma'(\mathbf{H}^{(2)}_{i*})(\mathbf{W}_2) - \sigma'(\mathbf{H}^{(2)}_{i*})(\mathbf{W}'_2)) \right\Vert_2 + \left\Vert ((\hat{\mathbf{y}}_i(\mathbf{W}_2) - \hat{\mathbf{y}}_i(\mathbf{W}'_2))\mathbf{W}^{'\top}_3) \odot \sigma'(\mathbf{H}^{(2)}_{i*}) \right\Vert_2  \\ 
    \leq & P \left\Vert (\hat{\mathbf{y}}_i - \mathbf{y}_i) \mathbf{W}^{\top}_3 \right\Vert_2 \left\Vert \mathbf{H}^{(2)}_{i*}(\mathbf{W}_2) - \mathbf{H}^{(2)}_{i*}(\mathbf{W}'_2) \right\Vert^{\tilde{\alpha}}_2 + 2c^2_W \left\Vert \mathbf{H}^{(2)}_{i*}(\mathbf{W}_2) - \mathbf{H}^{(2)}_{i*}(\mathbf{W}'_2) \right\Vert_2 \\
    \leq & P \sqrt{2} c_W \Delta^{\tilde{\alpha}}_{22} + 2 c^2_W \Delta_{22} \\
    = & \sqrt{2} c_W P \left( \frac{\beta_2 B_2}{C_2} \right)^{\tilde{\alpha}} \left\Vert {\rm vec}\left[\mathbf{W}_2\right] - {\rm vec}\left[\mathbf{W}'_2\right] \right\Vert^{\tilde{\alpha}}_2 + 2 c^2_W \frac{\beta_2 B_2}{C_2} \left\Vert {\rm vec}\left[\mathbf{W}_2\right] - {\rm vec}\left[\mathbf{W}'_2\right] \right\Vert_2.
\end{aligned}
\end{equation}
Besides,
\begin{equation}\label{delta11}
\begin{aligned}
    & \left\Vert \delta_i(\mathbf{W}_1) - \delta_i(\mathbf{W}'_1) \right\Vert_2 \leq \sqrt{2} c_W P \Delta^{\tilde{\alpha}}_{21} + 2 c^2_W \Delta_{21} \\
    = & \sqrt{2} c_W P  \left( (1-\alpha_2) \beta_1 \frac{B_1 C_2}{C_1} \left\Vert g(\tilde{\mathbf{A}}) \right\Vert_\infty \right)^{\tilde{\alpha}} \left\Vert {\rm vec}\left[\mathbf{W}_1 \right] - {\rm vec}\left[\mathbf{W}'_1 \right] \right\Vert^{\tilde{\alpha}}_2 \\
    & + 2(1-\alpha_2) \beta_1 c^2_W \frac{B_1 C_2}{C_1} \left\Vert g(\tilde{\mathbf{A}}) \right\Vert_\infty \left\Vert {\rm vec}\left[\mathbf{W}_1 \right] - {\rm vec}\left[\mathbf{W}'_1 \right] \right\Vert_2, \\
    & \left\Vert \delta_i(\mathbf{W}_0) - \delta_i(\mathbf{W}'_0) \right\Vert_2 \leq \sqrt{2} c_W P \Delta^{\tilde{\alpha}}_{20} + 2c^2_W \Delta_{20} \\
    = & \sqrt{2} c_W P  \left( \frac{B_2}{c_W} \right)^{\tilde{\alpha}} \left\Vert {\rm vec}\left[\mathbf{W}_0 \right] - {\rm vec}\left[\mathbf{W}'_0 \right] \right\Vert^{\tilde{\alpha}}_2 + 2 c_W B_2 \left\Vert {\rm vec}\left[\mathbf{W}_0 \right] - {\rm vec}\left[\mathbf{W}'_0 \right] \right\Vert_2.
\end{aligned}
\end{equation}

\textbf{Part D.} For $i \in [n]$, we have
\begin{equation}\label{delta23}
\begin{aligned}
    & \left\Vert \delta_{ij}(\mathbf{W}_3) - \delta_{ij}(\mathbf{W}'_3) \right\Vert_2 \\
    = & (1-\alpha_2) \left\Vert \sigma'(\mathbf{H}^{(1)}_{j*}) \odot ((\delta_i(\mathbf{W}_3) - \delta_i(\mathbf{W}'_3)) ((1-\beta_2)\mathbf{W}^\top_2+\beta_2 \mathbf{I})) \right\Vert_2 \\ 
    \leq & (1-\alpha_2) \left\Vert (\delta_i(\mathbf{W}_3) - \delta_i(\mathbf{W}'_3)) ((1-\beta_2)\mathbf{W}^\top_2+\beta_2 \mathbf{I}) \right\Vert_2 \\
    \leq & (1-\alpha_2) C_2 \left\Vert \delta_i(\mathbf{W}_3) - \delta_i(\mathbf{W}'_3) \right\Vert_2 \leq (1-\alpha_2) C_2 (\sqrt{2} + 2 c_W B_2) \left\Vert {\rm vec}\left[\mathbf{W}_3 \right] - {\rm vec}\left[\mathbf{W}'_3 \right] \right\Vert_2.
\end{aligned}
\end{equation}
Similarly, we have
\begin{equation}\label{delta22}
\begin{aligned}
    & \left\Vert \delta_{ij}(\mathbf{W}_2) - \delta_{ij}(\mathbf{W}'_2) \right\Vert_2 \\
    \leq & (1-\alpha_2) \left\Vert \sigma'(\mathbf{H}^{(1)}_{j*}) \odot ((\delta_i(\mathbf{W}_2) - \delta_i(\mathbf{W}'_2)) ((1-\beta_2)\mathbf{W}^\top_2+\beta_2 \mathbf{I})) \right\Vert_2 \\
    & + (1-\alpha_2) (1-\beta_2) \left\Vert \sigma'(\mathbf{H}^{(1)}_{j*}) \odot ((\delta_i (\mathbf{W}_2 - \mathbf{W}'_2)^\top)) \right\Vert_2 \\ 
    \leq & (1-\alpha_2) \left\Vert (\delta_i(\mathbf{W}_2) - \delta_i(\mathbf{W}'_2)) ((1-\beta_2)\mathbf{W}^\top_2+\beta_2 \mathbf{I}) \right\Vert_2 + (1-\alpha_2) (1-\beta_2) \left\Vert \delta_i (\mathbf{W}_2 - \mathbf{W}'_2)^\top \right\Vert_2 \\
    \leq & (1-\alpha_2) C_2 \left\Vert (\delta_i(\mathbf{W}_2) - \delta_i(\mathbf{W}'_2)) \right\Vert_2 + (1-\alpha_2) (1-\beta_2) \left\Vert \delta_i \right\Vert_2 \left\Vert \mathbf{W}_2 - \mathbf{W}'_2 \right\Vert \\
    \leq & \sqrt{2}(1-\alpha_2) c_W P C_2 \left( \frac{\beta_2 B_2}{C_2} \right)^{\tilde{\alpha}} \left\Vert {\rm vec}\left[\mathbf{W}_2 \right] - {\rm vec}\left[\mathbf{W}'_2 \right] \right\Vert^{\tilde{\alpha}}_2 \\
    & + (1-\alpha_2) \left( 2c^2_W \beta_2 B_2 + \sqrt{2} (1-\beta_2) c_W \right) \left\Vert {\rm vec}\left[\mathbf{W}_2 \right] - {\rm vec}\left[\mathbf{W}'_2 \right] \right\Vert_2.
\end{aligned}
\end{equation}
Besides, 
\begin{equation}\label{delta21}
\begin{aligned}
    & \left\Vert \delta_{ij}(\mathbf{W}_1) - \delta_{ij}(\mathbf{W}'_1) \right\Vert_2 \\
    \leq & (1-\alpha_2) \left\Vert \sigma'(\mathbf{H}^{(1)}_{j*}) \odot ((\delta_i(\mathbf{W}_1) - \delta_i(\mathbf{W}'_1)) ((1-\beta_2)\mathbf{W}^\top_2+\beta_2 \mathbf{I})) \right\Vert_2 \\
    & + (1-\alpha_2) \left\Vert (\sigma'(\mathbf{H}^{(1)}_{j*})(\mathbf{W}_1) - \sigma'(\mathbf{H}^{(1)}_{j*})(\mathbf{W}'_1)) \odot (\delta_i ((1-\beta_2)\mathbf{W}^\top_2+\beta_2 \mathbf{I})) \right\Vert_2 \\ 
    \leq & (1-\alpha_2) C_2 \Vert \delta_i(\mathbf{W}_1) - \delta_i(\mathbf{W}'_1) \Vert_2 + (1-\alpha_2) C_2 \Vert \delta_i \Vert \left\Vert \sigma'(\mathbf{H}^{(1)}_{j*})(\mathbf{W}_1) - \sigma'(\mathbf{H}^{(1)}_{j*})(\mathbf{W}'_1) \right\Vert_2 \\
    \leq & (1-\alpha_2) C_2 \Vert \delta_i(\mathbf{W}_1) - \delta_i(\mathbf{W}'_1) \Vert_2 + \sqrt{2} (1-\alpha_2) c_W C_2 P \left\Vert \mathbf{H}^{(1)}_{j*} (\mathbf{W}_1) - \mathbf{H}^{(1)}_{j*} (\mathbf{W}'_1) \right\Vert^{\tilde{\alpha}}_2 \\
    \leq & \sqrt{2} c_W C_2 P \left[ (1-\alpha_2)^{1+\tilde{\alpha}} C^{\tilde{\alpha}}_2 \left\Vert g(\tilde{\mathbf{A}}) \right\Vert^{\tilde{\alpha}}_\infty + (1-\alpha_2) \right] \left( \frac{\beta_1 B_1}{C_1} \right)^{\tilde{\alpha}} \left\Vert {\rm vec}\left[\mathbf{W}_1 \right] - {\rm vec}\left[\mathbf{W}'_1 \right] \right\Vert^{\tilde{\alpha}}_2 \\
    & + 2(1-\alpha_2)^2 \beta_1 c^2_W \left\Vert g(\tilde{\mathbf{A}}) \right\Vert_\infty \frac{B_1 C^2_2}{C_1} \left\Vert {\rm vec}\left[\mathbf{W}_1 \right] - {\rm vec}\left[\mathbf{W}'_1 \right] \right\Vert_2.
\end{aligned}
\end{equation}
Finally,
\begin{equation}\label{delta20}
\begin{aligned}
    & \left\Vert \delta_{ij}(\mathbf{W}_0) - \delta_{ij}(\mathbf{W}'_0) \right\Vert_2 \\
    \leq & (1-\alpha_2) \left\Vert \sigma'(\mathbf{H}^{(1)}_{j*}) \odot ((\delta_i(\mathbf{W}_0) - \delta_i(\mathbf{W}'_0)) ((1-\beta_2)\mathbf{W}^\top_2+\beta_2 \mathbf{I})) \right\Vert_2 \\
    & + (1-\alpha_2) \left\Vert (\sigma'(\mathbf{H}^{(1)}_{j*})(\mathbf{W}_0) - \sigma'(\mathbf{H}^{(1)}_{j*})(\mathbf{W}'_0)) \odot (\delta_i ((1-\beta_2)\mathbf{W}^\top_2+\beta_2 \mathbf{I})) \right\Vert_2 \\ 
    \leq & (1-\alpha_2) C_2 \Vert \delta_i(\mathbf{W}_0) - \delta_i(\mathbf{W}'_0) \Vert_2 + \sqrt{2} (1-\alpha_2) c_W C_2 P \left\Vert \mathbf{H}^{(1)}_{j*} (\mathbf{W}_0) - \mathbf{H}^{(1)}_{j*} (\mathbf{W}'_0) \right\Vert^{\tilde{\alpha}}_2 \\
    \leq & \sqrt{2} (1-\alpha_2) C_2 P c_W \left[ \left( \frac{B_2}{c_W} \right)^{\tilde{\alpha}} + \left( \frac{B_1}{c_W} \right)^{\tilde{\alpha}}  \right] \left\Vert {\rm vec}\left[\mathbf{W}_0 \right] - {\rm vec}\left[\mathbf{W}'_0 \right] \right\Vert^{\tilde{\alpha}}_2 \\
    & + 2(1-\alpha_2) c_W B_2 C_2 \left\Vert {\rm vec}\left[\mathbf{W}_0 \right] - {\rm vec}\left[\mathbf{W}'_0 \right] \right\Vert_2.
\end{aligned}
\end{equation}
Now we are ready to discuss each gradient term in the following four parts.

\textbf{Part F.} First
\begin{equation*}
\begin{aligned}
    & \left\Vert \frac{\partial \ell(\mathbf{W}_0, \mathbf{W}_1, \mathbf{W}_2, \mathbf{W}_3;z_i)}{\partial {\rm vec} \left[ \mathbf{W}_3 \right]} (\mathbf{W}_3) - \frac{\partial \ell(\mathbf{W}_0, \mathbf{W}_1, \mathbf{W}_2, \mathbf{W}_3;z_i)}{\partial {\rm vec} \left[ \mathbf{W}_3 \right]} (\mathbf{W}'_3) \right\Vert_2 \\
    \leq & \mathop{\rm max}_{i\in [n]} \left\Vert \mathbf{H}^{(2)}_{i*} \right\Vert_2 \Vert \hat{\mathbf{y}}_i(\mathbf{W}_3) - \hat{\mathbf{y}}_i(\mathbf{W}'_3) \Vert_2 \leq 2 B^2_2 \left\Vert {\rm vec}\left[\mathbf{W}_2\right] - {\rm vec}\left[\mathbf{W}'_2\right] \right\Vert_2.
\end{aligned}
\end{equation*}
Similarly,
\begin{equation*}
\begin{aligned}
    & \left\Vert \frac{\partial \ell(\mathbf{W}_0, \mathbf{W}_1, \mathbf{W}_2, \mathbf{W}_3;z_i)}{\partial {\rm vec} \left[ \mathbf{W}_3 \right]} (\mathbf{W}_2) - \frac{\partial \ell(\mathbf{W}_0, \mathbf{W}_1, \mathbf{W}_2, \mathbf{W}_3;z_i)}{\partial {\rm vec} \left[ \mathbf{W}_3 \right]} (\mathbf{W}'_2) \right\Vert_2 \\
    \leq & \left\Vert \mathbf{H}^{(2)}_{i*}(\mathbf{W}_2) - \mathbf{H}^{(2)}_{i*}(\mathbf{W}'_2) \right\Vert_2 \Vert \hat{\mathbf{y}}_i - \mathbf{y}_i \Vert_2 + \mathop{\rm max}_{i\in [n]} \left\Vert \mathbf{H}^{(2)}_{i*} \right\Vert_2 \Vert \hat{\mathbf{y}}_i(\mathbf{W}_2) - \hat{\mathbf{y}}_i(\mathbf{W}'_2) \Vert_2 \\
    = & (\sqrt{2} + 2c_W B_2) \Delta_{22} = (\sqrt{2} + 2c_W B_2) \frac{\beta_2 B_2}{C_2} \left\Vert {\rm vec}\left[\mathbf{W}_2\right] - {\rm vec}\left[\mathbf{W}'_2\right] \right\Vert_2. \\
\end{aligned}
\end{equation*}
Similarly, we can obtain
\begin{equation*}
\begin{aligned}
    & \left\Vert \frac{\partial \ell(\mathbf{W}_0, \mathbf{W}_1, \mathbf{W}_2, \mathbf{W}_3;z_i)}{\partial {\rm vec} \left[ \mathbf{W}_3 \right]} (\mathbf{W}_1) - \frac{\partial \ell(\mathbf{W}_0, \mathbf{W}_1, \mathbf{W}_2, \mathbf{W}_3;z_i)}{\partial {\rm vec} \left[ \mathbf{W}_3 \right]} (\mathbf{W}'_1) \right\Vert_2 \\
    \leq & \left\Vert \mathbf{H}^{(2)}_{i*}(\mathbf{W}_1) - \mathbf{H}^{(2)}_{i*}(\mathbf{W}'_1) \right\Vert_2 \Vert \hat{\mathbf{y}}_i - \mathbf{y}_i \Vert_2 + \mathop{\rm max}_{i\in [n]} \left\Vert \mathbf{H}^{(2)}_{i*} \right\Vert_2 \Vert \hat{\mathbf{y}}_i(\mathbf{W}_1) - \hat{\mathbf{y}}_i(\mathbf{W}'_1) \Vert_2 \\
    = & (\sqrt{2} + 2c_W B_2) \Delta_{21} = (1-\alpha_2) \beta_1 (\sqrt{2} + 2c_W B_2) \frac{B_1 C_2}{C_1} \left\Vert g(\tilde{\mathbf{A}}) \right\Vert_\infty \left\Vert {\rm vec}\left[\mathbf{W}_1\right] - {\rm vec}\left[\mathbf{W}'_1\right] \right\Vert_2,
\end{aligned}
\end{equation*}
and
\begin{equation*}
\begin{aligned}
    & \left\Vert \frac{\partial \ell(\mathbf{W}_0, \mathbf{W}_1, \mathbf{W}_2, \mathbf{W}_3;z_i)}{\partial {\rm vec} \left[ \mathbf{W}_3 \right]} (\mathbf{W}_0) - \frac{\partial \ell(\mathbf{W}_0, \mathbf{W}_1, \mathbf{W}_2, \mathbf{W}_3;z_i)}{\partial {\rm vec} \left[ \mathbf{W}_3 \right]} (\mathbf{W}'_0) \right\Vert_2 \\
    \leq & \left\Vert \mathbf{H}^{(2)}_{i*}(\mathbf{W}_0) - \mathbf{H}^{(2)}_{i*}(\mathbf{W}'_0) \right\Vert_2 \Vert \hat{\mathbf{y}}_i - \mathbf{y}_i \Vert_2 + \mathop{\rm max}_{i\in [n]} \left\Vert \mathbf{H}^{(2)}_{i*} \right\Vert_2 \Vert \hat{\mathbf{y}}_i(\mathbf{W}_0) - \hat{\mathbf{y}}_i(\mathbf{W}'_0) \Vert_2 \\
    = & (\sqrt{2} + 2 c_W B_2) \Delta_{20} = (\sqrt{2} + 2 c_W B_2) \frac{B_2}{c_W} \left\Vert {\rm vec}\left[\mathbf{W}_0\right] - {\rm vec}\left[\mathbf{W}'_0\right] \right\Vert_2.
\end{aligned}
\end{equation*}
Denote by 
\begin{equation*}
\begin{aligned}
    P_{33} & = 2B^2_2, \ P_{32} = (\sqrt{2} + 2 c_W B_2) \frac{\beta_2 B_2}{C_2}, \ P_{31} = (1-\alpha_2) \beta_1 (\sqrt{2} + 2 c_W B_2) \frac{B_1 C_2}{C_1} \left\Vert g(\tilde{\mathbf{A}}) \right\Vert_\infty, \\
    \ P_{30} & = (\sqrt{2} + 2 c_W B_2) \frac{B_2}{c_W}, \ \tilde{P}_{33} = \tilde{P}_{32} = \tilde{P}_{31} = \tilde{P}_{30} = 0,
\end{aligned}
\end{equation*}
we obtain that $\left\Vert \frac{\partial \ell(\mathbf{w};z_i)}{\partial {\rm vec} \left[ \mathbf{W}_3 \right]} - \frac{\partial \ell(\mathbf{w}';z_i)}{\partial {\rm vec} \left[ \mathbf{W}_3 \right]} \right\Vert_2 \leq \sum_{i=1}^4 P_{3i} \left\Vert {\rm vec}\left[\mathbf{W}_i\right] - {\rm vec}\left[\mathbf{W}'_i\right] \right\Vert_2 + \sum_{i=1}^4 \tilde{P}_{3i} \left\Vert {\rm vec}\left[\mathbf{W}_i\right] - {\rm vec}\left[\mathbf{W}'_i\right] \right\Vert^{\tilde{\alpha}}_2 $.

\textbf{Part G.} 
First,
\begin{equation*}
\begin{aligned}
    & \left\Vert \frac{\partial \ell(\mathbf{W}_0, \mathbf{W}_1, \mathbf{W}_2, \mathbf{W}_3;z_i)}{\partial {\rm vec} \left[ \mathbf{W}_2 \right]} (\mathbf{W}_3) - \frac{\partial \ell(\mathbf{W}_0, \mathbf{W}_1, \mathbf{W}_2, \mathbf{W}_3;z_i)}{\partial {\rm vec} \left[ \mathbf{W}_2 \right]} (\mathbf{W}'_3) \right\Vert_2 \\
    = & \left\Vert \alpha_2 \beta_2 \mathbf{H}^{(0)}_{i*} + (1-\alpha_2)\beta_2 \sum_{j=1}^n \left[ g(\tilde{\mathbf{A}}) \right]_{ij} \mathbf{H}^{(1)}_{j*} \right\Vert_2 \left\Vert \delta_i(\mathbf{W}_3) - \delta_i(\mathbf{W}'_3) \right\Vert_2 \\
    \leq & (\sqrt{2} + 2 c_W B_2) \left( \alpha_2 \beta_2 \mathop{\rm max}_{i\in[n]} \left\Vert \mathbf{H}^{(0)}_{i*} \right\Vert_2 + (1-\alpha_2)\beta_2 \left\Vert g(\tilde{\mathbf{A}}) \right\Vert_\infty \mathop{\rm max}_{i\in[n]} \left\Vert \mathbf{H}^{(1)}_{i*} \right\Vert_2 \right) \left\Vert {\rm vec}\left[\mathbf{W}_3\right] - {\rm vec}\left[\mathbf{W}'_3\right] \right\Vert_2 \\
    \leq & (\sqrt{2} + 2 c_W B_2) \left( \alpha_2 \beta_2 c_X c_W + (1-\alpha_2) \beta_2 B_1 \left\Vert g(\tilde{\mathbf{A}}) \right\Vert_\infty \right) \left\Vert {\rm vec}\left[\mathbf{W}_3\right] - {\rm vec}\left[\mathbf{W}'_3\right] \right\Vert_2 \\
    = & (\sqrt{2} + 2 c_W B_2) \frac{\beta_2 B_2}{C_2} \left\Vert {\rm vec}\left[\mathbf{W}_3\right] - {\rm vec}\left[\mathbf{W}'_3\right] \right\Vert_2.
\end{aligned}
\end{equation*}
Similarly,
\begin{equation*}
\begin{aligned}
    & \left\Vert \frac{\partial \ell(\mathbf{W}_0, \mathbf{W}_1, \mathbf{W}_2, \mathbf{W}_3;z_i)}{\partial {\rm vec} \left[ \mathbf{W}_2 \right]} (\mathbf{W}_2) - \frac{\partial \ell(\mathbf{W}_0, \mathbf{W}_1, \mathbf{W}_2, \mathbf{W}_3;z_i)}{\partial {\rm vec} \left[ \mathbf{W}_2 \right]} (\mathbf{W}'_2) \right\Vert_2 \\
    = & \left\Vert \alpha_2 \beta_2 \mathbf{H}^{(0)}_{i*} + (1-\alpha_2)\beta_2 \sum_{j=1}^n \left[ g(\tilde{\mathbf{A}}) \right]_{ij} \mathbf{H}^{(1)}_{j*} \right\Vert_2 \left\Vert \delta_i(\mathbf{W}_2) - \delta_i(\mathbf{W}'_2) \right\Vert_2 \\
    \leq & \sqrt{2} P c_W \left( \frac{\beta_2 B_2}{C_2} \right)^{1+\tilde{\alpha}} \left\Vert {\rm vec}\left[\mathbf{W}_2\right] - {\rm vec}\left[\mathbf{W}'_2\right] \right\Vert^{\tilde{\alpha}}_2 + c^2_W \left( \frac{\beta_2 B_2}{C_2} \right)^2 \left\Vert {\rm vec}\left[\mathbf{W}_2\right] - {\rm vec}\left[\mathbf{W}'_2\right] \right\Vert_2.
\end{aligned}
\end{equation*}
Besides,
\begin{equation*}
\begin{aligned}
    & \left\Vert \frac{\partial \ell(\mathbf{W}_0, \mathbf{W}_1, \mathbf{W}_2, \mathbf{W}_3;z_i)}{\partial {\rm vec} \left[ \mathbf{W}_2 \right]} (\mathbf{W}_1) - \frac{\partial \ell(\mathbf{W}_0, \mathbf{W}_1, \mathbf{W}_2, \mathbf{W}_3;z_i)}{\partial {\rm vec} \left[ \mathbf{W}_2 \right]} (\mathbf{W}'_1) \right\Vert_2 \\
    \leq & \left\Vert \alpha_2 \beta_2 \mathbf{H}^{(0)}_{i*} + (1-\alpha_2)\beta_2 \sum_{j=1}^n \left[ g(\tilde{\mathbf{A}}) \right]_{ij} \mathbf{H}^{(1)}_{j*} \right\Vert_2 \left\Vert \delta_i(\mathbf{W}_1) - \delta_i(\mathbf{W}'_1) \right\Vert_2 \\
    & + \left\Vert (1-\alpha_2)\beta_2 \sum_{j=1}^n \left[ g(\tilde{\mathbf{A}}) \right]_{ij} (\mathbf{H}^{(1)}_{j*}(\mathbf{W}_1) - \mathbf{H}^{(1)}_{j*}(\mathbf{W}_1))  \right\Vert_2 \left\Vert \delta_i \right\Vert_2 \\
    \leq & \sqrt{2} P c_W (1-\alpha_2)^{\tilde{\alpha}} \beta^{\tilde{\alpha}}_1 \left( \frac{B_1 C_2}{C_1} \right)^{\tilde{\alpha}} \left\Vert g(\tilde{\mathbf{A}}) \right\Vert^{\tilde{\alpha}}_\infty \frac{\beta_2 B_2}{C_2} \left\Vert {\rm vec}\left[\mathbf{W}_1\right] - {\rm vec}\left[\mathbf{W}'_1\right] \right\Vert^{\tilde{\alpha}}_2 \\
    & + (1-\alpha_2) c_W \frac{B_1 \beta_1 \beta_2}{C_1} \left\Vert g(\tilde{\mathbf{A}}) \right\Vert_\infty \left[ c_W B_2 + \sqrt{2} \right] \left\Vert {\rm vec}\left[\mathbf{W}_1\right] - {\rm vec}\left[\mathbf{W}'_1\right] \right\Vert_2 .
\end{aligned}
\end{equation*}
Finally,
\begin{equation*}
\begin{aligned}
    & \left\Vert \frac{\partial \ell(\mathbf{W}_0, \mathbf{W}_1, \mathbf{W}_2, \mathbf{W}_3;z_i)}{\partial {\rm vec} \left[ \mathbf{W}_2 \right]} (\mathbf{W}_0) - \frac{\partial \ell(\mathbf{W}_0, \mathbf{W}_0, \mathbf{W}_2, \mathbf{W}_3;z_i)}{\partial {\rm vec} \left[ \mathbf{W}_2 \right]} (\mathbf{W}'_0) \right\Vert_2 \\
    \leq & \left\Vert \alpha_2 \beta_2 \mathbf{H}^{(0)}_{i*} + (1-\alpha_2)\beta_2 \sum_{j=1}^n \left[ g(\tilde{\mathbf{A}}) \right]_{ij} \mathbf{H}^{(1)}_{j*} \right\Vert_2 \left\Vert \delta_i(\mathbf{W}_0) - \delta_i(\mathbf{W}'_0) \right\Vert_2 \\
    & + \left\Vert \alpha_2 \beta_2 (\mathbf{H}^{(0)}_{i*}(\mathbf{W}_0) - \mathbf{H}^{(0)}_{i*}(\mathbf{W}_0))  \right\Vert_2 \mathop{\rm max}_{i\in [n]} \left\Vert \delta_i \right\Vert_2 + (1-\alpha_2)\beta_2 \left\Vert g(\tilde{\mathbf{A}}) \right\Vert_\infty \mathop{\rm max}_{i\in [n]} \Vert \delta_i \Vert_2 \left\Vert \mathbf{H}^{(1)}_{i*}(\mathbf{W}_0) - \mathbf{H}^{(1)}_{i*}(\mathbf{W}_0) \right\Vert_2 \\
    \leq & \sqrt{2} P c_W \left( \frac{B_2}{c_W} \right)^{\tilde{\alpha}} \frac{\beta_2 B_2}{C_2} \left\Vert {\rm vec}\left[\mathbf{W}_0\right] - {\rm vec}\left[\mathbf{W}'_0\right] \right\Vert^{\tilde{\alpha}}_2 + \frac{\beta_2 B_2 (\sqrt{2}+B_2 c_W)}{C_2} \left\Vert {\rm vec}\left[\mathbf{W}_0\right] - {\rm vec}\left[\mathbf{W}'_0\right] \right\Vert_2.
\end{aligned}
\end{equation*}
Denote by
\begin{equation*}
\begin{aligned}
    P_{23} & = (\sqrt{2} + 2 c_W B_2) \frac{\beta_2 B_2}{C_2}, \ P_{22} = c^2_W \left( \frac{\beta_2 B_2}{C_2} \right)^2, \\
    P_{21} & = (1-\alpha_2) c_W \frac{B_1 \beta_1 \beta_2}{C_1} \left\Vert g(\tilde{\mathbf{A}}) \right\Vert_\infty \left[ c_W B_2 + \sqrt{2} \right], \ P_{20} = \frac{\beta_2 B_2 (\sqrt{2}+B_2 c_W)}{C_2} \\
    \tilde{P}_{23} & = 0, \ \tilde{P}_{22} = \sqrt{2} P c_W \left( \frac{\beta_2 B_2}{C_2} \right)^{1+\tilde{\alpha}}, \ \tilde{P}_{21} = \sqrt{2} P c_W (1-\alpha_2)^{\tilde{\alpha}} \beta^{\tilde{\alpha}}_1 \left( \frac{B_1 C_2}{C_1} \right)^{\tilde{\alpha}} \left\Vert g(\tilde{\mathbf{A}}) \right\Vert^{\tilde{\alpha}}_\infty \frac{\beta_2 B_2}{C_2}, \\ \tilde{P}_{20} & = \sqrt{2} P c_W \left( \frac{B_2}{c_W} \right)^{\tilde{\alpha}} \frac{\beta_2 B_2}{C_2},
\end{aligned}
\end{equation*}
we obtain that $\left\Vert \frac{\partial \ell(\mathbf{w};z_i)}{\partial {\rm vec} \left[ \mathbf{W}_2 \right]} - \frac{\partial \ell(\mathbf{w}';z_i)}{\partial {\rm vec} \left[ \mathbf{W}_2 \right]} \right\Vert_2 \leq \sum_{i=1}^4 P_{2i} \left\Vert {\rm vec}\left[\mathbf{W}_i\right] - {\rm vec}\left[\mathbf{W}'_i\right] \right\Vert_2 + \sum_{i=1}^4 \tilde{P}_{2i} \left\Vert {\rm vec}\left[\mathbf{W}_i\right] - {\rm vec}\left[\mathbf{W}'_i\right] \right\Vert^{\tilde{\alpha}}_2$.

\textbf{Part H.} Since (1) $\mathbf{H}^{(0)}$ is variable related to $\mathbf{W}_0$; (2) $\delta_{ij}$ is variable related to $\mathbf{W}_0$, $\mathbf{W}_1$, $\mathbf{W}_2$, $\mathbf{W}_3$, we have
\begin{equation*}
\begin{aligned}
    & \left\Vert \frac{\partial \ell(\mathbf{W}_0, \mathbf{W}_1, \mathbf{W}_2, \mathbf{W}_3;z_i)}{\partial {\rm vec} \left[ \mathbf{W}_1 \right]} (\mathbf{W}_3) - \frac{\partial \ell(\mathbf{W}_0, \mathbf{W}_1, \mathbf{W}_2, \mathbf{W}_3;z_i)}{\partial {\rm vec} \left[ \mathbf{W}_1 \right]} (\mathbf{W}'_3) \right\Vert_2 \\
    = & \alpha_1 \beta_1 \sum_{j=1}^n  \left[ g(\tilde{\mathbf{A}}) \right]_{ij} \left\Vert \mathbf{H}^{(0)}_{j*} \right\Vert_2 \Vert \delta_{ij}(\mathbf{W}_3) - \delta_{ij}(\mathbf{W}'_3) \Vert_2 \\
    & + (1-\alpha_1)\beta_1 \sum_{j=1}^n \sum_{k=1}^n \left[ g(\tilde{\mathbf{A}}) \right]_{ij} \left[ g(\tilde{\mathbf{A}}) \right]_{jk} \left\Vert \mathbf{H}^{(0)}_{k*} \right\Vert_2 \Vert \delta_{ij}(\mathbf{W}_3) - \delta_{ij}(\mathbf{W}'_3) \Vert_2 \\
    \leq & c_X c_W (1-\alpha_2) C_2 \left( \alpha_1 \beta_1 \left\Vert g(\tilde{\mathbf{A}}) \right\Vert_\infty + (1-\alpha_1) \beta_1 \left\Vert g(\tilde{\mathbf{A}}) \right\Vert^2_\infty \right) (\sqrt{2}+2c_W B_2) \left\Vert {\rm vec}\left[\mathbf{W}_3\right] - {\rm vec}\left[\mathbf{W}'_3\right] \right\Vert_2 \\
    = & \frac{ \beta_1 B_1 C_2}{C_1} \left\Vert g(\tilde{\mathbf{A}}) \right\Vert_\infty (\sqrt{2}+2c_W B_2) \left\Vert {\rm vec}\left[\mathbf{W}_3\right] - {\rm vec}\left[\mathbf{W}'_3\right] \right\Vert_2,
\end{aligned}
\end{equation*}
where we have used
\begin{equation*}
   \alpha_1 \beta_1 c_X c_W \left\Vert g(\tilde{\mathbf{A}}) \right\Vert_\infty + (1-\alpha_1) \beta_1 c_X c_W \left\Vert g(\tilde{\mathbf{A}}) \right\Vert^2_\infty = \beta_1 \frac{B_1}{C_1} \left\Vert g(\tilde{\mathbf{A}}) \right\Vert_\infty.
\end{equation*}
Similarly, we have
\begin{equation*}
\begin{aligned}
    & \left\Vert \frac{\partial \ell(\mathbf{W}_0, \mathbf{W}_1, \mathbf{W}_2, \mathbf{W}_3;z_i)}{\partial {\rm vec} \left[ \mathbf{W}_1 \right]} (\mathbf{W}_2) - \frac{\partial \ell(\mathbf{W}_0, \mathbf{W}_1, \mathbf{W}_2, \mathbf{W}_3;z_i)}{\partial {\rm vec} \left[ \mathbf{W}_1 \right]} (\mathbf{W}'_2) \right\Vert_2 \\
    = & \alpha_1 \beta_1 \sum_{j=1}^n  \left[ g(\tilde{\mathbf{A}}) \right]_{ij} \left\Vert \mathbf{H}^{(0)}_{j*} \right\Vert_2 \Vert \delta_{ij}(\mathbf{W}_2) - \delta_{ij}(\mathbf{W}'_2) \Vert_2 \\
    & + (1-\alpha_1)\beta_1 \sum_{j=1}^n \sum_{k=1}^n \left[ g(\tilde{\mathbf{A}}) \right]_{ij} \left[ g(\tilde{\mathbf{A}}) \right]_{jk} \left\Vert \mathbf{H}^{(0)}_{k*} \right\Vert_2 \Vert \delta_{ij}(\mathbf{W}_2) - \delta_{ij}(\mathbf{W}'_2) \Vert_2 \\
    \leq & \sqrt{2} (1-\alpha_2) c_W P  \frac{\beta_1 B_1 C_2}{C_1} \left\Vert g(\tilde{\mathbf{A}}) \right\Vert_\infty \left( \frac{\beta_2 B_2}{C_2}\right)^{\tilde{\alpha}} \left\Vert {\rm vec}\left[\mathbf{W}_2\right] - {\rm vec}\left[\mathbf{W}'_2\right] \right\Vert^{\tilde{\alpha}}_2 \\
    & + (1-\alpha_2) \frac{\beta_1 B_1}{C_1} \left\Vert g(\tilde{\mathbf{A}}) \right\Vert_\infty \left( c^2_W \beta_2 B_2 + \sqrt{2} (1-\beta_2) c_W \right) \left\Vert {\rm vec}\left[\mathbf{W}_2\right] - {\rm vec}\left[\mathbf{W}'_2\right] \right\Vert_2.
\end{aligned}
\end{equation*}
Besides,
\begin{equation*}
\begin{aligned}
    & \left\Vert \frac{\partial \ell(\mathbf{W}_0, \mathbf{W}_1, \mathbf{W}_2, \mathbf{W}_3;z_i)}{\partial {\rm vec} \left[ \mathbf{W}_1 \right]} (\mathbf{W}_1) - \frac{\partial \ell(\mathbf{W}_0, \mathbf{W}_1, \mathbf{W}_2, \mathbf{W}_3;z_i)}{\partial {\rm vec} \left[ \mathbf{W}_1 \right]} (\mathbf{W}'_1) \right\Vert_2 \\
    = & \alpha_1 \beta_1 \sum_{j=1}^n  \left[ g(\tilde{\mathbf{A}}) \right]_{ij} \left\Vert \mathbf{H}^{(0)}_{j*} \right\Vert_2 \Vert \delta_{ij}(\mathbf{W}_1) - \delta_{ij}(\mathbf{W}'_1) \Vert_2 \\
    & + (1-\alpha_1)\beta_1 \sum_{j=1}^n \sum_{k=1}^n \left[ g(\tilde{\mathbf{A}}) \right]_{ij} \left[ g(\tilde{\mathbf{A}}) \right]_{jk} \left\Vert \mathbf{H}^{(0)}_{k*} \right\Vert_2 \Vert \delta_{ij}(\mathbf{W}_1) - \delta_{ij}(\mathbf{W}'_1) \Vert_2 \\
    \leq & \sqrt{2} c_W C_2 P \left[ (1-\alpha_2)^{1+\tilde{\alpha}} C^{\tilde{\alpha}}_2 \left\Vert g(\tilde{\mathbf{A}}) \right\Vert^{\tilde{\alpha}}_\infty + (1-\alpha_2) \right] \left( \frac{\beta_1 B_1}{C_1} \right)^{\tilde{\alpha}+1} \left\Vert g(\tilde{\mathbf{A}}) \right\Vert_\infty \left\Vert {\rm vec}\left[\mathbf{W}_1 \right] - {\rm vec}\left[\mathbf{W}'_1 \right] \right\Vert^{\tilde{\alpha}}_2 \\
    & + (1-\alpha_2)^2 \beta^2_1 c^2_W \frac{B^2_1 C^2_2}{C^2_1} \left\Vert g(\tilde{\mathbf{A}}) \right\Vert_\infty \left\Vert {\rm vec}\left[\mathbf{W}_1 \right] - {\rm vec}\left[\mathbf{W}'_1 \right] \right\Vert_2 .
\end{aligned}
\end{equation*}
Finally,
\begin{equation*}
\begin{aligned}
    & \left\Vert \frac{\partial \ell(\mathbf{W}_0, \mathbf{W}_1, \mathbf{W}_2, \mathbf{W}_3;z_i)}{\partial {\rm vec} \left[ \mathbf{W}_1 \right]} (\mathbf{W}_0) - \frac{\partial \ell(\mathbf{W}_0, \mathbf{W}_1, \mathbf{W}_2, \mathbf{W}_3;z_i)}{\partial {\rm vec} \left[ \mathbf{W}_1 \right]} (\mathbf{W}'_0) \right\Vert_2 \\
    = & \alpha_1 \beta_1 \sum_{j=1}^n  \left[ g(\tilde{\mathbf{A}}) \right]_{ij} \left\Vert \mathbf{H}^{(0)}_{j*} \right\Vert_2 \Vert \delta_{ij}(\mathbf{W}_0) - \delta_{ij}(\mathbf{W}'_0) \Vert_2 \\
    & + (1-\alpha_1)\beta_1 \sum_{j=1}^n \sum_{k=1}^n \left[ g(\tilde{\mathbf{A}}) \right]_{ij} \left[ g(\tilde{\mathbf{A}}) \right]_{jk} \left\Vert \mathbf{H}^{(0)}_{k*} \right\Vert_2 \Vert \delta_{ij}(\mathbf{W}_0) - \delta_{ij}(\mathbf{W}'_0) \Vert_2 \\
    \leq & \sqrt{2} (1 - \alpha_2) c_W P \left[ \left( \frac{B_2}{c_W} \right)^{\tilde{\alpha}} + \left( \frac{B_1}{c_W} \right)^{\tilde{\alpha}} \right] \frac{\beta_1 B_1 C_2}{C_1} \left\Vert g(\tilde{\mathbf{A}}) \right\Vert_\infty \left\Vert {\rm vec}\left[\mathbf{W}_0 \right] - {\rm vec}\left[\mathbf{W}'_0 \right] \right\Vert^{\tilde{\alpha}}_2 \\
    & + (1-\alpha_2) c_W \frac{\beta_1 B_1 B_2}{C_1} \left\Vert g(\tilde{\mathbf{A}}) \right\Vert_\infty \left\Vert {\rm vec}\left[\mathbf{W}_0 \right] - {\rm vec}\left[\mathbf{W}'_0 \right] \right\Vert_2.
\end{aligned}
\end{equation*}
Denote by
\begin{equation*}
\begin{aligned}
    P_{13} & = \frac{ \beta_1 B_1 C_2}{C_1} \left\Vert g(\tilde{\mathbf{A}}) \right\Vert_\infty (\sqrt{2}+2c_W B_2), \ P_{12} = (1-\alpha_2) \frac{\beta_1 B_1}{C_1} \left\Vert g(\tilde{\mathbf{A}}) \right\Vert_\infty \left( c^2_W \beta_2 B_2 + \sqrt{2} (1-\beta_2) c_W \right)\\
    P_{11} & = (1-\alpha_2)^2 \beta^2_1 c^2_W \frac{B^2_1 C^2_2}{C^2_1} \left\Vert g(\tilde{\mathbf{A}}) \right\Vert_\infty,  \ P_{10} = (1-\alpha_2) c_W \frac{\beta_1 B_1 B_2}{C_1} \left\Vert g(\tilde{\mathbf{A}}) \right\Vert_\infty \\
    \tilde{P}_{13} & = 0, \ \tilde{P}_{12} = \sqrt{2} (1-\alpha_2) c_W P  \frac{\beta_1 B_1 C_2}{C_1} \left\Vert g(\tilde{\mathbf{A}}) \right\Vert_\infty \left( \frac{\beta_2 B_2}{C_2}\right)^{\tilde{\alpha}} \\
    \tilde{P}_{11} & = \sqrt{2} c_W C_2 P \left[ (1-\alpha_2)^{1+\tilde{\alpha}} C^{\tilde{\alpha}}_2 \left\Vert g(\tilde{\mathbf{A}}) \right\Vert^{\tilde{\alpha}}_\infty + (1-\alpha_2) \right] \left( \frac{\beta_1 B_1}{C_1} \right)^{\tilde{\alpha}+1} \left\Vert g(\tilde{\mathbf{A}}) \right\Vert_\infty, \\
    P_{10} & = \sqrt{2} (1 - \alpha_2) c_W P \left[ \left( \frac{B_2}{c_W} \right)^{\tilde{\alpha}} + \left( \frac{B_1}{c_W} \right)^{\tilde{\alpha}} \right] \frac{\beta_1 B_1 C_2}{C_1} \left\Vert g(\tilde{\mathbf{A}}) \right\Vert_\infty ,
\end{aligned}
\end{equation*}
we obtain that $\left\Vert \frac{\partial \ell(\mathbf{w};z_i)}{\partial {\rm vec} \left[ \mathbf{W}_1 \right]} - \frac{\partial \ell(\mathbf{w}';z_i)}{\partial {\rm vec} \left[ \mathbf{W}_1 \right]} \right\Vert_2 \leq \sum_{i=1}^4 P_{1i} \left\Vert {\rm vec}\left[\mathbf{W}_i\right] - {\rm vec}\left[\mathbf{W}'_i\right] \right\Vert_2 + \sum_{i=1}^4 \tilde{P}_{1i} \left\Vert {\rm vec}\left[\mathbf{W}_i\right] - {\rm vec}\left[\mathbf{W}'_i\right] \right\Vert^{\tilde{\alpha}}_2 $.

\textbf{Part I.}
First we have
\begin{equation*}
\begin{aligned}
    & \left\Vert \frac{\partial \ell(\mathbf{W}_0, \mathbf{W}_1, \mathbf{W}_2, \mathbf{W}_3;z_i)}{\partial {\rm vec} \left[ \mathbf{W}_0 \right]} (\mathbf{W}_3) - \frac{\partial \ell(\mathbf{W}_0, \mathbf{W}_1, \mathbf{W}_2, \mathbf{W}_3;z_i)}{\partial {\rm vec} \left[ \mathbf{W}_0 \right]} (\mathbf{W}'_3) \right\Vert_2 \\
    \leq & \alpha_2 \left\Vert \mathbf{X}_{i*} \right\Vert_2 \left\Vert \delta_i(\mathbf{W}_3) - \delta_i(\mathbf{W}'_3) \right\Vert_2 \left\Vert (1-\beta_2)\mathbf{I}+\beta_2 \mathbf{W}^{\top}_2 \right\Vert_2 \left\Vert \mathbf{H}^{(0)}_{i*} \right\Vert_2 \\
    & + \alpha_1 \sum_{j=1}^n  \left[g(\tilde{\mathbf{A}})\right]_{ij} \left\Vert \mathbf{X}_{j*} \right\Vert_2 \left\Vert \delta_{ij}(\mathbf{W}_3) - \delta_{ij}(\mathbf{W}'_3) \right\Vert_2 \left\Vert (1-\beta_1)\mathbf{I}+\beta_1\mathbf{W}_1^{\top}) \right\Vert_2 \left\Vert \mathbf{H}^{(0)}_{j*} \right\Vert_2 \\
    & + (1-\alpha_1) \sum_{j=1}^n  \sum_{k=1}^n \left[g(\tilde{\mathbf{A}})\right]_{ij} \left[g(\tilde{\mathbf{A}})\right]_{jk} \left\Vert \mathbf{X}_{k*} \right\Vert_2 \left\Vert \delta_{ij}(\mathbf{W}_3) - \delta_{ij}(\mathbf{W}'_3) \right\Vert_2 \left\Vert (1-\beta_1)\mathbf{I}+\beta_1\mathbf{W}_1^{\top}) \right\Vert_2 \left\Vert \mathbf{H}^{(0)}_{k*} \right\Vert_2 \\
    \leq & \left( \alpha_2 C_2 + \alpha_1 (1-\alpha_2) C_1 C_2 \left\Vert g(\tilde{\mathbf{A}}) \right\Vert_\infty + (1 - \alpha_1 )(1-\alpha_2) C_1 C_2 \left\Vert g(\tilde{\mathbf{A}}) \right\Vert^2_\infty \right) c^2_X c_W \\ 
    & \times (\sqrt{2} + 2c_W B_2) \left\Vert {\rm vec}\left[\mathbf{W}_3\right] - {\rm vec}\left[\mathbf{W}'_3\right] \right\Vert_2 \\
    = & c_X B_2 (\sqrt{2} + 2c_W B_2) \left\Vert {\rm vec}\left[\mathbf{W}_3\right] - {\rm vec}\left[\mathbf{W}'_3\right] \right\Vert_2.
\end{aligned}
\end{equation*}
Similarly,
\begin{equation*}
\begin{aligned}
    & \left\Vert \frac{\partial \ell(\mathbf{W}_0, \mathbf{W}_1, \mathbf{W}_2, \mathbf{W}_3;z_i)}{\partial {\rm vec} \left[ \mathbf{W}_0 \right]} (\mathbf{W}_2) - \frac{\partial \ell(\mathbf{W}_0, \mathbf{W}_1, \mathbf{W}_2, \mathbf{W}_3;z_i)}{\partial {\rm vec} \left[ \mathbf{W}_0 \right]} (\mathbf{W}'_2) \right\Vert_2 \\
    \leq & \alpha_2 \left\Vert \mathbf{X}_{i*} \right\Vert_2 \left\Vert \delta_i(\mathbf{W}_2) - \delta_i(\mathbf{W}'_2) \right\Vert_2 \left\Vert (1-\beta_2)\mathbf{I}+\beta_2 \mathbf{W}^{\top}_2 \right\Vert_2 \left\Vert \mathbf{H}^{(0)}_{i*} \right\Vert_2 \\
    & + \alpha_2 \beta_2 \left\Vert \mathbf{X}_{i*} \right\Vert_2 \Vert \delta_i \Vert_2 \left\Vert \mathbf{H}^{(0)}_{i*} \right\Vert_2 \left\Vert \mathbf{W}_2 - \mathbf{W}'_2 \right\Vert \\
    & + \alpha_1 \sum_{j=1}^n  \left[g(\tilde{\mathbf{A}})\right]_{ij} \left\Vert \mathbf{X}_{j*} \right\Vert_2 \left\Vert \delta_{ij}(\mathbf{W}_2) - \delta_{ij}(\mathbf{W}'_2) \right\Vert_2 \left\Vert (1-\beta_1)\mathbf{I}+\beta_1\mathbf{W}_1^{\top}) \right\Vert_2 \left\Vert \mathbf{H}^{(0)}_{j*} \right\Vert_2 \\
    & + (1-\alpha_1) \sum_{j=1}^n  \sum_{k=1}^n \left[g(\tilde{\mathbf{A}})\right]_{ij} \left[g(\tilde{\mathbf{A}})\right]_{jk} \left\Vert \mathbf{X}_{k*} \right\Vert_2 \left\Vert \delta_{ij}(\mathbf{W}_2) - \delta_{ij}(\mathbf{W}'_2) \right\Vert_2 \left\Vert (1-\beta_1)\mathbf{I}+\beta_1\mathbf{W}_1^{\top}) \right\Vert_2 \left\Vert \mathbf{H}^{(0)}_{k*} \right\Vert_2 \\
    \leq & B_2 c_X c_W \sqrt{2} P \left( \frac{\beta_2 B_2}{C_2} \right)^{\tilde{\alpha}} \left\Vert {\rm vec}\left[\mathbf{W}_2\right] - {\rm vec}\left[\mathbf{W}'_2\right] \right\Vert^{\tilde{\alpha}}_2 \\
    & + \left[ 2 c_X c^2_W \beta_2 \frac{B^2_2}{C_2} + \sqrt{2} \alpha_2 \beta_2 c^2_X c^2_W + \sqrt{2}(1-\alpha_2)(1-\beta_2) c_X c_W B_1 \left\Vert g(\tilde{\mathbf{A}}) \right\Vert_\infty \right] \left\Vert {\rm vec}\left[\mathbf{W}_2\right] - {\rm vec}\left[\mathbf{W}'_2\right] \right\Vert_2.
\end{aligned}
\end{equation*}
Besides,
\begin{equation*}
\begin{aligned}
    & \left\Vert \frac{\partial \ell(\mathbf{W}_0, \mathbf{W}_1, \mathbf{W}_2, \mathbf{W}_3;z_i)}{\partial {\rm vec} \left[ \mathbf{W}_0 \right]} (\mathbf{W}_1) - \frac{\partial \ell(\mathbf{W}_0, \mathbf{W}_1, \mathbf{W}_2, \mathbf{W}_3;z_i)}{\partial {\rm vec} \left[ \mathbf{W}_0 \right]} (\mathbf{W}'_1) \right\Vert_2 \\
    \leq & \alpha_2 \left\Vert \mathbf{X}_{i*} \right\Vert_2 \left\Vert \delta_i(\mathbf{W}_1) - \delta_i(\mathbf{W}'_1) \right\Vert_2 \left\Vert (1-\beta_2)\mathbf{I}+\beta_2 \mathbf{W}^{\top}_2 \right\Vert_2 \left\Vert \mathbf{H}^{(0)}_{i*} \right\Vert_2 \\
    & + \alpha_1 \beta_1  \sum_{j=1}^n \left[ g(\tilde{\mathbf{A}}) \right]_{ij} \left\Vert \mathbf{X}_{j*} \right\Vert_2 \Vert \delta_{ij} \Vert_2 \left\Vert \mathbf{H}^{(0)}_{j*} \right\Vert_2 \left\Vert \mathbf{W}_1 - \mathbf{W}'_1 \right\Vert \\
    & + \alpha_1 \sum_{j=1}^n  \left[g(\tilde{\mathbf{A}})\right]_{ij} \left\Vert \mathbf{X}_{j*} \right\Vert_2 \left\Vert \delta_{ij}(\mathbf{W}_1) - \delta_{ij}(\mathbf{W}'_1) \right\Vert_2 \left\Vert (1-\beta_1)\mathbf{I}+\beta_1\mathbf{W}_1^{\top}) \right\Vert_2 \left\Vert \mathbf{H}^{(0)}_{j*} \right\Vert_2 \\
    & + (1-\alpha_1) \sum_{j=1}^n  \sum_{k=1}^n \left[g(\tilde{\mathbf{A}})\right]_{ij} \left[g(\tilde{\mathbf{A}})\right]_{jk} \left\Vert \mathbf{X}_{k*} \right\Vert_2 \left\Vert \delta_{ij}(\mathbf{W}_1) - \delta_{ij}(\mathbf{W}'_1) \right\Vert_2 \left\Vert (1-\beta_1)\mathbf{I}+\beta_1\mathbf{W}_1^{\top}) \right\Vert_2 \left\Vert \mathbf{H}^{(0)}_{k*} \right\Vert_2 \\
    \leq & \left( B_2 (1-\alpha_2)^{\tilde{\alpha}} C^{\tilde{\alpha}}_2 \left\Vert g(\tilde{\mathbf{A}}) \right\Vert^{\tilde{\alpha}}_\infty + B_1 C_2 \left\Vert g(\tilde{\mathbf{A}}) \right\Vert_\infty \right) \sqrt{2} P c_X c_W \left( \frac{\beta_1 B_1}{C_1} \right)^{\tilde{\alpha}} \left\Vert {\rm vec}\left[\mathbf{W}_1\right] - {\rm vec}\left[\mathbf{W}'_1\right] \right\Vert^{\tilde{\alpha}}_2 \\
    & + \left( 2(1-\alpha_2) \beta_1 c_X c^2_W \frac{B_1 B_2 C_2}{C_1}  + \sqrt{2} \alpha_1 (1-\alpha_2) \beta_1 c^2_X c^2_W C_2 \right) \left\Vert g(\tilde{\mathbf{A}}) \right\Vert_\infty \left\Vert {\rm vec}\left[\mathbf{W}_1\right] - {\rm vec}\left[\mathbf{W}'_1\right] \right\Vert_2.
\end{aligned}
\end{equation*}
Finally,
\begin{equation*}
\begin{aligned}
    & \left\Vert \frac{\partial \ell(\mathbf{W}_0, \mathbf{W}_1, \mathbf{W}_2, \mathbf{W}_3;z_i)}{\partial {\rm vec} \left[ \mathbf{W}_0 \right]} (\mathbf{W}_0) - \frac{\partial \ell(\mathbf{W}_0, \mathbf{W}_1, \mathbf{W}_2, \mathbf{W}_3;z_i)}{\partial {\rm vec} \left[ \mathbf{W}_0 \right]} (\mathbf{W}'_0) \right\Vert_2 \\
    \leq & \alpha_2 \left\Vert \mathbf{X}_{i*} \right\Vert_2 \left\Vert \delta_i(\mathbf{W}_0) - \delta_i(\mathbf{W}'_0) \right\Vert_2 \left\Vert (1-\beta_2)\mathbf{I}+\beta_2 \mathbf{W}^{\top}_2 \right\Vert_2 \left\Vert \mathbf{H}^{(0)}_{i*} \right\Vert_2 \\
    & + \alpha_2 \left\Vert \mathbf{X}_{i*} \right\Vert_2 \left\Vert \delta_i ((1-\beta_2)\mathbf{I}+\beta_2 \mathbf{W}^{\top}_2) \right\Vert_2 \left\Vert \mathbf{H}^{(0)}_{i*}(\mathbf{W}_0) - \mathbf{H}^{(0)}_{i*}(\mathbf{W}'_0) \right\Vert_2 \\
    & + \alpha_1 \sum_{j=1}^n  \left[g(\tilde{\mathbf{A}})\right]_{ij} \left\Vert \mathbf{X}_{j*} \right\Vert_2 \left\Vert \delta_{ij}(\mathbf{W}_1) - \delta_{ij}(\mathbf{W}'_1) \right\Vert_2 \left\Vert (1-\beta_1)\mathbf{I}+\beta_1\mathbf{W}_1^{\top}) \right\Vert_2 \left\Vert \mathbf{H}^{(0)}_{j*} \right\Vert_2 \\
    & + \alpha_1 \sum_{j=1}^n  \left[g(\tilde{\mathbf{A}})\right]_{ij} \left\Vert \mathbf{X}_{j*} \right\Vert_2 \left\Vert \delta_{ij} ((1-\beta_1)\mathbf{I}+\beta_1 \mathbf{W}^{\top}_1) \right\Vert_2 \left\Vert \mathbf{H}^{(0)}_{j*}(\mathbf{W}_0) - \mathbf{H}^{(0)}_{j*}(\mathbf{W}'_0) \right\Vert_2 \\
    & + (1-\alpha_1) \sum_{j=1}^n  \sum_{k=1}^n \left[g(\tilde{\mathbf{A}})\right]_{ij} \left[g(\tilde{\mathbf{A}})\right]_{jk} \left\Vert \mathbf{X}_{k*} \right\Vert_2 \left\Vert \delta_{ij}(\mathbf{W}_0) - \delta_{ij}(\mathbf{W}'_0) \right\Vert_2 \left\Vert (1-\beta_1)\mathbf{I}+\beta_1\mathbf{W}_1^{\top}) \right\Vert_2 \left\Vert \mathbf{H}^{(0)}_{k*} \right\Vert_2 \\
    & + (1-\alpha_1) \sum_{j=1}^n  \sum_{k=1}^n \left[g(\tilde{\mathbf{A}})\right]_{ij} \left[g(\tilde{\mathbf{A}})\right]_{jk} \left\Vert \mathbf{X}_{k*} \right\Vert_2 \left\Vert \delta_{ij} ((1-\beta_1)\mathbf{I}+\beta_1 \mathbf{W}^{\top}_1) \right\Vert_2 \left\Vert \mathbf{H}^{(0)}_{k*}(\mathbf{W}_0) - \mathbf{H}^{(0)}_{k*}(\mathbf{W}'_0) \right\Vert_2 \\
    \leq & c_X B_2 (\sqrt{2}  + 2 c_W B_2) \left\Vert {\rm vec}\left[\mathbf{W}_0\right] - {\rm vec}\left[\mathbf{W}'_0\right] \right\Vert \\
    & + \sqrt{2} c_X c_W P \left[ B_2 \left( \frac{B_2}{c_W} \right)^{\tilde{\alpha}} + (1-\alpha_2) C_2 \left\Vert g(\tilde{\mathbf{A}}) \right\Vert_\infty B_1 \left( \frac{B_1}{c_W} \right)^{\tilde{\alpha}} \right] \left\Vert {\rm vec}\left[\mathbf{W}_0\right] - {\rm vec}\left[\mathbf{W}'_0\right] \right\Vert^{\tilde{\alpha}}_2.
\end{aligned}
\end{equation*}
Denote by 
\begin{equation*}
\begin{aligned}
    P_{03} & = c_X B_2 (\sqrt{2} + 2c_W B_2), \ P_{02} = \sqrt{2} \alpha_2 \beta_2 c^2_X c^2_W + \sqrt{2}(1-\alpha_2)(1-\beta_2) c_X c_W B_1 \left\Vert g(\tilde{\mathbf{A}}) \right\Vert_\infty , \\
    P_{01} & = \left( 2(1-\alpha_2) \beta_1 c_X c^2_W \frac{B_1 B_2 C_2}{C_1}  + \sqrt{2} \alpha_1 (1-\alpha_2) \beta_1 c^2_X c^2_W C_2 \right) \left\Vert g(\tilde{\mathbf{A}}) \right\Vert_\infty \ P_{00} = c_X B_2 (\sqrt{2} + 2 c_W B_2), \\
    \tilde{P}_{03} & = 0, \ \tilde{P}_{02} = B_2 c_X c_W \sqrt{2} P \left( \frac{\beta_2 B_2}{C_2} \right)^{\tilde{\alpha}}, \\
    \tilde{P}_{01} & = \left( B_2 (1-\alpha_2)^{\tilde{\alpha}} C^{\tilde{\alpha}}_2 \left\Vert g(\tilde{\mathbf{A}}) \right\Vert^{\tilde{\alpha}}_\infty + B_1 C_2 \left\Vert g(\tilde{\mathbf{A}}) \right\Vert_\infty \right) \sqrt{2} P c_X c_W \left( \frac{\beta_1 B_1}{C_1} \right)^{\tilde{\alpha}}, \\
    \tilde{P}_{00} & = \sqrt{2} c_X c_W P \left[ B_2 \left( \frac{B_2}{c_W} \right)^{\tilde{\alpha}} + (1-\alpha_2) C_2 \left\Vert g(\tilde{\mathbf{A}}) \right\Vert_\infty B_1 \left( \frac{B_1}{c_W} \right)^{\tilde{\alpha}} \right],
\end{aligned}
\end{equation*}
we obtain that $\left\Vert \frac{\partial \ell(\mathbf{w};z_i)}{\partial {\rm vec} \left[ \mathbf{W}_0 \right]} - \frac{\partial \ell(\mathbf{w}';z_i)}{\partial {\rm vec} \left[ \mathbf{W}_0 \right]} \right\Vert_2 \leq \sum_{i=1}^4 P_{0i} \left\Vert {\rm vec}\left[\mathbf{W}_i\right] - {\rm vec}\left[\mathbf{W}'_i\right] \right\Vert_2 + \sum_{i=1}^4 \tilde{P}_{0i} \left\Vert {\rm vec}\left[\mathbf{W}_i\right] - {\rm vec}\left[\mathbf{W}'_i\right] \right\Vert^{\tilde{\alpha}}_2$. Combing the results in Part~F, Part~G, Part~H, Part~I, we conclude that $\Vert \nabla \ell(\mathbf{w}) - \nabla \ell(\mathbf{w}') \Vert_2 \leq P_{\mathcal{F}} \mathop{\rm max} \{ \Vert \mathbf{w} - \mathbf{w}' \Vert_2, \Vert \mathbf{w} - \mathbf{w}' \Vert^{\tilde{\alpha}}_2 \}$ holds where $\mathbf{w}= \left[{\rm vec}\left[ \mathbf{W}_0 \right]; {\rm vec}\left[ \mathbf{W}_1 \right]; {\rm vec}\left[ \mathbf{W}_2 \right]; {\rm vec}\left[ \mathbf{W}_3 \right]\right]$ by Lemma~\ref{bound_alpha}.

\subsubsection{Proof of Proposition~\ref{sgc}}
For two layer SGC, we have $g(\tilde{\mathbf{A}}) = \tilde{\mathbf{A}}^2$. Note that
\begin{equation*}
\begin{aligned}
    |\ell(\mathbf{W}_1 \mathbf{W}_2; z_i) - \ell(\mathbf{W}_1 \mathbf{W}'_2; z_i)| \leq & \sqrt{2} \left\Vert \sum_{j=1}^n {\left[g(\tilde{{\mathbf{A}}})\right]_{ij} \mathbf{X}_{j*} \mathbf{W}_1 (\mathbf{W}_2 - \mathbf{W}'_2) }\right\Vert_2\\
    \leq & \sqrt{2} \sum_{j=1}^n \left[g(\tilde{{\mathbf{A}}})\right]_{ij} \left\Vert \mathbf{X}_{j*} \mathbf{W}_1 \right\Vert_2 \left\Vert \mathbf{W}_2 - \mathbf{W}'_2 \right\Vert \\
    \leq & c_X c_W \sqrt{2} \left\Vert g(\tilde{{\mathbf{A}}}) \right\Vert_\infty \left\Vert \mathbf{W}_2 - \mathbf{W}'_2 \right\Vert .
\end{aligned}
\end{equation*}
Similarly,
\begin{equation*}
\begin{aligned}
    |\ell(\mathbf{W}_1 \mathbf{W}_2; z_i) - \ell(\mathbf{W}'_1 \mathbf{W}_2; z_i)| \leq c_X c_W \sqrt{2} \left\Vert g(\tilde{{\mathbf{A}}}) \right\Vert_\infty \left\Vert \mathbf{W}_1 - \mathbf{W}'_1 \right\Vert.
\end{aligned}
\end{equation*}
Denote by $L_1 = L_2 = c_X c_W \sqrt{2} \left\Vert g(\tilde{\mathbf{A}}) \right\Vert_{\infty}$, we conclude that$\vert \ell(\mathbf{w}) - \ell(\mathbf{w}') \vert \leq L_{\mathcal{F}} \Vert \mathbf{w} - \mathbf{w}' \Vert_2$ holds with $L_\mathcal{F} = 2c_X c_W \left\Vert g(\tilde{\mathbf{A}}) \right\Vert_{\infty}$ by Lemma~\ref{bound_alpha}. Then we discuss the H\"{o}lder smoothness. By the chain rule, the gradients are
\begin{equation*}
\begin{aligned}
    \frac{\partial \ell(\mathbf{W}_1, \mathbf{W}_2;z_i)}{\partial {\rm vec} \left[\mathbf{W}_2\right]} & = \sum_{j=1}^n \left[g(\tilde{\mathbf{A}})\right]_{ij} (\hat{\mathbf{y}}_i - \mathbf{y}_i) \otimes (\mathbf{X}_{j*}\mathbf{W}_1), \\
    \frac{\partial \ell(\mathbf{W}_1, \mathbf{W}_2;z_i)}{\partial {\rm vec} \left[\mathbf{W}_1\right]} & = \sum_{j=1}^n \left[g(\tilde{\mathbf{A}})\right]_{ij} ((\hat{\mathbf{y}}_i - \mathbf{y}_i) \mathbf{W}^\top_2) \otimes \mathbf{X}_{j*}.
\end{aligned}
\end{equation*}
First,
\begin{equation*}
\begin{aligned}
    & \left\Vert \frac{\partial \ell(\mathbf{W}_1, \mathbf{W}_2;z_i)}{\partial {\rm vec} \left[ \mathbf{W}_2 \right]} (\mathbf{W}_1) -  \frac{\partial \ell(\mathbf{W}_1, \mathbf{W}_2;z_i)}{\partial {\rm vec} \left[ \mathbf{W}_2 \right]} (\mathbf{W}'_1) \right\Vert_2 \\
    \leq & \sum_{j=1}^n \left[g(\tilde{\mathbf{A}})\right]_{ij} \left\Vert  (\hat{\mathbf{y}}_i - \mathbf{y}_i) \otimes (\mathbf{X}_{j*} (\mathbf{W}_1 - \mathbf{W}'_1)) \right\Vert + \sum_{j=1}^n \left[g(\tilde{\mathbf{A}})\right]_{ij} \left\Vert  (\hat{\mathbf{y}}_i(\mathbf{W}_2) - \mathbf{y}_i(\mathbf{W}'_2)) \otimes (\mathbf{X}_{j*} \mathbf{W}_1) \right\Vert \\
    \leq & \left( \sqrt{2} c_X \left\Vert g(\tilde{\mathbf{A}}) \right\Vert_{\infty} + c^2_X c^2_W \left\Vert g(\tilde{\mathbf{A}}) \right\Vert^2_{\infty} \right) \left\Vert {\rm vec}\left[\mathbf{W}_1\right] - {\rm vec}\left[\mathbf{W}'_1\right] \right\Vert.
\end{aligned}
\end{equation*}
Similarly,
\begin{equation*}
\begin{aligned}
    & \left\Vert \frac{\partial \ell(\mathbf{W}_1, \mathbf{W}_2;z_i)}{\partial {\rm vec} \left[ \mathbf{W}_2 \right]} (\mathbf{W}_2) -  \frac{\partial \ell(\mathbf{W}_1, \mathbf{W}_2;z_i)}{\partial {\rm vec} \left[ \mathbf{W}_2 \right]} (\mathbf{W}'_2) \right\Vert_2 \\ 
    \leq & \sum_{j=1}^n \left[g(\tilde{\mathbf{A}})\right]_{ij} \left\Vert  (\hat{\mathbf{y}}_i(\mathbf{W}_2) - \mathbf{y}_i(\mathbf{W}'_2)) \otimes (\mathbf{X}_{j*} \mathbf{W}_1) \right\Vert \leq c^2_X c^2_W \left\Vert g(\tilde{\mathbf{A}}) \right\Vert^2_{\infty} \left\Vert {\rm vec}\left[\mathbf{W}_2\right] - {\rm vec}\left[\mathbf{W}'_2\right] \right\Vert.
\end{aligned}
\end{equation*}
Denote by $P_{21} = \sqrt{2} c_X \left\Vert g(\tilde{\mathbf{A}}) \right\Vert_{\infty} + c^2_X c^2_W \left\Vert g(\tilde{\mathbf{A}}) \right\Vert^2_{\infty}, P_{22} = c^2_X c^2_W \left\Vert g(\tilde{\mathbf{A}}) \right\Vert^2_{\infty}$ and $\tilde{P}_{21} = \tilde{P}_{22} = 0$, we obtain that $\left\Vert \frac{\partial \ell(\mathbf{w};z_i)}{\partial {\rm vec} \left[ \mathbf{W}_2 \right]} - \frac{\partial \ell(\mathbf{w}';z_i)}{\partial {\rm vec} \left[ \mathbf{W}_2 \right]} \right\Vert_2 \leq \sum_{i=1}^2 P_{2i} \left\Vert {\rm vec}\left[\mathbf{W}_i\right] - {\rm vec}\left[\mathbf{W}'_i\right] \right\Vert_2 + \tilde{P}_{2i} \left\Vert {\rm vec}\left[\mathbf{W}_i\right] - {\rm vec}\left[\mathbf{W}'_i\right] \right\Vert^{\tilde{\alpha}}_2$. By the same way, denote by $P_{11} = P_{22}, \tilde{P}_{11} = \tilde{P}_{22}$ and $P_{12} = P_{21}, \tilde{P}_{12} = \tilde{P}_{21}$ as well as $\alpha_{11} = \alpha_{12} = 1$, we obtain that $\left\Vert \frac{\partial \ell(\mathbf{w};z_i)}{\partial {\rm vec} \left[ \mathbf{W}_1 \right]} - \frac{\partial \ell(\mathbf{w}';z_i)}{\partial {\rm vec} \left[ \mathbf{W}_1 \right]} \right\Vert_2 \leq \sum_{i=1}^2 P_{1i} \left\Vert {\rm vec}\left[\mathbf{W}_i\right] - {\rm vec}\left[\mathbf{W}'_i\right] \right\Vert_2 + \tilde{P}_{1i} \left\Vert {\rm vec}\left[\mathbf{W}_i\right] - {\rm vec}\left[\mathbf{W}'_i\right] \right\Vert^{\tilde{\alpha}}_2 $. By Lemma~\ref{bound_alpha}, we conclude that $\Vert \nabla \ell(\mathbf{w}) - \nabla \ell(\mathbf{w}') \Vert_2 \leq P_{\mathcal{F}} \mathop{\rm max} \{ \Vert \mathbf{w} - \mathbf{w}' \Vert_2, \Vert \mathbf{w} - \mathbf{w}' \Vert^{\tilde{\alpha}}_2 \}$ holds where $\mathbf{w}=\left[{\rm vec}\left[ \mathbf{W}_1 \right]; {\rm vec}\left[ \mathbf{W}_2 \right]\right]$.

\subsubsection{Proof of Proposition~\ref{appnp}}
We first show that the objective $\ell(\mathbf{W}_1, \mathbf{W}_2)$ is Lipschitz continuous w.r.t. $\mathbf{W}_1$ and $\mathbf{W}_2$. Note that
\begin{equation*}
\begin{aligned}
    & \vert \ell(\mathbf{W}_1, \mathbf{W}_2, z_i) - \ell(\mathbf{W}'_1, \mathbf{W}_2, z_i) \vert_2\\
    \leq & \sqrt{2} \left\Vert \sum_{j=1}^n \left[g(\tilde{\mathbf{A}})\right]_{ij}\sigma(\sigma(\mathbf{X}_{j*}\mathbf{W}_1)\mathbf{W}_2) - \sum_{j=1}^n \left[ g(\tilde{\mathbf{A}})\right]_{ij}\sigma(\sigma(\mathbf{X}_{j*}\mathbf{W}'_1)\mathbf{W}_2) \right\Vert_2 \\
    \leq & \sqrt{2} \sum_{j=1}^n \left\vert \left[g(\tilde{\mathbf{A}})\right]_{ij} \right\vert \left\Vert (\sigma(\mathbf{X}_{j*}\mathbf{W}_1) - \sigma(\mathbf{X}_{j*}\mathbf{W}'_1)) \mathbf{W}_2 \right\Vert_2 \\
    \leq & \sqrt{2} \sum_{j=1}^n \left\vert \left[g(\tilde{\mathbf{A}})\right]_{ij} \right\vert \Vert \sigma(\mathbf{X}_{j*}\mathbf{W}_1) - \sigma(\mathbf{X}_{j*}\mathbf{W}'_1) \Vert_2 \Vert \mathbf{W}_2 \Vert \\
    \leq & \sqrt{2} \sum_{j=1}^n \left\vert \left[g(\tilde{\mathbf{A}})\right]_{ij} \right\vert \Vert \mathbf{X}_{j*}(\mathbf{W}_1 - \mathbf{W}'_1) \Vert_2 \Vert \mathbf{W}_2 \Vert  \\
    \leq & c_{X}c_{W} \sqrt{2} \left\Vert g(\tilde{\mathbf{A}}) \right\Vert_{\infty} \Vert {\rm vec}(\mathbf{W}_1) - {\rm vec}({\mathbf{W}'_1}) \Vert.
\end{aligned}
\end{equation*}
Besides,
\begin{equation*}
\begin{aligned}
    & \vert \ell(\mathbf{W}_1, \mathbf{W}_2, z_i) - \ell(\mathbf{W}_1, \mathbf{W}'_2, z_i) \vert_2\\ 
    \leq & \sqrt{2} \left\Vert \sum_{j=1}^n \left[ g(\tilde{\mathbf{A}}) \right]_{ij} \left( \sigma(\sigma(\mathbf{X}_{j,:}\mathbf{W}_1)\mathbf{W}_2) - \sigma(\sigma(\mathbf{X}_{j,:}\mathbf{W}_1)\mathbf{W}'_2) \right) \right\Vert_2 \\
    \leq & \sqrt{2} \sum_{j=1}^n \left\vert \left[ g(\tilde{\mathbf{A}}) \right]_{ij} \right\vert \Vert \sigma(\mathbf{X}_{j,:}\mathbf{W}_1) (\mathbf{W}_2 - \mathbf{W}'_2) \Vert_2 \leq c_{X}c_{W} \sqrt{2} \left\Vert g(\tilde{\mathbf{A}}) \right\Vert_{\infty} \Vert \mathbf{W}_2 - \mathbf{W}'_2 \Vert.
\end{aligned}
\end{equation*}
By Lemma~\ref{bound_alpha}, we conclude that $\vert \ell(\mathbf{w}) - \ell(\mathbf{w}') \vert \leq L_{\mathcal{F}} \Vert \mathbf{w} - \mathbf{w}' \Vert_2$ holds with $L_\mathcal{F}=2c_X c_W \left\Vert g(\tilde{\mathbf{A}}) \right\Vert_{\infty}$. The gradients of $\ell$ w.r.t. $\mathbf{W}_1$ and $\mathbf{W}_2$ are
\begin{equation*}
\begin{aligned}
    & \frac{\partial \ell(\mathbf{W}_1, \mathbf{W}_2;z_i)}{\partial {\rm vec} \left[\mathbf{W}_2\right]} = \sum_{j=1}^n \left[g(\tilde{\mathbf{A}})\right]_{ij} (\sigma'(\mathbf{H}^{(1)}\mathbf{W}_2)_{j*} \odot (\hat{\mathbf{y}}_i - \mathbf{y}_i)) \otimes (\mathbf{X}\mathbf{W}_1)_{j*}, \\
    & \frac{\partial \ell(\mathbf{W}_1, \mathbf{W}_2;z_i)}{\partial {\rm vec} \left[\mathbf{W}_1\right]} = \sum_{j=1}^n \left[g(\tilde{\mathbf{A}})\right]_{ij} (\sigma'(\mathbf{X}\mathbf{W}_1)_{j*} \odot (((\hat{\mathbf{y}}_i - \mathbf{y}_i) \odot \sigma'(\mathbf{H}^{(1)}\mathbf{W}_2)_{j*}) \mathbf{W}^{\top}_2)) \otimes \mathbf{X}_{j*},
\end{aligned}
\end{equation*}
where $\mathbf{H}^{(1)} = \sigma(\mathbf{X}\mathbf{W}_1)$. Note that
\begin{equation*}
\begin{aligned}
    & \Vert \hat{\mathbf{y}}_i (\mathbf{W}_2) - \hat{\mathbf{y}}_i (\mathbf{W}'_2) \Vert_2 \leq \sum_{j=1}^n \left| \left[g(\tilde{\mathbf{A}})\right]_{ij} \right| \Vert \sigma(\mathbf{H}^{(1)}(\mathbf{W}_2 - \mathbf{W}'_2) )\Vert_2 \leq c_X c_W \left\Vert g(\tilde{\mathbf{A}}) \right\Vert_\infty \left\Vert {\rm vec} \left[\mathbf{W}_2\right] - {\rm vec} \left[ \mathbf{W}'_2 \right] \right\Vert_2.
\end{aligned}
\end{equation*}
Similarly,
\begin{equation*}
\begin{aligned}
    & \Vert \hat{\mathbf{y}}_i (\mathbf{W}_1) - \hat{\mathbf{y}}_i (\mathbf{W}'_1) \Vert_2
    \leq c_X c_W \left\Vert g(\tilde{\mathbf{A}}) \right\Vert_\infty \left\Vert {\rm vec} \left[\mathbf{W}_1 \right] - {\rm vec} \left[ \mathbf{W}'_1 \right] \right\Vert_2.
\end{aligned}
\end{equation*}

\textbf{Part A.} First we have
\begin{equation*}
\begin{aligned}
    & \left\Vert \frac{\partial \ell(\mathbf{W}_1, \mathbf{W}_2;z_i)}{\partial {\rm vec} \left[ \mathbf{W}_2 \right]} (\mathbf{W}_2) -  \frac{\partial \ell(\mathbf{W}_1, \mathbf{W}_2;z_i)}{\partial {\rm vec} \left[ \mathbf{W}_2 \right]} (\mathbf{W}'_2) \right\Vert_2 \\
    \leq & \sum_{j=1}^n \left| \left[g(\tilde{\mathbf{A}})\right]_{ij} \right| \sqrt{2}\Vert \mathbf{X}_{j*} \mathbf{W}_1 \Vert_2 \Vert \sigma'(\mathbf{H}^{(1)}\mathbf{W}_2)_{j*} - \sigma'(\mathbf{H}^{(1)}\mathbf{W}'_2)_{j*} \Vert_2 \\
    & + \sum_{j=1}^n \left| \left[g(\tilde{\mathbf{A}})\right]_{ij} \right| \Vert \mathbf{X}_{j*} \mathbf{W}_1 \Vert_2 \Vert \hat{\mathbf{y}}_i (\mathbf{W}_2) - \hat{\mathbf{y}}_i (\mathbf{W}'_2) \Vert_2 \\
    \leq & \left\Vert g(\tilde{\mathbf{A}}) \right\Vert_{\infty} \sqrt{|\mathcal{Y}|} c^{1+\tilde{\alpha}}_X c^{1+\tilde{\alpha}}_W \left\Vert {\rm vec} \left[\mathbf{W}_2\right] - {\rm vec} \left[ \mathbf{W}'_2 \right] \right\Vert^{\tilde{\alpha}}_2 + c^2_X c^2_W \left\Vert g(\tilde{\mathbf{A}}) \right\Vert^2_\infty \left\Vert {\rm vec} \left[\mathbf{W}_2 \right] - {\rm vec} \left[ \mathbf{W}'_2 \right] \right\Vert_2.
\end{aligned}
\end{equation*}
Also,
\begin{equation*}
\begin{aligned}
    & \left\Vert \frac{\partial \ell(\mathbf{W}_1, \mathbf{W}_2;z_i)}{\partial {\rm vec} \left[ \mathbf{W}_2 \right]} (\mathbf{W}_1) -  \frac{\partial \ell(\mathbf{W}_1, \mathbf{W}_2;z_i)}{\partial {\rm vec} \left[ \mathbf{W}_2 \right]} (\mathbf{W}'_1) \right\Vert_2 \\
    \leq & \sqrt{2} \sum_{j=1}^n \left| \left[g(\tilde{\mathbf{A}})\right]_{ij} \right| \Vert \mathbf{X}_{j*} \mathbf{W}_1 \Vert_2 \Vert \sigma'(\mathbf{H}^{(1)}(\mathbf{W}_1)\mathbf{W}_2)_{j*} - \sigma'(\mathbf{H}^{(1)}(\mathbf{W}'_1)\mathbf{W}_2)_{j*} \Vert_2 \\
    & + \sqrt{2} \sum_{j=1}^n \left| \left[g(\tilde{\mathbf{A}})\right]_{ij} \right| \Vert \mathbf{X}_{j*}(\mathbf{W}_1 - \mathbf{W}'_1) \Vert_2 + \sum_{j=1}^n \left| \left[g(\tilde{\mathbf{A}})\right]_{ij} \right| \Vert \hat{\mathbf{y}}_i(\mathbf{W}_1) - \hat{\mathbf{y}}_i(\mathbf{W}'_1) \Vert_2 \Vert \mathbf{X}_{j*} \mathbf{W}_1 \Vert_2 \\
    \leq & \sqrt{2} P \left\Vert g(\tilde{\mathbf{A}}) \right\Vert_{\infty} c^{1+\tilde{\alpha}}_X c^{1+\tilde{\alpha}}_W \Vert {\rm vec} \left[\mathbf{W}_1\right] - {\rm vec} \left[\mathbf{W}'_1 \right] \Vert^{\tilde{\alpha}}_2 \\
    & + \left( \sqrt{2} c_X \left\Vert g(\tilde{\mathbf{A}}) \right\Vert_{\infty} + c^2_X c^2_W \left\Vert g(\tilde{\mathbf{A}}) \right\Vert^2_\infty \right) \Vert {\rm vec} \left[ \mathbf{W}_1 \right] - {\rm vec} \left[\mathbf{W}'_1 \right] \Vert_2 . 
\end{aligned}    
\end{equation*}
Denote by 
\begin{equation*}
\begin{aligned}
    & P_{21} = \left( \sqrt{2} c_X \left\Vert g(\tilde{\mathbf{A}}) \right\Vert_{\infty} + c^2_X c^2_W \left\Vert g(\tilde{\mathbf{A}}) \right\Vert^2_\infty \right), P_{22} = c^2_X c^2_W \left\Vert g(\tilde{\mathbf{A}}) \right\Vert^2_\infty , \\
    & \tilde{P}_{21} = \left\Vert g(\tilde{\mathbf{A}}) \right\Vert_{\infty} \sqrt{|\mathcal{Y}|} c^{1+\tilde{\alpha}}_X c^{1+\tilde{\alpha}}_W, \ \tilde{P}_{22} = \sqrt{2} P \left\Vert g(\tilde{\mathbf{A}}) \right\Vert_{\infty} c^{1+\tilde{\alpha}}_X c^{1+\tilde{\alpha}}_W,
\end{aligned}
\end{equation*}
we obtain that 
$$\left\Vert \frac{\partial \ell(\mathbf{w};z_i)}{\partial {\rm vec} \left[ \mathbf{W}_2 \right]} - \frac{\partial \ell(\mathbf{w}';z_i)}{\partial {\rm vec} \left[ \mathbf{W}_2 \right]} \right\Vert_2 \leq \sum_{i=1}^2 P_{2i} \left\Vert {\rm vec}\left[\mathbf{W}_i\right] - {\rm vec}\left[\mathbf{W}'_i\right] \right\Vert_2 + \tilde{P}_{2i} \left\Vert {\rm vec}\left[\mathbf{W}_i\right] - {\rm vec}\left[\mathbf{W}'_i\right] \right\Vert^{\tilde{\alpha}}_2.$$

\textbf{Part B.}
We have
\begin{equation*}
\begin{aligned}
    & \left\Vert \frac{\partial \ell(\mathbf{W}_1, \mathbf{W}_2;z_i)}{\partial {\rm vec} \left[ \mathbf{W}_1 \right]} (\mathbf{W}_2) -  \frac{\partial \ell(\mathbf{W}_1, \mathbf{W}_2;z_i)}{\partial {\rm vec} \left[ \mathbf{W}_1 \right]} (\mathbf{W}'_2) \right\Vert_2 \\
    \leq & c_X c_W \sqrt{2} \sum_{j=1}^n \left[g(\tilde{\mathbf{A}})\right]_{ij} \left\Vert \sigma'(\mathbf{H}^{(1)}\mathbf{W}_2)_{j*} - \sigma'(\mathbf{H}^{(1)}\mathbf{W}'_2)_{j*} \right\Vert_2 + c_X \sqrt{2} \sum_{j=1}^n \left[g(\tilde{\mathbf{A}})\right]_{ij} \left\Vert \mathbf{W}_2 - \mathbf{W}'_2 \right\Vert_2 \\
    & + c_W \sum_{j=1}^n \left[g(\tilde{\mathbf{A}})\right]_{ij} \Vert \mathbf{X}_{j*} \Vert_2 \Vert \hat{\mathbf{y}}_i (\mathbf{W}_2) - \hat{\mathbf{y}}_i (\mathbf{W}'_2) \Vert_2  \\
    \leq & c^{1+\tilde{\alpha}}_X c^{1+\tilde{\alpha}}_W P \sqrt{2} \left\Vert g(\tilde{\mathbf{A}}) \right\Vert_\infty \Vert {\rm vec} \left[ \mathbf{W}_2 \right] - {\rm vec} \left[\mathbf{W}'_2 \right] \Vert^{\tilde{\alpha}}_2 \\
    & + \left( c_X \sqrt{2} \left\Vert g(\tilde{\mathbf{A}}) \right\Vert_\infty + c^2_X c^2_W \left\Vert g(\tilde{\mathbf{A}}) \right\Vert^2_\infty \right) \Vert {\rm vec} \left[ \mathbf{W}_2 \right] - {\rm vec} \left[\mathbf{W}'_2 \right] \Vert_2.
\end{aligned}
\end{equation*}
Also, one can find that
\begin{equation*}
\begin{aligned}
    & \left\Vert \frac{\partial \ell(\mathbf{W}_1, \mathbf{W}_2;z_i)}{\partial {\rm vec} \left[ \mathbf{W}_1 \right]} (\mathbf{W}_1) -  \frac{\partial \ell(\mathbf{W}_1, \mathbf{W}_2;z_i)}{\partial {\rm vec} \left[ \mathbf{W}_1 \right]} (\mathbf{W}'_1) \right\Vert_2 \\
    \leq & c_X c_W \sqrt{2} \sum_{j=1}^n \left[g(\tilde{\mathbf{A}})\right]_{ij} \left\Vert \sigma'(\mathbf{X}\mathbf{W}_1)_{j*} - \sigma'(\mathbf{X}\mathbf{W}'_1)_{j*} \right\Vert_2 +c_W \sum_{j=1}^n \left[g(\tilde{\mathbf{A}})\right]_{ij} \Vert \mathbf{X}_{j*} \Vert_2 \Vert \hat{\mathbf{y}}_i (\mathbf{W}_1) - \hat{\mathbf{y}}_i (\mathbf{W}'_1) \Vert_2 \\
    & + c_X c_W \sqrt{2} \sum_{j=1}^n \left[g(\tilde{\mathbf{A}})\right]_{ij} \left\Vert \sigma'(\mathbf{H}^{(1)}(\mathbf{W}_1)\mathbf{W}_2) - \sigma'(\mathbf{H}^{(1)}(\mathbf{W}'_1)\mathbf{W}_2) \right\Vert_2 \\
    \leq & c_X c_W P \sqrt{2} \sum_{j=1}^n \left[g(\tilde{\mathbf{A}})\right]_{ij} \left\Vert \mathbf{X}_{j*}(\mathbf{W}_1 - \mathbf{W}'_1 ) \right\Vert^{\tilde{\alpha}}_2  + c_X c_W P \sqrt{2} \sum_{j=1}^n \left[g(\tilde{\mathbf{A}})\right]_{ij} \left\Vert (\mathbf{H}^{(1)}(\mathbf{W}_1) - \mathbf{H}^{(1)}(\mathbf{W}_1) )\mathbf{W}_2 \right\Vert^{\tilde{\alpha}}_2 \\
    \leq & \left[ c^{1+\tilde{\alpha}}_X c_W + c^{1+\tilde{\alpha}}_X c^{1+\tilde{\alpha}}_W \right] P \sqrt{2} \left\Vert g(\tilde{\mathbf{A}}) \right\Vert_\infty \Vert {\rm vec} \left[ \mathbf{W}_1 \right] - {\rm vec} \left[\mathbf{W}'_1 \right] \Vert^{\tilde{\alpha}}_2 + c^2_X c^2_W \left\Vert g(\tilde{\mathbf{A}}) \right\Vert^2_\infty \Vert {\rm vec} \left[ \mathbf{W}_1 \right] - {\rm vec} \left[\mathbf{W}'_1 \right] \Vert_2.
\end{aligned}
\end{equation*}
Denote by 
\begin{equation*}
\begin{aligned}
    & P_{11} = c^2_X c^2_W \left\Vert g(\tilde{\mathbf{A}}) \right\Vert^2_\infty, \ P_{12} = \left( c_X \sqrt{2} \left\Vert g(\tilde{\mathbf{A}}) \right\Vert_\infty + c^2_X c^2_W \left\Vert g(\tilde{\mathbf{A}}) \right\Vert^2_\infty \right), \\
    & \tilde{P}_{11} = c^{1+\tilde{\alpha}}_X c^{1+\tilde{\alpha}}_W P \sqrt{2} \left\Vert g(\tilde{\mathbf{A}}) \right\Vert_\infty, \ \tilde{P}_{12} = \left[ c^{1+\tilde{\alpha}}_X c_W + c^{1+\tilde{\alpha}}_X c^{1+\tilde{\alpha}}_W \right] P \sqrt{2} \left\Vert g(\tilde{\mathbf{A}}) \right\Vert_\infty, 
\end{aligned}
\end{equation*}
we obtain that 
$$\left\Vert \frac{\partial \ell(\mathbf{w};z_i)}{\partial {\rm vec} \left[ \mathbf{W}_1 \right]} - \frac{\partial \ell(\mathbf{w}';z_i)}{\partial {\rm vec} \left[ \mathbf{W}_1 \right]} \right\Vert_2 \leq \sum_{i=1}^2 P_{1i} \left\Vert {\rm vec}\left[\mathbf{W}_i\right] - {\rm vec}\left[\mathbf{W}'_i\right] \right\Vert_2 + \sum_{i=1}^2 \tilde{P}_{1i} \left\Vert {\rm vec}\left[\mathbf{W}_i\right] - {\rm vec}\left[\mathbf{W}'_i\right] \right\Vert^{\tilde{\alpha}}_2.$$ By Lemma~\ref{bound_alpha}, we conclude that $\Vert \nabla \ell(\mathbf{w}) - \nabla \ell(\mathbf{w}') \Vert_2 \leq P_{\mathcal{F}} \mathop{\rm max} \{ \Vert \mathbf{w} - \mathbf{w}' \Vert_2, \Vert \mathbf{w} - \mathbf{w}' \Vert^{\tilde{\alpha}}_2 \}$ holds where $\mathbf{w}=\left[{\rm vec}\left[ \mathbf{W}_1 \right]; {\rm vec}\left[ \mathbf{W}_2 \right]\right]$.

\subsubsection{Proof of Proposition~\ref{gpr}}
We first show that the objective $\ell(\mathbf{W}_1, \mathbf{W}_2, \gamma)$ is Lipschitz continuous w.r.t. $\mathbf{W}_1$, $\mathbf{W}_2$ and $\gamma$. Note that
\begin{equation*}
\begin{aligned}
    & \vert \ell(\mathbf{W}_1, \mathbf{W}_2, \bm \gamma) - \ell(\mathbf{W}_1, \mathbf{W}_2, {\bm \gamma}') \vert \\
    \leq & \sqrt{2} \left\Vert \sum_{j=1}^n \left[g(\tilde{\mathbf{A}}, {\bm \gamma})\right]_{ij} \sigma(\sigma(\mathbf{X}_{j*}\mathbf{W}_1)\mathbf{W}_2) - \sum_{j=1}^n \left[g(\tilde{\mathbf{A}}, {\bm \gamma}')\right]_{ij} \sigma(\sigma(\mathbf{X}_{j*}\mathbf{W}_1)\mathbf{W}_2) \right\Vert_2\\
    = & \sqrt{2} \left\Vert \sum_{k=0}^K (\gamma_k - \gamma'_k) \left( \sum_{j=1}^n \left[ \tilde{\mathbf{A}}^k \right]_{ij} \sigma(\sigma(\mathbf{X}_{j*}\mathbf{W}_1)\mathbf{W}_2) \right) \right\Vert_2 \\
    \leq & \sqrt{2} \Vert {\bm \gamma} -{\bm \gamma}' \Vert_2 \sqrt{ \sum_{k=0}^K \left\Vert \sum_{j=1}^n \left[ \tilde{\mathbf{A}}^k \right]_{ij} \sigma(\sigma(\mathbf{X}_{j*}\mathbf{W}_1)\mathbf{W}_2) \right\Vert^2_2} \\
    \leq & \sqrt{2} \Vert {\bm \gamma} -{\bm \gamma}' \Vert_2 \sum_{k=0}^K \left\Vert \sum_{j=1}^n \left[ \tilde{\mathbf{A}}^k \right]_{ij} \sigma(\sigma(\mathbf{X}_{j*}\mathbf{W}_1)\mathbf{W}_2) \right\Vert_2 \leq \sqrt{2} c_X c^2_W \left( \sum_{k=0}^K \left\Vert \tilde{\mathbf{A}}^k \right\Vert_\infty \right) \Vert {\bm \gamma} -{\bm \gamma}' \Vert_2.
\end{aligned}
\end{equation*}
Denote by
\begin{equation*}
    L_{\mathcal{F}} = \sqrt{2 c^2_X c^4_W \left( \sum_{k=0}^K \left\Vert \tilde{\mathbf{A}}^k \right\Vert_\infty \right)^2 + 4 c^2_X c^2_W \left\Vert g(\tilde{\mathbf{A}}, \bm \gamma) \right\Vert^2_\infty},
\end{equation*}
we conclude that $\vert \ell(\mathbf{w}) - \ell(\mathbf{w}') \vert \leq L_{\mathcal{F}} \Vert \mathbf{w} - \mathbf{w}' \Vert_2$ holds. Then we discuss the smoothness of this model. The gradients of $\ell(\mathbf{W}_1,\mathbf{W}_2,\gamma)$ w.r.t. $\mathbf{W}_1$, $\mathbf{W}_2$, and $\gamma$ are
\begin{equation*}\small
\begin{aligned}
    & \frac{\partial \ell(\mathbf{W}_1, \mathbf{W}_2, \bm \gamma;z_i)}{\partial {\rm vec} \left[\mathbf{W}_2\right]} = \sum_{j=1}^n \left[ g(\tilde{\mathbf{A}}, \bm \gamma)\right]_{ij} (\sigma'(\sigma(\mathbf{X}\mathbf{W}_1)\mathbf{W}_2)_{j*} \odot (\hat{\mathbf{y}}_i - \mathbf{y}_i)) \otimes (\mathbf{X}\mathbf{W}_1)_{j*}, \\
    & \frac{\partial \ell(\mathbf{W}_1, \mathbf{W}_2, \bm \gamma;z_i)}{\partial {\rm vec} \left[\mathbf{W}_1\right]} = \sum_{j=1}^n \left[g(\tilde{\mathbf{A}}, \gamma)\right]_{ij} (\sigma'(\mathbf{X}\mathbf{W}_1)_{j,:} \odot (((\hat{\mathbf{y}}_i - \mathbf{y}_i) \odot \sigma'(\sigma(\mathbf{X}\mathbf{W}_1)\mathbf{W}_2)_{j*}) \mathbf{W}^{\top}_2)) \otimes \mathbf{X}_{j*} , \\
    & \frac{\partial \ell(\mathbf{W}_1, \mathbf{W}_2, \bm \gamma;z_i)}{\partial {\bm \gamma}} = (\hat{\mathbf{y}}_i - \mathbf{y}_i) \left[ \sum_{j=1}^n \left[\tilde{\mathbf{A}}^0 \right]_{ij}\mathbf{H}^\top_{j*}, \ldots, \sum_{j=1}^n \left[\tilde{\mathbf{A}}^K \right]_{ij}\mathbf{H}^\top_{j*} \right],
\end{aligned}
\end{equation*}
where $\mathbf{H} = \sigma(\sigma(\mathbf{X}\mathbf{W}_1)\mathbf{W}_2)$. Note that
\begin{equation*}
\begin{aligned}
    & \Vert \hat{\mathbf{y}}_i({\bm \gamma}) - \hat{\mathbf{y}}_i({\bm \gamma}') \Vert_2 \leq c_X c^2_W \left( \sum_{k=0}^K \left\Vert \tilde{\mathbf{A}}^k \right\Vert_\infty \right) \Vert {\bm \gamma} -{\bm \gamma}' \Vert_2.
\end{aligned}
\end{equation*}
Then one can find that
\begin{equation*}\small
\begin{aligned}
    & \left\Vert \frac{\partial \ell(\mathbf{W}_1, \mathbf{W}_2, \bm \gamma;z_i)}{\partial {\rm vec} \left[\mathbf{W}_2\right]}(\bm \gamma) - \frac{\partial \ell(\mathbf{W}_1, \mathbf{W}_2, \bm \gamma;z_i)}{\partial {\rm vec} \left[\mathbf{W}_2\right]} ({\bm \gamma}') \right\Vert_2 \\
    \leq & \left\Vert \sum_{k=0}^K (\gamma_k - \gamma'_k) \left( \sum_{j=1}^n \left[ \tilde{\mathbf{A}}^k \right]_{ij} (\sigma'(\sigma(\mathbf{X}\mathbf{W}_1)\mathbf{W}_2)_{j*} \odot (\hat{\mathbf{y}}_i - \mathbf{y}_i)) \otimes (\mathbf{X}\mathbf{W}_1)_{j*} \right) \right\Vert \\
    & + \sum_{j=1}^n \left\vert \left[ g(\tilde{\mathbf{A}}, \bm \gamma)\right]_{ij} \right\vert \Vert \hat{\mathbf{y}}_i(\bm \gamma) - \hat{\mathbf{y}}_i({\bm \gamma}') \Vert_2 \Vert (\mathbf{X}\mathbf{W}_1)_{j*} \Vert_2 \\
    \leq & \left( \sqrt{2} + c_X c_W \left\Vert g(\tilde{\mathbf{A}}, \bm \gamma) \right\Vert_\infty \right) c_X c^2_W \left( \sum_{k=0}^K \left\Vert \tilde{\mathbf{A}}^k \right\Vert_\infty \right) \Vert {\bm \gamma} - {\bm \gamma}' \Vert_2 .
\end{aligned}
\end{equation*}
Denote by
\begin{equation*}
\begin{aligned}
    & P_{21} = c^2_X c^2_W \left\Vert g(\tilde{\mathbf{A}}, \bm \gamma) \right\Vert^2_\infty, \ P_{22} =  \left( \sqrt{2} c_X \left\Vert g(\tilde{\mathbf{A}}, \bm \gamma) \right\Vert_{\infty} + c^2_X c^2_W \left\Vert g(\tilde{\mathbf{A}}, \bm \gamma) \right\Vert^2_\infty \right), \\
    & P_{23} = \left( \sqrt{2} + c_X c_W \left\Vert g(\tilde{\mathbf{A}}, \bm \gamma) \right\Vert_\infty \right) c_X c^2_W \left( \sum_{k=0}^K \left\Vert \tilde{\mathbf{A}}^k \right\Vert_\infty \right), \\
    & \tilde{P}_{21} = \left\Vert g(\tilde{\mathbf{A}}, \bm \gamma) \right\Vert_{\infty} \sqrt{|\mathcal{Y}|} c^{1+\tilde{\alpha}}_X c^{1+\tilde{\alpha}}_W, \ \tilde{P}_{22} = \sqrt{2} P \left\Vert g(\tilde{\mathbf{A}}, \bm \gamma) \right\Vert_{\infty} c^{1+\tilde{\alpha}}_X c^{1+\tilde{\alpha}}_W, \ \tilde{P}_{23} = 0,
\end{aligned}
\end{equation*}
we obtain that 
\begin{equation*}
\begin{aligned}
    \left\Vert \frac{\partial \ell(\mathbf{w};z_i)}{\partial {\rm vec} \left[ \mathbf{W}_1 \right]} - \frac{\partial \ell(\mathbf{w}';z_i)}{\partial {\rm vec} \left[ \mathbf{W}_1 \right]} \right\Vert_2 \leq & \sum_{i=1}^2 P_{2i} \left\Vert {\rm vec}\left[\mathbf{W}_i\right] - {\rm vec}\left[\mathbf{W}'_i\right] \right\Vert_2 + \sum_{i=1}^2 \tilde{P}_{2i} \left\Vert {\rm vec}\left[\mathbf{W}_i\right] - {\rm vec}\left[\mathbf{W}'_i\right] \right\Vert^{\tilde{\alpha}}_2 \\
    & + P_{23} \Vert {\bm \gamma} - {\bm \gamma}' \Vert_2 + \tilde{P}_{23} \Vert {\bm \gamma} - {\bm \gamma}' \Vert^{\tilde{\alpha}}_2.    
\end{aligned}
\end{equation*}

Besides,
\begin{equation*}\small
\begin{aligned}
    & \left\Vert \frac{\partial \ell(\mathbf{W}_1, \mathbf{W}_2, \bm \gamma;z_i)}{\partial {\rm vec} \left[\mathbf{W}_1\right]}(\bm \gamma) - \frac{\partial \ell(\mathbf{W}_1, \mathbf{W}_2, \bm \gamma;z_i)}{\partial {\rm vec} \left[\mathbf{W}_1\right]} ({\bm \gamma}') \right\Vert \\
    = & \left\Vert \sum_{k=0}^{K} (\gamma_k - \gamma'_k) \left( \sum_{j=1}^n \left[\tilde{\mathbf{A}}^k \right]_{ij} (\sigma'(\mathbf{X}\mathbf{W}_1)_{j,:} \odot (((\hat{\mathbf{y}}_i - \mathbf{y}_i) \odot \sigma'(\sigma(\mathbf{X}\mathbf{W}_1)\mathbf{W}_2)_{j*}) \mathbf{W}^{\top}_2)) \otimes \mathbf{X}_{j*} \right) \right\Vert_2 \\
    & + \sum_{j=1}^n c_X c_W \left\vert \left[ g(\tilde{\mathbf{A}}, \bm \gamma)\right]_{ij} \right\vert \Vert \hat{\mathbf{y}}_i(\bm \gamma) - \hat{\mathbf{y}}_i({\bm \gamma}') \Vert_2 \\
    \leq & \left( \sqrt{2} + c_X c_W \left\Vert g(\tilde{\mathbf{A}}, \bm \gamma) \right\Vert_\infty \right) c_X c^2_W \left( \sum_{k=0}^K \left\Vert \tilde{\mathbf{A}}^k \right\Vert_\infty \right) \Vert {\bm \gamma} - {\bm \gamma}' \Vert_2.
\end{aligned}
\end{equation*}
Denote by 
\begin{equation*}
\begin{aligned}
    & P_{11} =  c^2_X c^2_W \left\Vert g(\tilde{\mathbf{A}}, \bm \gamma) \right\Vert^2_\infty, \ P_{12} =  \left( c_X \sqrt{2} \left\Vert g(\tilde{\mathbf{A}}, \bm \gamma) \right\Vert_\infty + c^2_X c^2_W \left\Vert g(\tilde{\mathbf{A}}, \bm \gamma) \right\Vert^2_\infty \right), \\
    & P_{13} = \left( \sqrt{2} + c_X c_W \left\Vert g(\tilde{\mathbf{A}}, \bm \gamma) \right\Vert_\infty \right) c_X c^2_W \left( \sum_{k=0}^K \left\Vert \tilde{\mathbf{A}}^k \right\Vert_\infty \right) \\
    & \tilde{P}_{11} = \sqrt{2} P \left[ c^{1+\tilde{\alpha}}_X c_W + c^{1+\tilde{\alpha}}_X c^{1+\tilde{\alpha}}_W \right] , \ \tilde{P}_{12} = c^{1+\tilde{\alpha}}_X c^{1+\tilde{\alpha}}_W P \sqrt{2} \left\Vert g(\tilde{\mathbf{A}}, \bm \gamma) \right\Vert_\infty, \tilde{P}_{13} = 0, \\
\end{aligned}
\end{equation*}
we obtain that 
\begin{equation*}
\begin{aligned}
    \left\Vert \frac{\partial \ell(\mathbf{w};z_i)}{\partial {\rm vec} \left[ \mathbf{W}_1 \right]} - \frac{\partial \ell(\mathbf{w}';z_i)}{\partial {\rm vec} \left[ \mathbf{W}_1 \right]} \right\Vert_2 \leq & \sum_{i=1}^2 P_{1i} \left\Vert {\rm vec}\left[\mathbf{W}_i\right] - {\rm vec}\left[\mathbf{W}'_i\right] \right\Vert_2 + \sum_{i=1}^2 \tilde{P}_{1i} \left\Vert {\rm vec}\left[\mathbf{W}_i\right] - {\rm vec}\left[\mathbf{W}'_i\right] \right\Vert^{\tilde{\alpha}}_2 \\
    & + P_{13} \Vert {\bm \gamma} - {\bm \gamma}' \Vert_2 + \tilde{P}_{13} \Vert {\bm \gamma} - {\bm \gamma}' \Vert^{\tilde{\alpha}}_2.    
\end{aligned}
\end{equation*}

Lastly, since
\begin{equation*}
\begin{aligned}
    & \left\Vert \frac{\partial \ell(\mathbf{W}_1, \mathbf{W}_2, \bm \gamma;z_i)}{\partial {\bm \gamma}}(\mathbf{W}_2) - \frac{\partial \ell(\mathbf{W}_1, \mathbf{W}_2, \bm \gamma;z_i)}{\partial {\bm \gamma}}(\mathbf{W}'_2) \right\Vert_2 \\
    \leq & \Vert \hat{\mathbf{y}}_i - \mathbf{y}_i \Vert_2 \sqrt{ \sum_{k=0}^K \left\Vert \sum_{j=1}^n [\tilde{\mathbf{A}}^k]_{ij} (\sigma'(\sigma(\mathbf{X}_{j*}\mathbf{W}_1)\mathbf{W}_2) - \sigma'(\sigma(\mathbf{X}_{j*}\mathbf{W}_1)\mathbf{W}'_2)) \right\Vert^2_2} \\
    & + \Vert \hat{\mathbf{y}}_i(\mathbf{W}_2) - \hat{\mathbf{y}}_i(\mathbf{W}_2) \Vert_2 \sqrt{ \sum_{k=0}^K \left\Vert \sum_{j=1}^n [\tilde{\mathbf{A}}^k]_{ij} \sigma'(\sigma(\mathbf{X}_{j*}\mathbf{W}_1)\mathbf{W}_2) \right\Vert^2_2} \\ 
    \leq & \sqrt{2} P c^{\tilde{\alpha}}_X c^{\tilde{\alpha}}_W \left( \sum_{k=0}^K \left\Vert \tilde{\mathbf{A}}^k \right\Vert_\infty \right) \Vert {\rm vec} \left[ \mathbf{W}_2 \right] - {\rm vec} \left[\mathbf{W}'_2 \right] \Vert^{\tilde{\alpha}}_2 \\
    & + c^2_X c^3_W \left\Vert g(\tilde{\mathbf{A}}, \bm \gamma) \right\Vert_\infty \left( \sum_{k=0}^K \left\Vert \tilde{\mathbf{A}}^k \right\Vert_\infty \right) \Vert {\rm vec} \left[ \mathbf{W}_2 \right] - {\rm vec} \left[\mathbf{W}'_2 \right] \Vert_2.
\end{aligned}
\end{equation*}
Similarly, we have
\begin{equation*}
\begin{aligned}
    & \left\Vert \frac{\partial \ell(\mathbf{W}_1, \mathbf{W}_2, \bm \gamma;z_i)}{\partial {\bm \gamma}}(\mathbf{W}_1) - \frac{\partial \ell(\mathbf{W}_1, \mathbf{W}_2, \bm \gamma;z_i)}{\partial {\bm \gamma}}(\mathbf{W}'_1) \right\Vert_2 \\
    \leq & \sqrt{2} P c^{\tilde{\alpha}}_X c^{\tilde{\alpha}}_W \left( \sum_{k=0}^K \left\Vert \tilde{\mathbf{A}}^k \right\Vert_\infty \right) \Vert {\rm vec} \left[ \mathbf{W}_1 \right] - {\rm vec} \left[\mathbf{W}'_1 \right] \Vert^{\tilde{\alpha}}_2 \\
    & + c^2_X c^3_W \left\Vert g(\tilde{\mathbf{A}}, \bm \gamma) \right\Vert_\infty \left( \sum_{k=0}^K \left\Vert \tilde{\mathbf{A}}^k \right\Vert_\infty \right) \Vert {\rm vec} \left[ \mathbf{W}_1 \right] - {\rm vec} \left[\mathbf{W}'_1 \right] \Vert_2,
\end{aligned}
\end{equation*}
and
\begin{equation*}
\begin{aligned}
    & \left\Vert \frac{\partial \ell(\mathbf{W}_1, \mathbf{W}_2, \bm \gamma;z_i)}{\partial {\bm \gamma}}(\bm \gamma) - \frac{\partial \ell(\mathbf{W}_1, \mathbf{W}_2, \bm \gamma;z_i)}{\partial {\bm \gamma}}({\bm \gamma}') \right\Vert_2 \\
    \leq & \Vert \hat{\mathbf{y}}_i(\bm \gamma) - \hat{\mathbf{y}}_i({\bm \gamma}') \Vert_2 \sqrt{ \sum_{k=0}^K \left\Vert \sum_{j=1}^n \left[ \tilde{\mathbf{A}}^k \right]_{ij} \sigma(\sigma(\mathbf{X}_{j*}\mathbf{W}_1)\mathbf{W}_2) \right\Vert^2_2} \leq c^2_X c^4_W \left( \sum_{k=0}^K \left\Vert \tilde{\mathbf{A}}^k \right\Vert_\infty \right)^2 \Vert {\bm \gamma} - {\bm \gamma}' \Vert_2 .
\end{aligned}
\end{equation*}
Denote by 
\begin{equation*}
\begin{aligned}
    & P_{31} = P_{32} = c^2_X c^3_W \left\Vert g(\tilde{\mathbf{A}}, \bm \gamma) \right\Vert_\infty \left( \sum_{k=0}^K \left\Vert \tilde{\mathbf{A}}^k \right\Vert_\infty \right), \ P_{33} = c^2_X c^4_W \left( \sum_{k=0}^K \left\Vert \tilde{\mathbf{A}}^k \right\Vert_\infty \right)^2, \\
    & \tilde{P}_{31} = \tilde{P}_{32} = \sqrt{2} P c^{\tilde{\alpha}}_X c^{\tilde{\alpha}}_W \left( \sum_{k=0}^K \left\Vert \tilde{\mathbf{A}}^k \right\Vert_\infty \right), \ \tilde{P}_{33} = 0,
\end{aligned}
\end{equation*}
we obtain that 
\begin{equation*}
\begin{aligned}
    \left\Vert \frac{\partial \ell(\mathbf{w};z_i)}{\partial {\bm \gamma}} - \frac{\partial \ell(\mathbf{w}';z_i)}{\partial {\bm \gamma}} \right\Vert_2 \leq & \sum_{i=1}^2 P_{3i} \left\Vert {\rm vec}\left[\mathbf{W}_i\right] - {\rm vec}\left[\mathbf{W}'_i\right] \right\Vert_2 + \sum_{i=1}^2 \tilde{P}_{3i} \left\Vert {\rm vec}\left[\mathbf{W}_i\right] - {\rm vec}\left[\mathbf{W}'_i\right] \right\Vert^{\tilde{\alpha}}_2 \\
    & + P_{33} \Vert {\bm \gamma} - {\bm \gamma}' \Vert_2 + \tilde{P}_{33} \Vert {\bm \gamma} - {\bm \gamma}' \Vert^{\tilde{\alpha}}_2.    
\end{aligned}
\end{equation*}
By Lemma~\ref{bound_alpha}, we conclude that $\Vert \nabla \ell(\mathbf{w}) - \nabla \ell(\mathbf{w}') \Vert_2 \leq P_{\mathcal{F}} \mathop{\rm max} \{ \Vert \mathbf{w} - \mathbf{w}' \Vert_2, \Vert \mathbf{w} - \mathbf{w}' \Vert^{\tilde{\alpha}}_2 \}$ holds where $\mathbf{w}=\left[{\rm vec}\left[ \mathbf{W}_1 \right]; {\rm vec}\left[ \mathbf{W}_2 \right]; {\bm \gamma}\right]$.

\section{Experiments Details}

For GCN, GAT, SGC, APPNP and GCNII, we adopt the official PyTorch Geometric library implementations \cite{pyg}. For GPR-GNN, we adopt the released codes \footnote{\href{https://github.com/jianhao2016/GPRGNN}{https://github.com/jianhao2016/GPRGNN}} with commit number 2507f10. Following \cite{cong2021on}, we remove all dropout layers and adopt the Adam optimizer with default setting. The batch size is set to $512$ and the number of hidden units are set to $64$ for all baseline models. $K$ is set to $10$ for APPNP and GPR-GNN.

\end{document}